%% file: main.tex
\documentclass[10pt]{article}

\usepackage{meowarxiv}

\usepackage{graphicx}
\usepackage{makecell}
\usepackage{booktabs}
\usepackage{multirow}
\usepackage{multicol}
\usepackage{algorithm}
\usepackage{algorithmicx}
\usepackage{colortbl}
\usepackage{ntheorem}
\usepackage{wrapfig}
\usepackage{tabularx}
\usepackage{amsmath}
\usepackage{xcolor}
\usepackage{tikz}
\usepackage{url}

\definecolor{myblue3}{rgb}{0.1019,0.1372,0.4941}
\colorlet{mypurpleee}{meowplum}
\definecolor{explorationRose}{HTML}{B15C79}
\definecolor{explorationBlue}{HTML}{495B96}

\newcommand{\ind}{\mathbf{1}}
\newcommand{\clip}{\operatorname{clip}}

\newcommand{\nntask}[0]{ten~}

\newcommand{\pass}[0]{\textit{Pass}~}
\newcommand{\thumbaround}[0]{\textit{Thumb-Around}~}
\newcommand{\sonic}[0]{\textit{Sonic}~}
\newcommand{\sonicdown}[0]{\textit{Sonic}\texttt{ (down)}~}
\newcommand{\trianglepass}[0]{\textit{Triangle Pass}~}
\newcommand{\trianglepassup}[0]{\textit{Triangle Pass} \texttt{(palm-up)}~}
\newcommand{\trianglepassdown}[0]{\textit{Triangle Pass} \texttt{(palm-down)}~}
\newcommand{\trianglepassdownreversed}[0]{\textit{Triangle Pass} \texttt{(palm-down, reversed)}~}
\newcommand{\fourfingerpassleap}[0]{\texttt{Four-Finger-Pass (palm-down, LEAP)}~}
\newcommand{\fourfingerpass}[0]{\texttt{Four-Finger-Pass}~}

\newcommand{\rps}[0]{RoboPSer}
\newcommand{\passvzero}[0]{\textit{Pass}-V-R~}
\newcommand{\passvone}[0]{\textit{Pass}-V-NR~}
\newcommand{\passvtwo}[0]{\textit{Pass}-H-NR~}
\newcommand{\passvthree}[0]{\textit{Pass}-H-R~}

\newcommand{\passns}[0]{\textit{Pass}}
\newcommand{\passvzerons}[0]{\textit{Pass}-V-R}
\newcommand{\passvonens}[0]{\textit{Pass}-V-NR}
\newcommand{\passvtwons}[0]{\textit{Pass}-H-NR}
\newcommand{\passvthreens}[0]{\textit{Pass}-H-R}
\newcommand{\thumbaroundns}[0]{\textit{Thumb-Around}}

\newcommand{\sonicdownns}[0]{\textit{Sonic}\texttt{ (down)}}
\newcommand{\trianglepassns}[0]{\textit{Triangle Pass}}
\newcommand{\trianglepassupns}[0]{\textit{Triangle Pass} \texttt{(palm-up)}}
\newcommand{\trianglepassdownns}[0]{\textit{Triangle Pass} \texttt{(palm-down)}}
\newcommand{\trianglepassdownreversedns}[0]{\textit{Triangle Pass} \texttt{(palm-down, reversed)}}
\newcommand{\fourfingerpassleapns}[0]{\texttt{Four-Finger-Pass (palm-down, LEAP)}}
\newcommand{\fourfingerpassns}[0]{\texttt{Four-Finger-Pass}}

\newcommand{\LogU}{\operatorname{LogUniform}}

\makeatletter
\newcommand{\namedlabel}[2]{%
  \begingroup
  \def\@currentlabelname{#2}%
  \label{#1}%
  \endgroup
}
\makeatother

\newcommand{\compleedtodo}[1]{{}}

\def\scititle{%
    Reliable and Diverse Robotic Pen Spinning\\
    via Guided Model-to-Policy Refinement%
}

\projectname{\rps}                       
\projecticon{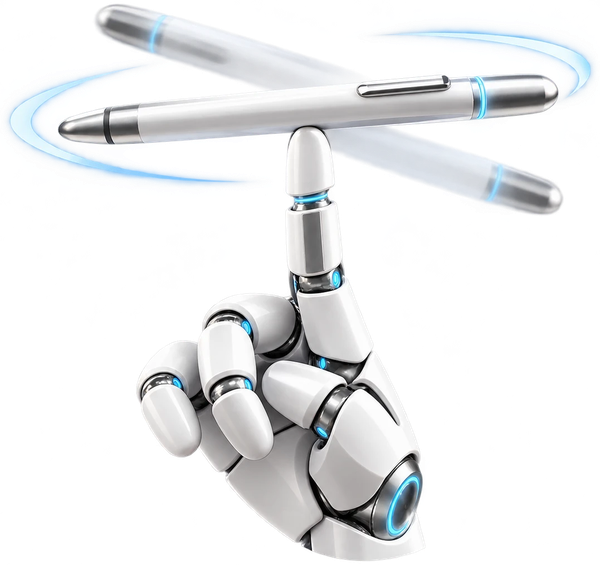}     
\title{\scititle}

\author{%
  Xueyi~Liu$^{1,2}$,
  Haozhe~Cui$^{3,2}$,
  Xizhe~Feng$^{1,2}$,
  He~Wang$^{4,2}$,
  Li~Yi$^{1,5,2}$%
}
\affiliations{%
  $^{1}$Tsinghua University\quad
  $^{2}$Galbot\quad
  $^{3}$ShanghaiTech University\quad
  $^{4}$Peking University\quad
  $^{5}$Shanghai Qi Zhi Institute%
}

\AtBeginDocument{\hypersetup{pdfauthor={Xueyi Liu, Haozhe Cui, Xizhe Feng, He Wang, Li Yi}}}

\begin{document}

\maketitle

\input{articles_c/abs}

\input{articles_c/intro}

\input{articles_c/result}

\input{articles_c/discussion}

\input{articles_c/method}

\input{articles_c/acknowledgments}


\clearpage
\bibliographystyle{meowarxiv}  
\bibliography{references}

\input{supp/supp}

\clearpage
\section*{Captions for Movies S1 to S16 and Data S1 to S3}
\input{supp/supp_caption}

\end{document}

%% file: articles_c/abs.tex
\begin{abstract} \bfseries \boldmath

    Simulation-trained policies can fail entirely on real robots when dexterous manipulation requires precisely timed forces and contact transitions, even after actuator, delay, and contact calibration.
    We present a three-stage framework that advances such policies from zero success to reliable execution across ten pen-spinning tasks spanning rapid motion, changing frictional contacts, and geometrically constrained finger gaiting. The framework uses real-world experience to improve both the training simulator and the deployed controller.
    Offline human guidance and coarse-physics-guided analysis identify targeted corrections, which we optimize through physical trials to establish initial success. 
    The resulting trajectories support learning a hybrid simulator with shared per-contact neural compensation, reducing contact-model mismatch and enabling policy refinement in simulation. The refined controller provides a stronger high-frequency feedback base, while its rollouts define a structured action representation for learning coordinated residual sequences through real-world reinforcement learning.
    This enables direct real-world optimization without further dynamics-model refinement. 
    Across six dynamic and four slower tasks on two hand embodiments, mean success rises from 0\% to 98\% over 50 trials per task (all $\geq90\%$, nine $\geq96\%$, six 100\%). 
    Sustained rotation exceeds 20 minutes on selected tasks and 30 minutes ($268\pi$ radians) beyond pen spinning. 
    Selected policies partially generalize zero-shot to new objects; 250--500 additional real-world rollouts yield 96--100\% success on seven variants and 16+ minutes of continuous rotation on an eighth.
    These results suggest a practical 
    sim-to-real path from zero success under direct transfer to reliable real-world execution of challenging, dynamic, contact-sensitive dexterous skills.
\end{abstract}


%% file: articles_c/intro.tex
\section*{Introduction}

\begin{figure}[htbp]
  \centering
  \includegraphics[width=0.98\textwidth,height=0.89\textheight,keepaspectratio]{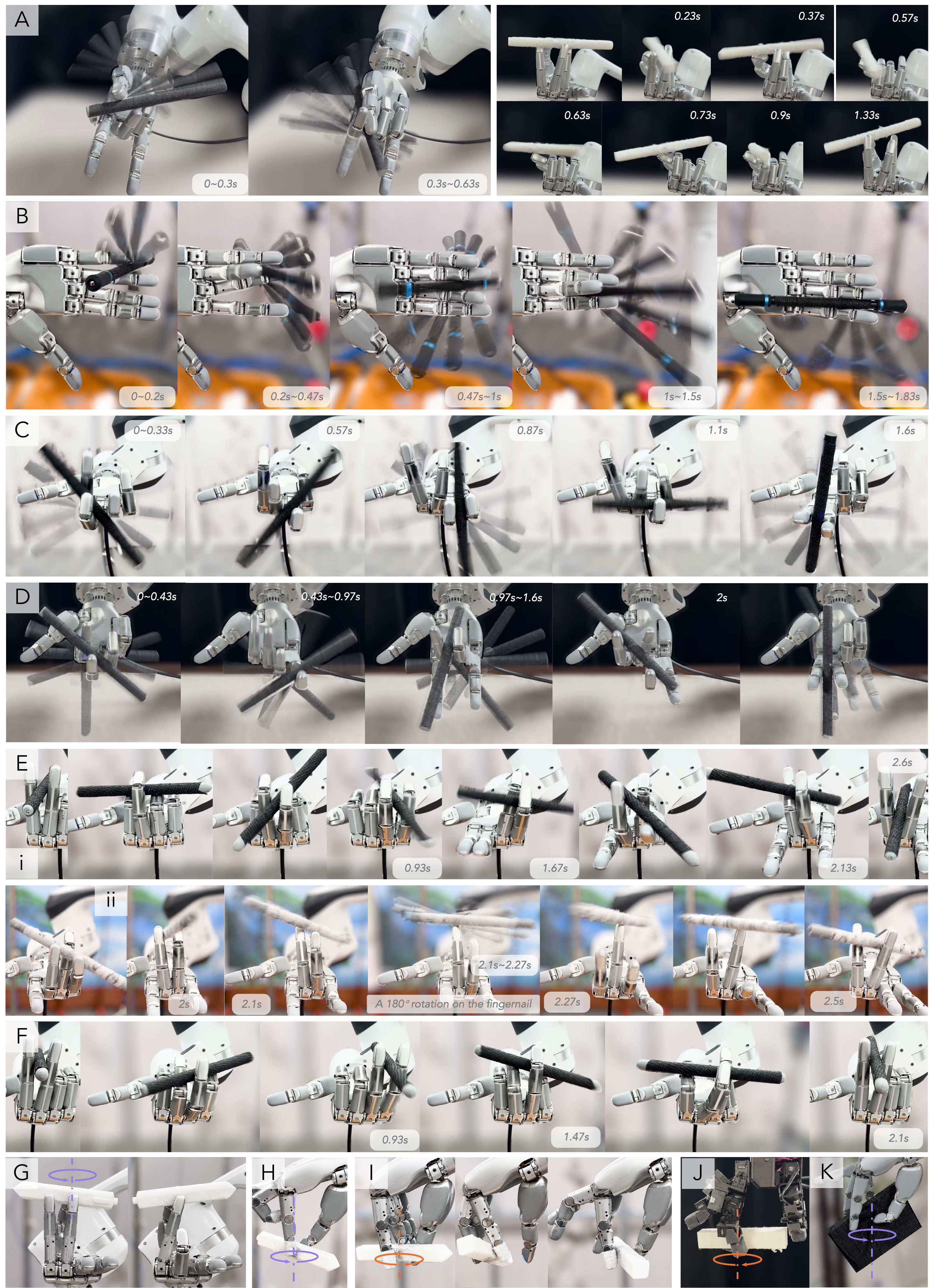}
  \caption{\footnotesize
  \textbf{Diverse Pen Spinning Skills.}  
  \textbf{(A)} \thumbaround ($2.4\pi$--$3.1\pi$ rad/s; left) and a variant (right). \textbf{(B)} \sonicdown ($1.5\pi$--$2.6\pi$ rad/s). \textbf{(C--F)} \pass variants with different initial pen poses and cycle reversal behavior ($1.48\pi$--$2.5\pi$ rad/s). 
  \textbf{(E-ii)} A half-turn spinning on the middle fingernail tip. 
  \textbf{(G--I)} Three \trianglepass variants with different palm orientations and rotation directions.
  \textbf{(J)} Validation on LEAP hand.
  \textbf{(K)} Beyond pen spinning: \fourfingerpassns.
  }
  \label{fig:teaser_task}
\end{figure}


Contact-rich dexterous manipulation requires fine-grained finger control and precise contact adjustments to move the object as desired while keeping it in hand.
Fingers must precisely prompt object motion, regulate momentum, and release and re-establish contacts while avoiding unwanted collisions and respecting geometric constraints.
Transferring policies for these tasks from simulation to real robots remains challenging because of sim-to-real modeling gaps~\cite{Wang2025PenSpin,Liu2025DexNDMCT,chen2023visualdexterity,handa2024dextremetransferagileinhand,feng2026minimalistretargetingguidedreinforcementlearning,hu2023dexterousinhandmanipulationslender,Singh2025InHandMO}. Rapid actuation and precisely timed finger gaiting amplify the effects of mismatches in contact and actuator dynamics~\cite{Wang2025PenSpin,Liu2025DexNDMCT,Fey2025BridgingTS}: small force or timing errors can turn successful simulated behavior into jammed fingers or dropped objects. 
Beyond initial success, reliable execution must tolerate trial-to-trial variations in hand--object configurations, contact interactions, and actuator response. 
For cyclic skills, higher reliability can also enable stable multi-episode execution despite accumulating errors.

Robotic pen spinning brings these demands together in a compact but demanding testbed. Towards this vision, prior studies have explored sim-real co-training~\cite{Wang2025PenSpin}, tracking~\cite{Wu2026TopoRetarget,Li2026TeleDexter}, and direct optimization of motion primitives or model-based control in the real world~\cite{Yao2025SWIFT,Nakatani2023PenSpinning,Ishihara2006DynamicPS}. Despite progress, they face one or more of the following constraints: low-speed rotations, limited validations on the real-world reliability, customized hardware, or constrained styles.
Pen spinning styles require varied combinations of rapid rotation, sliding contact, release and capture, and geometrically constrained finger gaiting, making reliable execution a demanding control problem. Multi-episode execution of cyclic variants further tests policy robustness to accumulating errors.
Here, we study ten pen-spinning tasks across two hand embodiments, nine on the SharpaWave hand (Sharpa, Singapore) and one on the LEAP hand~\cite{shaw2023leap}: six dynamic maneuvers and four slower tasks requiring delicate finger gaiting (Fig.~\ref{fig:teaser_task}). Together, these tasks provide structured coverage of diverse sim-to-real challenges (Fig.~\ref{fig:rollout_figure}). 
The ``pens'' are elongated objects sized for robotic manipulation, including prisms as well as cylinders or capsules with radii of $0.54$--$0.7\,\mathrm{cm}$ and object-to-hand width ratios on Sharpa comparable to those in human pen spinning.
Directly transferred policies achieve no successful trials on any task in our evaluation. Our central goal is to transform these unsuccessful policies into skills that are repeatable across trials and, where the task permits, sustainable over continued execution.

Existing approaches offer two complementary routes toward this goal: improving the dynamics used for policy training and improving control directly through real-world interaction. System identification calibrates simulator parameters~\cite{An1985EstimationOI,Mayeda1988BasePO,Lee2023RobotMI,Sobanbabu2025SamplingBasedSI,Allevato2019TuneNetOR}, including through learned state–action-dependent mappings~\cite{Jiang2021SimGANHS}, to support policy refinement.
Domain randomization broadens training dynamics to promote robustness~\cite{Loquercio2019DeepDR,Peng2017SimtoRealTO,Tan2018SimtoRealLA,Yu2019SimtoRealTF,Mozifian2019LearningDR,Siekmann2020SimtoRealLO,Jakobi1995NoiseAT}. Neural dynamics models learn effects that analytical formulations may miss~\cite{Hwangbo2019LearningAA,He2025ASAPAS,Fey2025BridgingTS,Shi2024AnEM,Liu2025DexNDMCT}.
Beyond accurately modeling local contact dynamics, learned models should generalize to new behaviors introduced by policy updates. Joint-wise formulations can help with generalization~\cite{Liu2025DexNDMCT}, but capture contact dynamics indirectly through joint responses. 
Analytical--neural hybrids combine analytical simulation components with learned contact representations, force augmentations, or robot dynamics models~\cite{Heiden2020NeuralSimAD,Xu2025NeuralRD,Pfrommer2020ContactNetsLO,Ataei2026FlashNeRDPC}.
This physical inductive bias can support generalization, but balancing model expressivity and generalization for effective policy refinement depends on what is learned and at what granularity.
Iterative data collection and model updates can expand coverage~\cite{Shi2024AnEM,Liu2025DexNDMCT,Farchy2013HumanoidRL}, but may require substantial real-world data.
The central challenge for approaches based on learned dynamics is not simply to fit recorded trajectories, but to learn precise and generalizable dynamics that support effective policy improvement with limited real-world data.


Direct real-world reinforcement learning (RL) improves policies through interaction with the physical system, making efficient exploration a central challenge.
Human demonstrations and interventions can guide learning~\cite{luo2024hilserl,Han2026DexHiLAH,jiang2024transic}, but embodiment differences and limited feedback complicate precisely timed, force-sensitive corrections~\cite{Wang2025PenSpin,Yin2025DexterityGenFC,Li2026TeleDexter}. Controller guidance supplies prior trajectories and online assistance~\cite{Kktabak2026RealWorldRL}, yet uncoordinated joint exploration can still disrupt critical contacts. 
Model-based fine-tuning can combine simulation-trained value functions with dynamics models learned from real-world data to guide short-horizon exploration~\cite{Yin2025RapidlyAP}, but relies on reliable local predictions and informative value transfer.
Generative policies and latent action spaces support coordinated exploration by generating actions through learned representations, often using action chunking for temporal structure or latent compression for efficient exploration~\cite{Wagenmaker2025SteeringYD,Yang2026SERNFSR,Zhou2020PLASLA,Allshire2021LASERLA,Yu2026BeyondAR,Zhu2026TowardsHD}. 
Two design choices matter for demanding tasks:
open-loop chunk execution can reduce feedback responsiveness, while action compression and reconstruction errors can compromise fine-grained control and constrain corrective exploration. 
Adding per-step residuals to pretrained policy actions supports fine-grained corrections~\cite{Su2026RFS,Ankile2025ResidualOR}, but exploration without an explicit spatiotemporal action prior may disrupt critical contacts and limit policy improvement.
The key challenge is enabling sample-efficient, coordinated policy improvement while preserving feedback responsiveness and precise commands, without requiring complete actions to be generated through a fixed learned representation.
Moreover, effective fine-tuning often starts from an already capable policy~\cite{rl100,Yu2026BeyondAR,Sun2026FromPT,Zhu2026TowardsHD}.
When direct transfer yields zero success, both dynamics learning and direct policy learning face a difficult starting point. Failed rollouts may never reach later task transitions, limiting informative data for dynamics learning, while physical exploration must discover coordinated corrections before successful behavior is available to build upon.

We introduce a unified sim-to-real framework based on guided model-to-policy refinement, with three closely coupled stages that improve both the simulator and controller within practical interaction budgets (Fig.~\ref{fig:framework_stages}).
Human-guided corrective optimization first establishes successful execution and collects trajectories spanning essential task transitions, enabling hybrid simulator learning and subsequent policy refinement in simulation. 
Without this corrective stage, direct policy transfer typically produces failed rollouts that offer limited support for refinement (Fig.~\ref{fig:framework_stages}(B)).
With these informative trajectories, model-based refinement improves performance, but its gains diminish over time, possibly because limited real-world data constrain the learning of accurate, generalizable contact dynamics. 
To overcome this limitation, we transition to model-free real-world RL, which directly improves reliability through structured residual corrections while preserving rapid feedback and fine-grained commands. Despite its diminishing returns, model-based refinement remains a valuable intermediate stage: compared with the first-stage policy, it provides a stronger initialization for real-world RL, reducing exploration costs and improving final performance. Omitting this stage yields lower reliability than the full framework, even with more than twice as much real-world data.
We call the overall system RoboPSer, after ``PSer,'' the pen-spinning community's shorthand for a pen spinner.

To establish initial success, we separate offline human guidance from dynamically precise execution. Our key idea is coarse-to-fine corrective optimization in both physics fidelity and correction granularity. Humans inspect recorded trajectories to identify failures and specify desired object poses when needed. Human-guided kinematic correction converts intuitive rigid-body edits into candidate joints and motion directions to remove obstructing contacts or establish new ones. Coarse-physics-guided contact correction uses reasoning under a coarse contact model to propose contact operations and corresponding joints and directions that could drive the object toward the specified pose. 
These proposals define compact search spaces for low-dimensional corrective primitives, whose parameters we optimize through real-world trials via the covariance matrix adaptation evolution strategy (CMA-ES)~\cite{Hansen2001CompletelyDS}. 
Real-world physics thus determines the precise timing and force-sensitive adjustments that coarse reasoning cannot supply. Finally, we distill the resulting corrections into an observation-conditioned policy that modulates the base controller.

For model-based refinement, we learn local contact corrections while retaining the simulator's analytical physics loop. The central challenge is balancing expressivity and generalization with limited real-world data. 
Our key design choice is to 
embed a shared per-contact neural compensator within MuJoCo Warp~\cite{mujocowarp2025}, which models contact through convex soft constraints. 
Using local contact features, it modulates key parameters of these constraints---reference accelerations and constraint regularization---before the coupled solve. 
This retains the analytical solver and its feasibility constraints while enabling expressive corrections, with shared parameters and local inputs encouraging cross-contact event generalization.
We further use joint-wise delta-action models at the simulator's action interface to compensate for residual contact-model and other simulator errors~\cite{Liu2025DexNDMCT,Farchy2013HumanoidRL,He2025ASAPAS,Fey2025BridgingTS,Hanna2017GroundedAT,Karnan2020ReinforcedGA}.
Together, these corrections support further policy refinement in the hybrid simulator.

Our final stage separates the temporal structure of exploration from the feedback frequency of execution. Using rollouts after model-based refinement, we learn a structured latent action space for task-relevant exploration~\cite{Zhu2026TowardsHD,Yu2026BeyondAR,Yang2026LAMPLM}.
With the base policy and action representation fixed, real-world RL trains a latent modulator to perturb observation-conditioned reference latents.
The difference between action chunks decoded from the perturbed and reference latents forms a residual action sequence.
At every control step, we add the corresponding residual to the base-policy command computed from the latest observation.
This supports coordinated exploration across joints and time while preserving high-frequency feedback and fine-grained base-policy commands.
The decoder thus parameterizes corrections rather than replacing the base policy as the source of full control commands.
Our experiments demonstrate the importance of these properties for dynamic, contact-sensitive manipulation.

Across ten tasks, \rps{}
raises mean success from 0\% under direct transfer to  98\% with six reaching 100\% and all $\geq$90\% (Fig.~\ref{fig:rollout_figure}(C)).
This reliability supports multi-episode execution: mean accumulated rotation before failure reaches $79.8\pi$, $80.2\pi$, and $21.4\pi$ radians on three cyclic tasks, with the longest runs on two exceeding 20 minutes ($182\pi$ and $154\pi$ radians). 
Beyond pen-like objects, four-finger rotation of a rectangular board ran uninterrupted for over 30~min, totaling $268\pi$~rad.
Selected policies also partially generalize zero-shot to changes in object geometry, compliance, and contact friction. Adaptation with 250–500 additional real-world rollouts per variant raises success to 96–100\% across seven variants and enables over 16 minutes of uninterrupted rotation on an additional long-board variant.
Ablations validate the key designs and the roles of the three stages.


Our contributions are:
\begin{itemize}
    \item \textbf{Reliable Dynamic Dexterity.} Our sim-to-real framework raises mean success from 0\% under direct transfer to 98\% on ten pen-spinning tasks across two hand embodiments.
    \item \textbf{Human-Guided Corrective Optimization.} Offline human guidance and coarse contact reasoning define compact search spaces for real-world primitive optimization, enabling precisely timed, force-sensitive corrections  to critical transfer failures.
    \item \textbf{Hybrid Simulator for Sim-to-Real Policy Refinement.} A shared per-contact compensator learned from real-world data modulates contact-constraint parameters before the coupled solve, capturing task-relevant contact dynamics within the analytical solver's feasibility constraints and supporting generalization.
    \item \textbf{Structured Residual Real-World RL.} Coordinated residual sequences from a learned action representation correct, rather than reconstruct, per-step base-policy commands, enabling structured, sample-efficient exploration while preserving high-frequency feedback and fine-grained control.
\end{itemize}

%% file: articles_c/result.tex
\section*{Results} \label{sec:results}

\begin{figure}[htbp]
  \centering
  \includegraphics[width=0.98\textwidth,height=0.84\textheight,keepaspectratio]{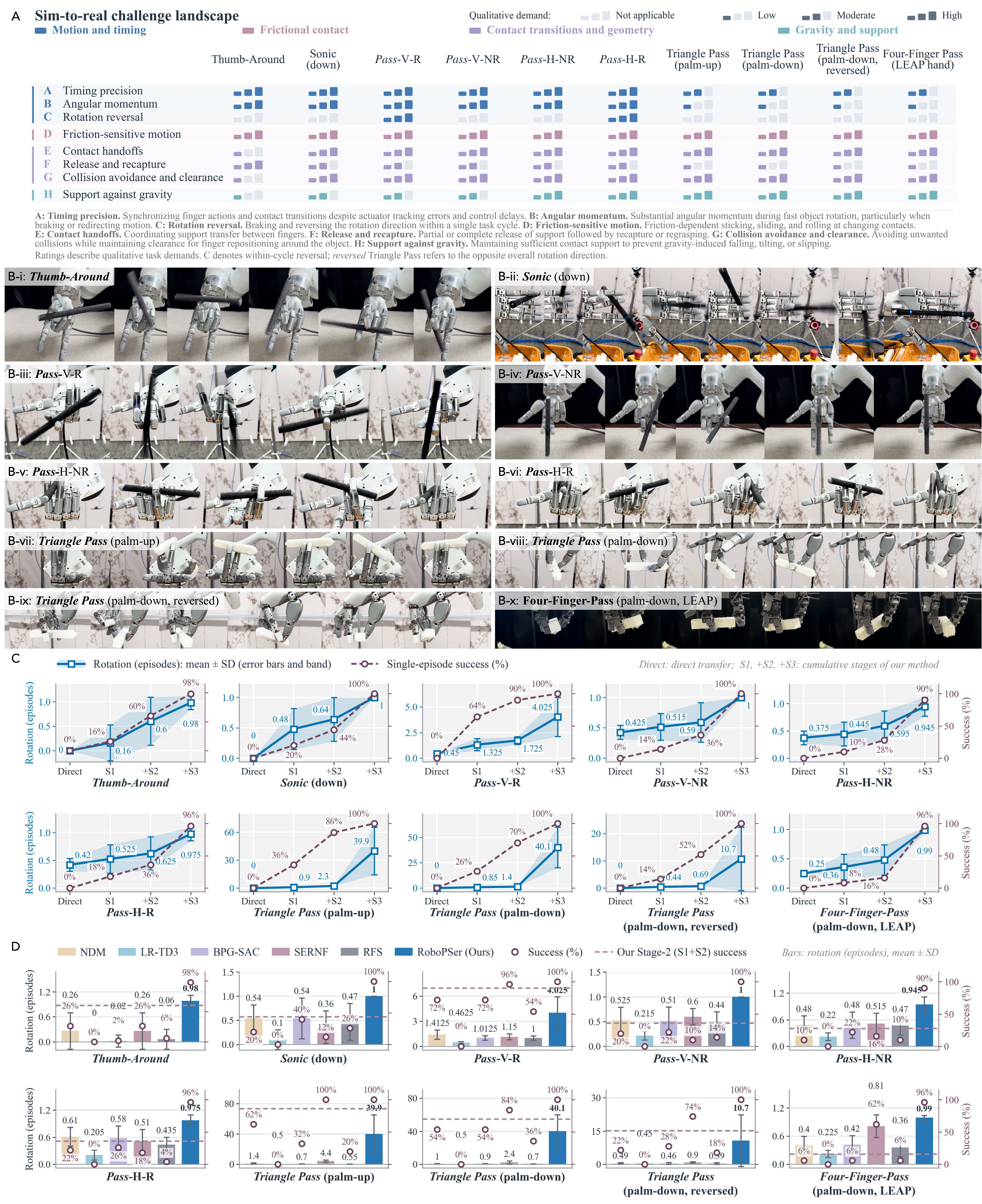}
\caption{\footnotesize
    \textbf{Sim-to-Real Challenges and Real-World Evaluation of Ten Pen-Spinning Tasks.}
    \textbf{(A)} Columns denote tasks; rows show eight challenge subcategories grouped into four color-coded categories. Definitions appear beneath the matrix.
    \textbf{(B)} Panels (i)--(x) show each task's rollout snapshots.
    \textbf{(C)} Performance across stages for each task. 
    \textbf{(D)} Comparison with baselines for each task. 
    Success rates are measured over 50 evaluation trials per task.
    An episode denotes a complete finger-slot cycle returning to the starting slot (\passns), a $2\pi$ revolution (\trianglepass and \fourfingerpassns), or one complete spinning maneuver (\thumbaround and \sonicdownns).
    Rotation (episodes) is summarized as the mean $\pm$ population standard deviation (SD) across evaluation trials (5--50 per task; Supplementary Section~S2); error bars and shaded bands indicate SD.
    }
  \label{fig:rollout_figure}
\end{figure}

\begin{figure}[htbp]
  \centering
  \includegraphics[width=1.0\textwidth,height=0.81\textheight,keepaspectratio]{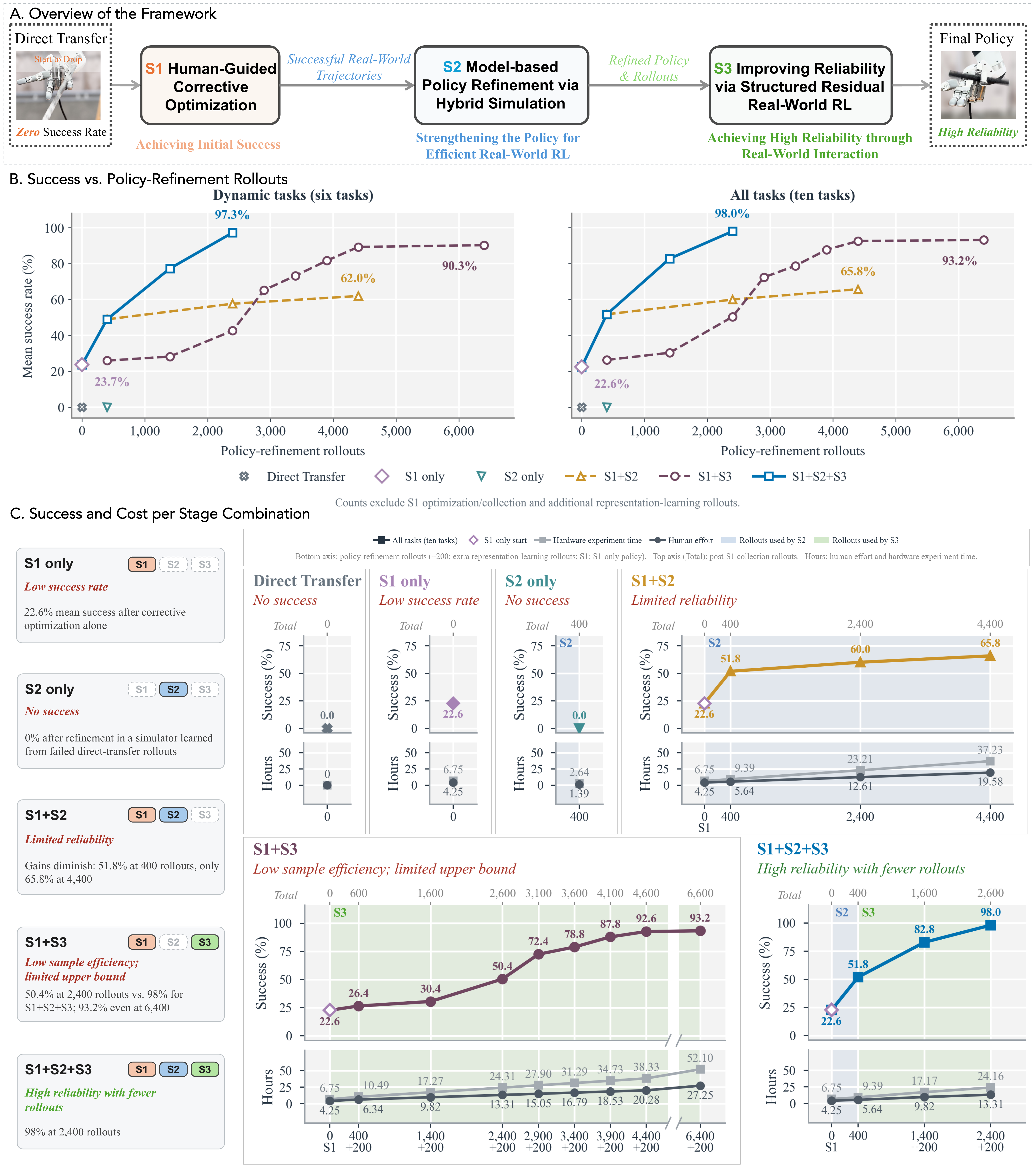}
\caption{\footnotesize
  \textbf{Framework Overview and Cross-Stage Interdependence.}
  \textbf{(A)} Overview of guided model-to-policy refinement, from initial success to high reliability. S1--S3 denote corrective optimization, model-based refinement, and model-free real-world RL.
  \textbf{(B)} Mean success rate versus policy-refinement rollouts per task. The S1+S2+S3 curve includes its S1 and S1+S2 checkpoints (0 and 400 rollouts).
  \textbf{(C)} Success rate, human effort, and hardware experiment time for each setting. Top axes give total post-S1 collection rollouts, which include 200 representation-learning rollouts for settings with S3.
  Rollout counts exclude S1 optimization trials and correction-policy collection. Human effort covers rollout collection and resets, plus corrective optimization and the full correction-policy data-collection budget for settings with S1; hardware time also includes cooldowns. Time estimates are approximate upper bounds excluding computation (Supplementary Section~S2). Success is evaluated over 50 trials per task; values are averaged across tasks.
}
\label{fig:framework_stages}
\end{figure}

We evaluate \nntask in-hand pen-spinning tasks---six dynamic tasks at $1.48\pi$--$3.1\pi\,\mathrm{rad/s}$ and four slower tasks---spanning rapid object motion, frictional contact-mode transitions, precise finger gaiting under geometric constraints, and balance under varying gravitational torques and support conditions, plus rotation of a large board (Figs.~\ref{fig:teaser_task} to~\ref{fig:framework_stages}; movie S1). At deployment, all policies use hand proprioception alone~\cite{Qi2022InHandOR,Liu2025DexNDMCT,Wang2025PenSpin}, without tactile sensing or external object-state observations.


Direct transfer fails on every task, even after calibration. We identify effective actuator parameters and action/observation delays, learn a free-hand action compensator, and identify contact parameters for each task from open-loop replays of real-world commands. Policies trained in this calibrated simulator, using proprioception alone or with object-state histories, achieve 0/50 success on every task, as do Isaac Lab policies on the two tasks tested. The compensator lowers free-hand joint error by $17.4\%$, but on 20 \passvzero hand--object replays it barely changes joint error ($0.8\%$) and does not improve object-orientation tracking, motivating explicit contact modeling in the model-based refinement stage.
Concurrent pen-spinning systems that transfer from simulation~\cite{Wu2026TopoRetarget,Li2026TeleDexter} use more forgiving contact conditions in their public demonstrations, such as gloves or extensive tape covering the fingertips and the finger links below them. See details in Supplementary Sections~S2 and~S3.  

We measure reliability in two ways. 
\emph{Single-episode success rate} is the fraction of successful trials from initial states within the policy's intended operating range; because visually subtle differences in these states change contact locations and forces, repeated trials of the same deterministic policy test reproducibility (Supplementary Section~S3).
\emph{Average episodes}, the total rotation in task-specific episodes divided by the number of trials, measures sustained execution despite accumulating errors. 
\rps{}
achieves at least 90\% success on every task (98\% mean) and sustains multi-episode rotation, including an uninterrupted run of over 30 minutes ($268\pi$ radians). Movies S2--S12 show all single-episode evaluation trials and the continuous runs.
We further evaluate zero-shot generalization, adaptation, and robustness to human disturbances (Fig.~\ref{fig:few_shot_zero_shot_tasks}, movies S13--S15), and ablate each stage and key design (Figs.~\ref{fig:human_takeover_failures} to~\ref{fig:structural_and_residual_exploration}).

\subsection*{Dynamic Pen Spinning}
Six dynamic tasks from three human pen-spinning families---\textit{Thumb-Around}, \textit{Sonic}, and \textit{Pass}---require rapid motion, precisely timed finger gaiting, and control through changing frictional contacts and contact modes on a non-customized, fully actuated hand. Over 50 trials per task, our method raises success from 0\% under direct transfer to 90--100\%, with five tasks reaching at least 96\% (movies S2--S7).

\paragraph{\thumbaround spinning style.}
The pen rotates around the thumb through sliding contact and aerial rotation without a stabilizing pinch grasp, requiring a controlled launch and a precisely timed catch. Our method achieves 98\% success (49/50 trials) at $2.4\pi$--$3.1\pi\,\mathrm{rad/s}$. A \textit{Thumb-Point-Around} variant (Fig.~\ref{fig:teaser_task}(A, right)) was discontinued because its forceful actions destabilized the hardware.

\paragraph{\sonic spinning style.}
\sonicdown couples two successive finger-slot transfers with a conical pen-axis sweep, requiring precise force regulation and timely release and re-establishment of contact. Our method achieves 100\% success (50/50 trials) at $1.5\pi$--$2.6\pi\,\mathrm{rad/s}$, as does a variant with more \textit{Charge}-style rotations between transfers.

\paragraph{\pass spinning family.}
The four \textit{Pass} variants combine direction reversal, dorsal finger-slot transfers, and gravity compensation. Names indicate the initial pen orientation (V: vertical; H: horizontal) and within-episode reversal (R: reversal; NR: no reversal). \passvzerons, with repeated transfers and a reversal in every episode, achieves 100\% success at $2\pi$--$2.5\pi\,\mathrm{rad/s}$ and averages 4.025 episodes over 20 continuous trials (up to 10.5). \passvonens, whose dorsal return to the initial slot without reversal makes contact maintenance harder, achieves 100\% at $1.5\pi$--$1.76\pi\,\mathrm{rad/s}$ (up to four episodes). \passvtwons, which must also prevent gravity-induced tilting during the transfer, achieves 90\% at $1.48\pi$--$1.54\pi\,\mathrm{rad/s}$ (up to two episodes), and \passvthreens, which reverses direction with tilt prevention throughout the cycle, achieves 96\% at $1.875\pi$--$2\pi\,\mathrm{rad/s}$ (up to three episodes).

\paragraph{Emergent contact transitions.}
Real-world training also discovers alternative transitions: on \passvtwons, the pen spins about $180^\circ$ around the middle fingernail before entering the target slot (Fig.~\ref{fig:teaser_task}(E-ii)).

\subsection*{Pen Rotation with Delicate Finger Gaits}
Three slower \trianglepass variants require precise finger gaiting under different gravitational, geometric, and contact constraints. Policies for the original, human-inspired dynamic style failed within one-third of a revolution even in simulation, motivating slower variants with controlled fingertip grasping and regrasping that must prevent drops, axial drift, and obstructive contacts. Our method raises full-episode success from 0\% under direct transfer to 100\% on all three variants (50/50 trials each; one episode spans about $2\pi$ radians). Continuous runs are limited to five per task because of their length and actuator heating after 20--30 minutes (movies S8--S10).

\paragraph{\trianglepassupns.}
The fingers must rotate the object while supporting it against gravity, maintaining its height, and regulating friction to prevent axial drift. The policy sustains 39.9 episodes on average across five evaluation trials. A separate demonstration reaches 91 consecutive episodes.

\paragraph{\trianglepassdownns.}
The fingers must provide upward friction against dropping without displacing the object palmward and obstructing rotation. The policy sustains 40.1 episodes on average (up to 77).

\paragraph{\trianglepassdownreversedns.}
Reversed rotation requires a dynamic transfer: the thumb retracts to avoid collision while the other fingers support and rotate the object, then precisely re-establishes contact. The policy sustains 10.7 episodes on average (up to 29).

\subsection*{Validation on LEAP Hand}
We assess our framework's applicability to LEAP Hand by training a LEAP-specific policy for \texttt{Four-Finger-Pass}, a task adapted from DexNDM~\cite{Liu2025DexNDMCT} that requires coordinated four-finger transfers of a $3\,\mathrm{cm}$-wide, $16\,\mathrm{cm}$-long object. Our method raises full-episode success from 0\% under direct transfer to 96\% (48/50 trials), sustaining up to two consecutive episodes (movie S11). We did not evaluate the dynamic tasks, which rely on dorsal finger contacts where LEAP's servomotors sit.

\subsection*{Beyond Pen Spinning}
On the Sharpa hand, \fourfingerpass rotates a $15\times10\times2\,\mathrm{cm}$ board, including a rapid finger-gaiting phase that reorients it by about $90^\circ$. Direct transfer fails (0\% full-episode success). Our policy completes all 50 assessed episodes within one uninterrupted run that lasts over 30 minutes and totals 134 consecutive episodes ($268\pi$ radians) without human intervention or resets (movie S12).

\subsection*{Sample Efficiency}
Figure~\ref{fig:framework_stages}(B, C) compares the full framework with S2 only, S1+S2, and S1+S3 across policy-refinement budgets, capped at 4,400 rollouts per task to limit hardware wear and dynamics drift; we extend S1+S3 to 6,400 to test whether more interaction closes the gap. Settings with S3 additionally collect 200 representation-learning rollouts.

With 400 rollouts per task, S2 alone remains at 0\% mean success, whereas S2 after S1 raises success from 22.6\% to 51.8\%, suggesting that S1's successes cover the task transitions needed for dynamics learning.
Further model-based refinement yields diminishing gains (60.0\% and 65.8\% at 2,400 and 4,400 rollouts).

At 2,400 rollouts, the full framework reaches 98\% mean success versus 50.4\% for S1+S3, which reaches only 92.6\% and 93.2\% at 4,400 and 6,400 rollouts, possibly because its weaker starting policy needs larger or qualitatively different corrections than the residual parameterization readily provides. Model-based refinement before real-world RL thus achieves higher success at lower estimated human effort and hardware time (Fig.~\ref{fig:framework_stages}(C); Supplementary Section~S2).

\subsection*{Comparison with Baselines}

Figure~\ref{fig:rollout_figure}(D) compares \rps{}
with NDM, a joint-wise dynamics refinement following DexNDM~\cite{Liu2025DexNDMCT}, and four real-world RL baselines: Latent-Residual TD3 (LR-TD3)~\cite{Zhu2026TowardsHD}, Base-Policy-Guided Soft Actor-Critic (BPG-SAC)~\cite{Kktabak2026RealWorldRL}, SERNF~\cite{Yang2026SERNFSR}, and RFS~\cite{Su2026RFS}.
NDM starts from our first-stage policy and the RL baselines from our second-stage policy, with matched simulation pretraining and hardware rollout budgets. BPG-SAC and RFS additionally receive simulation-only conservative tuning. Otherwise, we follow each paper's settings.

All five baselines underperform our method overall.
NDM achieves lower average rotation than our second-stage policy on all ten tasks, and NDM-style joint-wise compensation alone also trails contact compensation on most tasks (Fig.~\ref{fig:model_based_abl}, ``w/ delta-action model''). 
The two chunk-based methods start from chunked policies weaker than our per-step second-stage controller: LR-TD3's reference flow drifts on hardware and ends at 0\% success after fine-tuning, and SERNF loses performance during action-chunk distillation on some tasks and stays below the second-stage controller on several dynamic tasks
after fine-tuning, consistent with the sensitivity to action chunking and learned action interfaces in Fig.~\ref{fig:structural_and_residual_exploration}. 
The two per-step joint-space methods, BPG-SAC and RFS, fall below the second-stage controller on all ten tasks despite conservative tuning, suggesting that uncoordinated exploration disrupts critical contacts; RFS must also collect its offline data by policy exploration because human takeover is ineffective in our setup (Fig.~\ref{fig:human_takeover_failures}). 
See detailed settings in Supplementary Section~S2 and per-baseline analyses in~S3. 


\subsection*{Generalization and Adaptation across Object Variants}
\label{sec:generalization_main}
\label{sec:adaptation_main}

\begin{figure}[!htbp]
  \centering
  \includegraphics[width=0.83\textwidth,height=0.86\textheight,keepaspectratio]{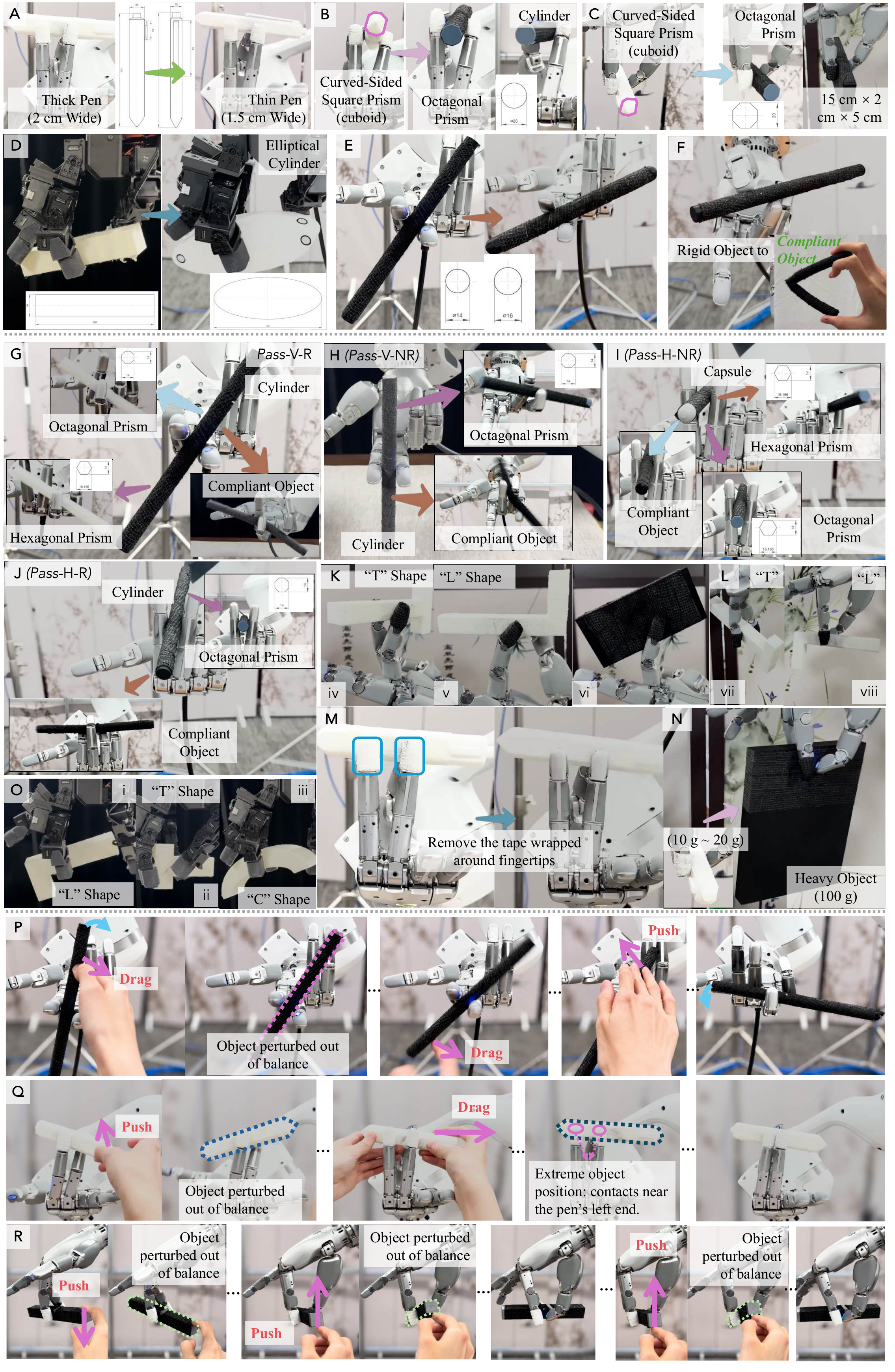}
  \par\vspace{2pt}
  \begin{minipage}{0.9\textwidth}
    \scriptsize
    \raggedright
    Tasks:
    \trianglepassup (A, B, K, M, Q);
    \trianglepassdown (C, L, N, R);
    \fourfingerpassleap (D, O);
    \passvzero (E, G, P);
    \thumbaround (F);
    \passvone (H);
    \passvtwo (I);
    \passvthree (J).
  \end{minipage}
  \caption{\footnotesize
  \textbf{Generalization and Adaptation across Task Variants.}
  (A--F) Adaptation to objects with reduced width (A), increased radius (E), novel geometries (B--D), or new materials (F), using 250--500 additional real-world rollouts per variant.
  (G--O) Zero-shot generalization to novel geometries (G, H, I, J, K, L, O), materials (G, H, I, J), low friction (M), and increased mass (N; 100 g versus 10–20 g for the objects used in real-world training).
  (P--R) Sustained multi-episode rotations under frequent human disturbances.
  }
  \label{fig:few_shot_zero_shot_tasks}
\end{figure}

Without additional training, policies for five tasks were tested on ten altered-object conditions with 10--50 trials each (Fig.~\ref{fig:few_shot_zero_shot_tasks}(G--O), movie S13). Five conditions kept at least 80\% success, including \passvzero on octagonal prisms (10/10), \trianglepassup on a lower-friction object (47/50), and \passvzero on compliant thermoplastic polyurethane (TPU) cylinders (18/20); the other five fell to 50--70\% (median 75\% overall). In continuous runs, \passvzero completed up to five consecutive episodes on altered objects. 
Further demonstrations without repeated-trial evaluation include L-, C-, and T-shaped objects, a board (\trianglepassupns), and a $100\,\mathrm{g}$ book-like object.

In more difficult transfer settings, where direct generalization success ranged from below 10\% to about 50\%, 
adaptation with 250--500 additional real-world rollouts per variant raises success to 100\% on six variants and 96\% on the seventh (Fig.~\ref{fig:few_shot_zero_shot_tasks}(A--F), movie S14), including a wider ($1.6\,\mathrm{cm}$) cylinder for \passvzerons, compliant TPU objects for \thumbaroundns, and elliptical cylinders for \fourfingerpassleapns. 
On an additional $15\times2\times5\,\mathrm{cm}$ board, where the zero-shot transfer completed at most one episode, the adapted \trianglepassdown policy rotated for over 16 minutes (more than 45 consecutive episodes) without a reset.
Experimental details are deferred to Supplementary Section~S2.


\subsection*{Robustness to Human Disturbances}
\label{sec:human_disturbances_main}
Human-applied pushes and drags of unmeasured magnitude
perturb object pose and contact locations, which can interrupt rapid contact transitions in \passvzero and destabilize fingertip support in the \trianglepass variants (Fig.~\ref{fig:few_shot_zero_shot_tasks}(P--R), movie S15). With disturbances applied in every trial, closed-loop policies complete all trials: single episodes on \trianglepassup (50/50) and \trianglepassdown (10/10), and two consecutive episodes on \passvzero (10/10); separate sustained runs reach up to 3.5 episodes under disturbance. Open-loop replay of successful action sequences completes only 0--16\%, underscoring the role of closed-loop responses (Supplementary Section~S2).

\subsection*{Enabling Effective Human-Guided Corrective Optimization}  \namedlabel{sec:corrective_analysis}{Analysis on Correction Strategy}

\begin{figure}[!htbp]
  \centering
  \includegraphics[width=1.0\textwidth,height=0.88\textheight,keepaspectratio]{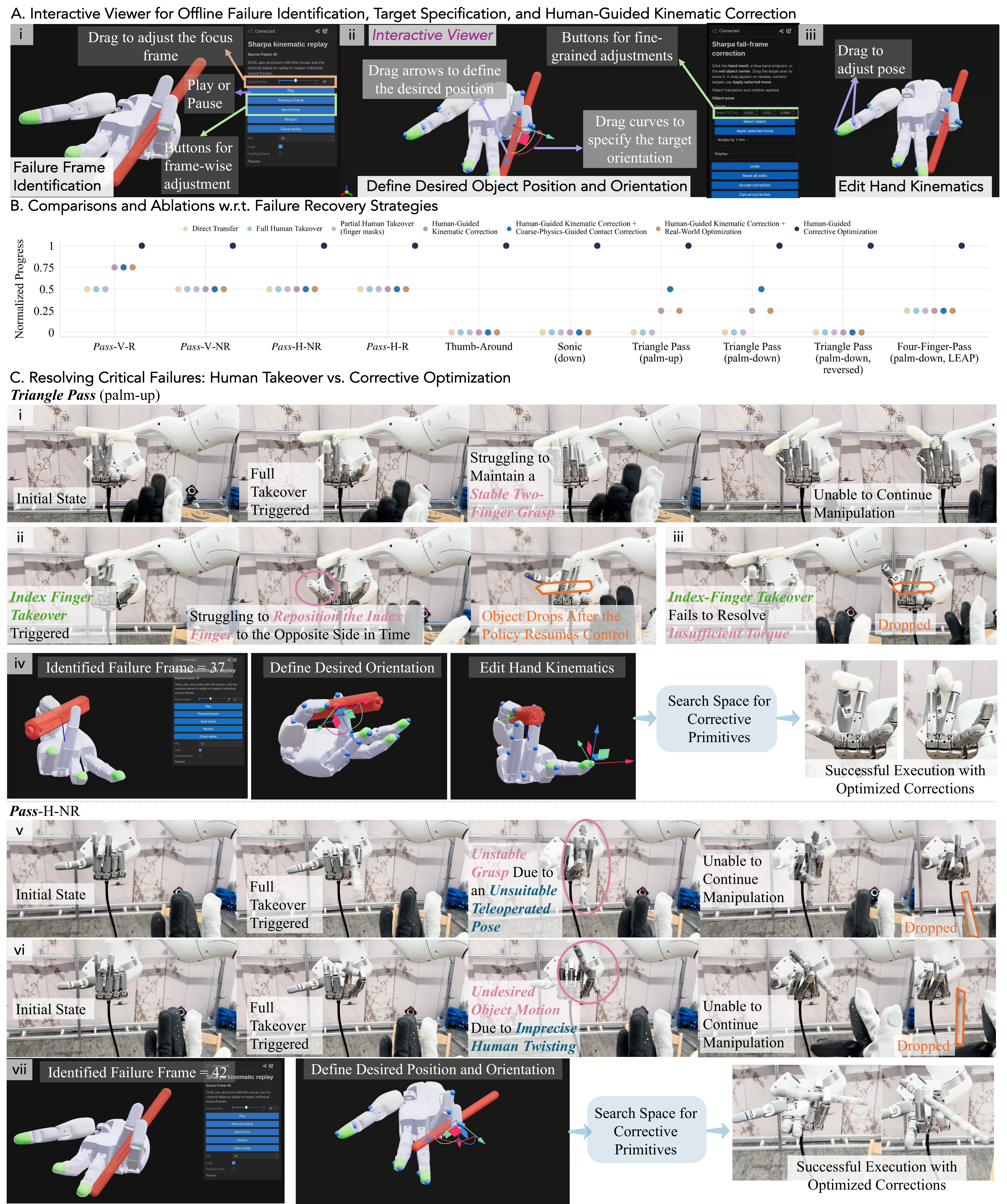}
\caption{\footnotesize
    \textbf{Human-Guided Corrective Optimization: Comparisons and Ablations.}
    \textbf{(A)} Offline viewer for failure identification (i), target-pose specification (ii), and kinematic correction (iii).
    \textbf{(B)} Maximum normalized single-episode progress by method and task; $1$ denotes completion.
    \textbf{(C)} Recovery at the first failure in \trianglepassup (i--iv) and \passvtwo (v--vii). Full-hand takeover disrupts fingertip support (i) or the grasp (v), or induces unwanted motion through imprecise forces (vi). Partial takeover fails to reposition the index finger within $\sim1\,\mathrm{s}$ (ii) or generate sufficient torque (iii). Optimized corrections succeed (iv,vii).
}
  \label{fig:human_takeover_failures}
\end{figure}

We compare our first stage against human takeover and three ablations (Fig.~\ref{fig:human_takeover_failures}).

\paragraph{Human takeover.}
\label{para:human_takeover}
After practice, neither full-hand nor finger-masked partial takeover achieves effective recovery in our setup, with up to 240 attempts per operator per failure case and early stopping when operators judged further improvement unlikely (B,C; movie S16). Full takeover often disrupts supporting contacts; partial takeover struggles because corrections must be precisely timed, force-sensitive, and compatible with the policy-controlled fingers within short intervention windows, even during slow rotation. Only the full method, combining contact-guided candidates with primitive optimization, completes every task (B).

\paragraph{Search-space proposals and primitive optimization.}
With fixed triangular joint-offset pulses, kinematic candidates extend progress on three tasks (\passvzerons, \trianglepassupns, and \trianglepassdownns), and coarse-physics-guided contact candidates add further gains on the two \trianglepass variants, but neither completes any task (B). 
Optimizing Gaussian primitives over kinematic candidates alone does not improve maximum single-episode progress over fixed kinematic pulses, suggesting that kinematic proposals may miss necessary contact-force adjustments. These results support combining contact-guided search spaces with optimization of timing, magnitude, and temporal structure.

\subsection*{Hybrid Simulation for Model-Based Refinement} 
\namedlabel{sec:model_based_analysis}{Analysis on Model-based Designs}



\begin{figure}[htbp]
  \centering
  \includegraphics[width=1.0\textwidth,height=0.86\textheight,keepaspectratio]{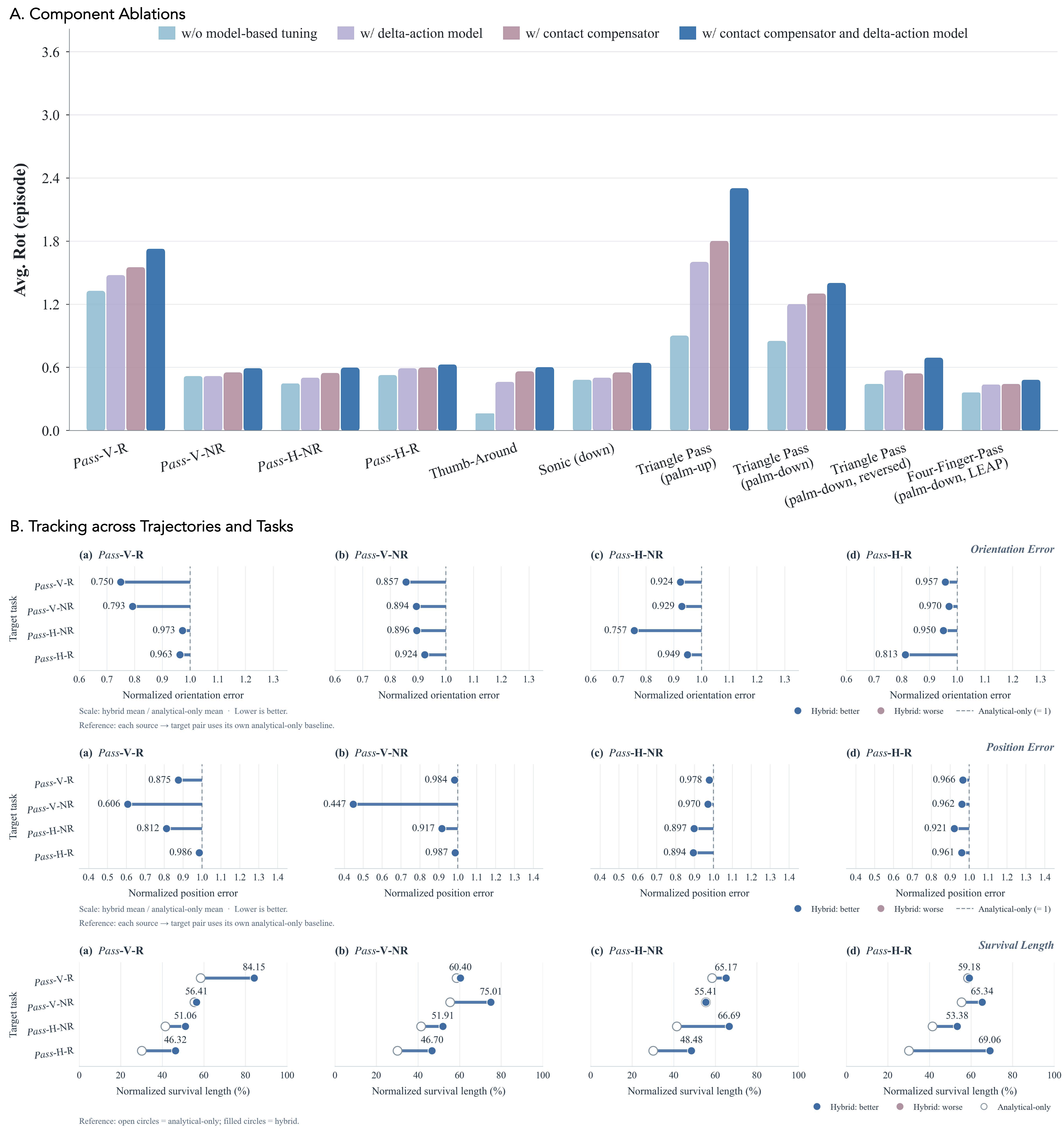}
    \caption{\footnotesize
        \textbf{Hybrid Simulator Components Ablations and Tracking Analysis.}
        \textbf{(A)} Ablations of contact and delta-action compensation, evaluated by real-world accumulated rotation. Bars show mean accumulated rotation over the evaluation rollouts specified in Supplementary Section~S2 (5--50 per task).
        \textbf{(B)} Cross-task tracking among four \pass variants: orientation error, position error, and survival length (top to bottom). Subpanels denote source tasks; rows denote targets (20 held-out trajectories per task). 
        Errors are relative to the calibrated analytical simulator ($<1$ is better); survival is the percentage of each trajectory replayed before the simulated replay fails (higher is better).
    }
  \label{fig:model_based_abl}
\end{figure}

\paragraph{Component ablations.}
Combining per-contact and joint-wise delta-action compensation achieves higher mean real-world performance than either alone, suggesting potential synergy (Fig.~\ref{fig:model_based_abl}(A)).
Contact-only compensation generally outperforms action-only compensation under limited data, consistent with a useful inductive bias from contact-level correction. The exception on \trianglepassdownreversed may reflect limited contact coverage or trajectory quality when successful transitions are scarce.

\paragraph{Tracking across tasks and policy updates.}
Relative to the calibrated analytical simulator, contact compensation reduces held-out orientation error to 0.75--0.97 and position error to 0.45--0.99 of the analytical values across all 16 source--target pairs of the four \pass variants, and generally extends survival (Fig.~\ref{fig:model_based_abl}(B)). 
Across all ten tasks, this advantage persists but narrows as the policy changes (Supplementary Section~S3).

\subsection*{Structured Latent Exploration and Residual Execution} \namedlabel{sec:moodel_free_analysis}{Analysis on Real-World RL}

\begin{figure}[htbp]
  \centering
    \includegraphics[width=1.0\textwidth,height=0.82\textheight,keepaspectratio]{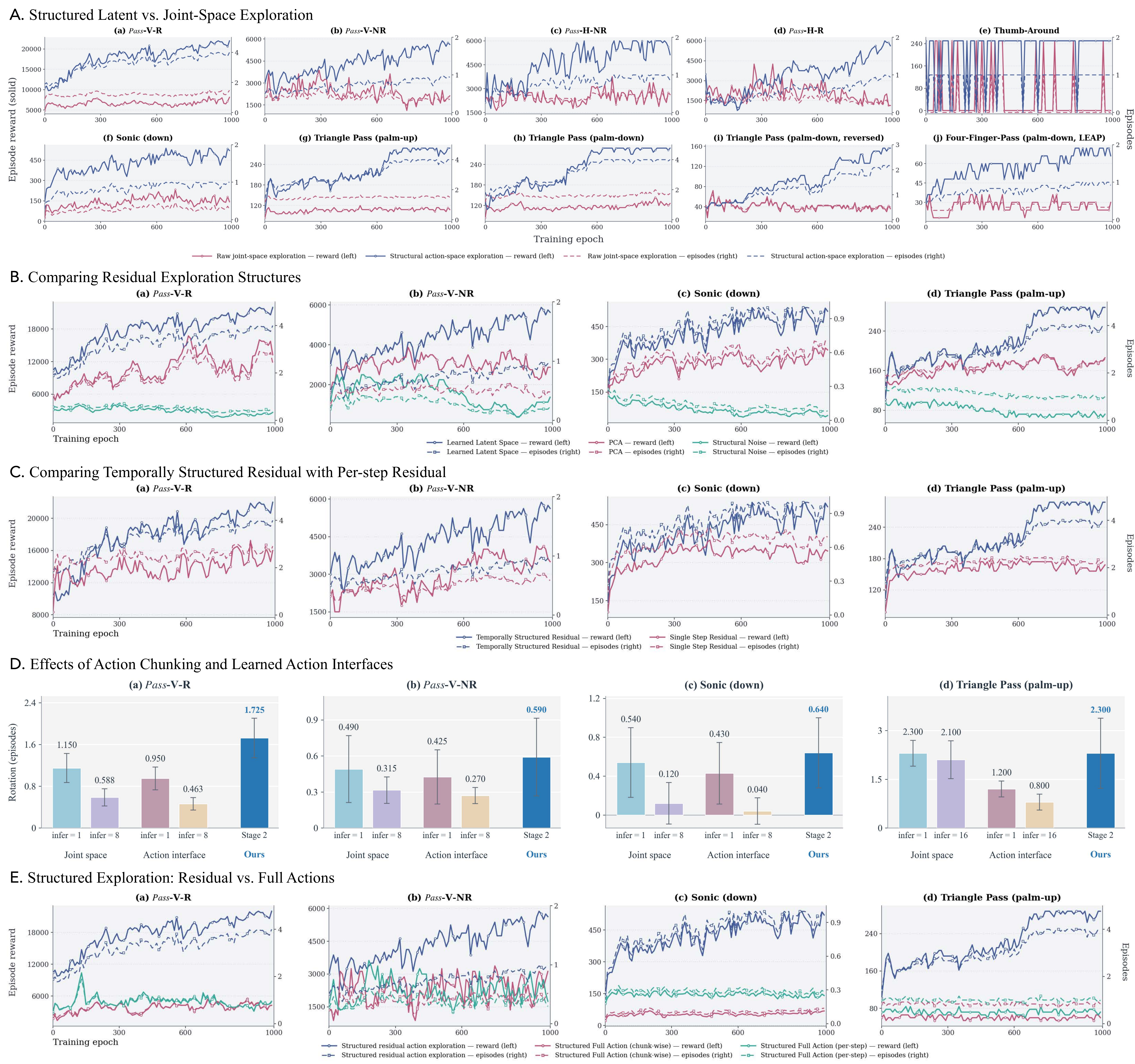}
    \caption{\footnotesize
    \textbf{Structured Latent Exploration and Residual Execution.}
    \textbf{(A)} Structural action-space exploration versus raw joint-space residual exploration across ten tasks.
    \textbf{(B)} Learned nonlinear latent, PCA-based linear, and chunk-constant joint residuals (``Structural Noise''), with matched initial residual magnitudes and per-step base-policy feedback.
    \textbf{(C)} Temporally structured residual chunks versus per-step residuals.
    \textbf{(D)} Base-policy sensitivity to action chunking: action-chunk
    sequences predicted directly or through a learned decoder, executed in full or replanned after the first action, compared with our single-step Stage 2 policy. Bars show mean $\pm$ population SD over 20, 50, 50, and 5 trials on \passvzerons, \passvonens, \sonicdownns, and \trianglepassupns, respectively.
    \textbf{(E)} Chunk-wise residual exploration versus chunk-wise and per-step full decoded-action exploration; the per-step variant executes only the first decoded action before replanning.
    Curves in (A--C,E) report real-world training performance; solid and dashed lines denote episode reward and episodes, respectively. 
    Experimental details appear in Supplementary Section~S2.
}
  \label{fig:structural_and_residual_exploration}
\end{figure}

Our design coordinates corrections across joints and time while retaining per-step base-policy feedback: the learned representation parameterizes only corrections, and base-policy commands bypass it. We examine four questions (Fig.~\ref{fig:structural_and_residual_exploration}):
\textbf{(1)} Does learned latent structure improve on joint-space exploration or simpler spatial and temporal coordination (A,B)?
\textbf{(2)} Do temporally coordinated residuals improve on per-step residuals (C)?
\textbf{(3)} Prior real-world tuning often uses chunk-wise policies~\cite{Yang2026SERNFSR} or learned action interfaces~\cite{Zhu2026TowardsHD,Yang2026LAMPLM};
how do these choices affect base-policy performance on dynamic tasks (D)?
\textbf{(4)} Does residual execution outperform full decoded actions during RL, even with per-step replanning (E)?

\paragraph{(1) Learned structure.}
Learned structured residuals outperform per-step joint-space residuals across ten tasks, despite additional noise calibration and conservative tuning of the joint-space baseline (A).
On four tasks, they also achieve the highest final reward and episodes, above linear residuals based on principal components analysis (PCA), which perform intermediately, and joint residuals held constant within each chunk, which show limited gains or deteriorate (B).

\paragraph{(2) Temporal coordination.}
With the same representation and per-step base feedback, chunked residuals outperform per-step residuals on all four tasks (C).

\paragraph{(3) Action chunking in the base policy.}
Behavior-cloned policies that predict action chunks, 
directly or through a learned decoder, improve when replanned after the first action, and our single-step Stage-2 policy performs best or ties the best alternative (D). Per-step base feedback should therefore be preserved during exploration.

\paragraph{(4) Residual versus full-action execution.}
Full decoded-action exploration improves with per-step replanning on selected tasks 
but remains well below residual exploration on all four tasks, despite conservative tuning (E):
complete commands stay constrained by the learned representation, and first-action-only execution forgoes within-chunk coordination.

Learning curves (A--C,E) report training-rollout performance to avoid frequent separate hardware evaluations~\cite{rl100}; exploration noise can cause an initial dip below deterministic base-policy performance.
Details of the calibration and conservative tuning are deferred to Supplementary Section~S2.

%% file: articles_c/discussion.tex
\section*{Discussion}

\paragraph{Toward reliable dynamic dexterity.}
Using a single framework, \rps{} reliably executes six dynamic ``pen'' spinning skills on an off-the-shelf, high-degree-of-freedom hand at pen-axis angular speeds of
$1.48\pi$--$3.1\pi\,\mathrm{rad/s}$, each succeeding in at least 90\% of 50 trials. Prior systems rotate pens more
slowly~\cite{Wang2025PenSpin,Liu2025DexNDMCT,Li2026TeleDexter,Wu2026TopoRetarget}, and prior fast spins are task-specific, relying on specialized hardware or soft
manipulators~\cite{Yao2025SWIFT,Nakatani2023PenSpinning,Ishihara2006DynamicPS}. See Supplementary Section~S2 for speed estimates.

\paragraph{Reliable and sustained rotation.}
We sustain rotation for over 20 minutes on slower tasks requiring delicate, geometrically constrained finger gaiting, and over 30 minutes with a rectangular board beyond pen spinning. These results extend the duration of prior demonstrations on challenging elongated objects~\cite{Liu2025DexNDMCT,Wang2025PenSpin,Wu2026TopoRetarget,Li2026TeleDexter}.

\paragraph{Analytical structure as an inductive bias.}
Our shared per-contact neural compensator retains analytical structure and encourages cross-contact generalization for policy refinement. In preliminary studies, a NeRD-style hybrid~\cite{Xu2025NeuralRD} and a fully neural model, each trained on millions of simulated trajectories, predicted held-out action sequences reasonably well but degraded under closed-loop feedback and policy optimization (Supplementary Section~S3). 
Collision geometry is not corrected explicitly. The compensator's contact-local inputs may absorb remaining discrepancies (Supplementary Sections~S2 and~S3).

\paragraph{Practical costs.}
Our framework requires human involvement in all three stages, but the workload remains manageable: corrective optimization typically takes under 30 minutes per failure, and beyond initial annotation and editing, human effort mainly involves resetting the setup and scoring outcomes. Figure~\ref{fig:framework_stages}(C) summarizes human effort and hardware time, with full rollout budgets, timing, and compute details in Supplementary Section~S2.

\paragraph{Limitations and future directions.}
Several tasks require relatively thick pens because of the hand's size, and training (but not deployment) relies on motion capture for trajectories and rewards. The system assumes task-appropriate initial grasps and does not address autonomous initialization, recovery from arbitrary states, or skill sequencing. Zero-shot transfer remains sensitive to geometry, compliance, and friction, and variants evaluated without repeated trials offer only qualitative evidence (Supplementary Section~S2). Future directions include tactile sensing, generalization across initial grasps, and automated resets.

%% file: articles_c/method.tex
\section*{Materials and Methods}

\begin{figure}[htbp]
  \centering
  \includegraphics[width=1.0\textwidth,height=0.84\textheight,keepaspectratio]{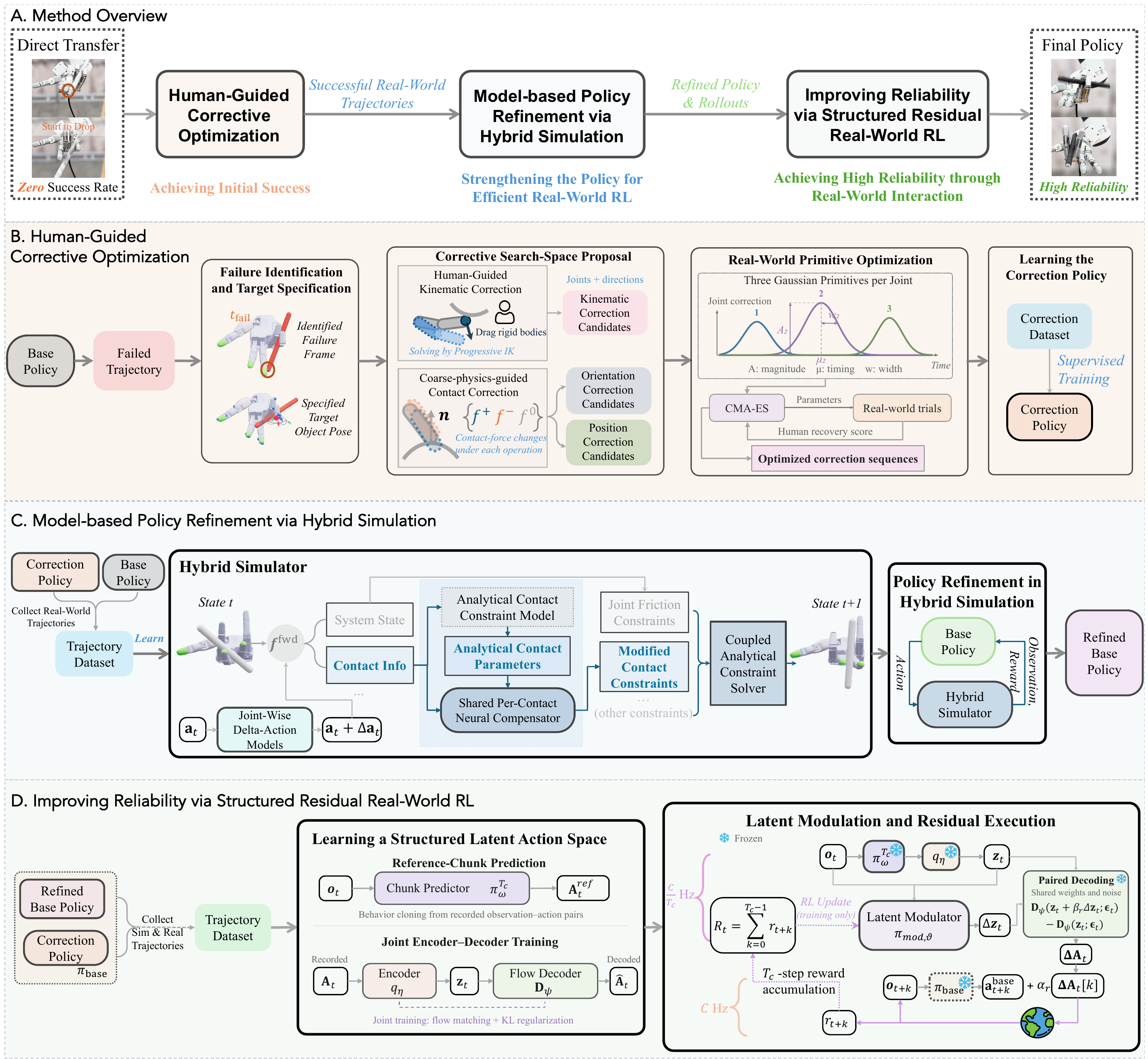}
  \caption{\footnotesize
    \textbf{Overview of the Three-Stage Sim-to-Real Framework.}
    \textbf{(A) Method Overview:} Corrective optimization enables initial success, model-based refinement improves performance, and model-free RL further improves reliability.
    \textbf{(B) Human-guided corrective optimization.} Offline human kinematic guidance and coarse contact analysis define a compact search space of key joints and motion directions for each targeted failure. Real-world optimization of corrective primitives within this space yields precise, effective correction sequences.
    \textbf{(C) Model-based refinement.} A shared per-contact neural compensator corrects the simulator's contact computation; joint-wise delta-action models address remaining discrepancies.
    \textbf{(D) Structured Residual Real-World RL.} Exploring residual action sequences through a structured action space preserves high-frequency closed-loop base-policy execution and fine-grained control while keeping the base controller outside the action codec.
}
  \label{fig:method}
\end{figure}

We aim to develop a sim-to-real framework for reliable execution of challenging in-hand manipulation, using ten pen-spinning tasks with diverse transfer challenges as a structured testbed (Fig.~\ref{fig:rollout_figure}(A)). Direct transfer yields zero success, making initial completion challenging, and reliable execution further demands robustness to trial-to-trial variations in initial hand--object configurations, contact interactions, and actuator response, as well as to accumulating errors.

Without initial success, common model-based and model-free approaches face coupled challenges: failed rollouts provide limited coverage for learning accurate, generalizable dynamics~\cite{Fey2025BridgingTS,He2025ASAPAS,Liu2025DexNDMCT,Xu2025NeuralRD,Shi2024AnEM}, and weak starting policies can make real-world exploration prohibitively costly~\cite{Zhu2026TowardsHD,Yu2026BeyondAR,Yang2026LAMPLM,luo2024hilserl,Han2026DexHiLAH,rl100,Sun2026FromPT}. We therefore design three complementary stages that build on one another, from task completion to reliability, within a practical real-world interaction budget (Fig.~\ref{fig:method}).

Offline human-guided corrective optimization first resolves critical failures and establishes initial success. The resulting trajectories span essential task transitions, providing informative data for learning a simulator in which we refine the policy. In our experiments, continued model-based refinement yields diminishing returns (Fig.~\ref{fig:framework_stages}(B)); further gains may require substantially more real-world data covering a much broader distribution for accurate, generalizable contact modeling. With the refined policy reducing exploration costs, we therefore transition to model-free real-world RL to directly optimize task performance and reliability.

\subsection*{Human-Guided Corrective Optimization}
\label{sec:method_failure_recovery}

Preventing catastrophic failures---object drops, stalled rotation, or unintended motion---can require substantial changes in hand configuration, contact modes, and forces. Correcting such failures offers a route to initial success, but existing approaches are limited in our tasks. Human takeover~\cite{Kelly2019HGDAgger,Zhu2026TowardsHD,Li2026HandintheLoopIV,Han2026DexHiLAH} demands rapid, force-sensitive corrections that anticipate contact dynamics and permit policy resumption (Fig.~\ref{fig:human_takeover_failures}(B,C)); online repair by direct finger guidance~\cite{Li2026ReDexRS} faces the same demands and additionally requires compliant hardware. Task-specific heuristics correct finger alignment in piano playing~\cite{Xie2026HandelBotRP}, but corrections for our coupled articulated-body and hand--object contact dynamics, often during rapid motion, are difficult to prescribe by intuition alone. Foundation-model agents~\cite{liu2023reflect,guo2024doremi,dai2025racer,lin2026failsafe} lack contact-dynamics grounding, and dedicated recovery policies~\cite{thananjeyan2021recovery} face a circular dependency: learning recovery requires the capabilities needed for the original task. Direct real-world optimization offers an alternative, but unrestricted search is prohibitively costly. 

Our method builds on real-world corrective optimization. We make it tractable by narrowing the search space through offline human guidance and coarse contact analysis and by parameterizing corrections with low-dimensional primitives that remain expressive for brief, targeted corrections; optimization often takes fewer than 10 iterations and typically no more than 30 minutes per failure.

\paragraph{Failure identification and target specification.}
From motion-capture recordings of hand and object states and applied actions, an annotator identifies the onset of the targeted failure, $t_{\mathrm{fail}}$, offline and, when needed, specifies a desired object position, orientation, or both in an interactive viewer (Fig.~\ref{fig:human_takeover_failures}).

\paragraph{Corrective search-space proposal.}
Given the identified failure and specified targets, we design two complementary routes to construct a compact search space for real-world optimization of corrective primitives. Each route proposes candidate joints and motion directions and is invoked only when needed.
In \textit{human-guided kinematic correction}, the annotator edits the hand configuration, and progressive inverse kinematics favors few modified joints, whose indices and displacement directions form a kinematic candidate. Because kinematic edits give limited guidance for force adjustments, \textit{coarse-physics-guided contact correction} proposes joints and directions for fine-grained contact-force adjustments. For each contact or near-contact $i$, evaluated with the object at its original pose and the hand after any kinematic edit, an operation $o_i$ strengthens ($+$), softens ($-$), or keeps ($0$) the contact. A coarse contact model assigns each operation a force-increment proxy $\Delta\widehat{\mathbf{f}}_i(o_i)$, and contact-point Jacobians identify feasible joints and motion signs. An orientation target $R^\star$ defines a desired torque direction $\widehat{\boldsymbol{\tau}}_d$; we choose tangential directions favoring this rotation and rank feasible combinations $\mathbf{o}=(o_1,\ldots,o_m)$ by their torque alignment $s_R(\mathbf{o})=\widehat{\boldsymbol{\tau}}_d^\top\sum_{i=1}^{m}\mathbf{r}_i\times\Delta\widehat{\mathbf{f}}_i(o_i)$, where $\mathbf{r}_i$ points from the object reference point to contact $i$. A position target can reveal drift that began earlier, so we trace back to the frame whose hand--object contact relations best match the edited failure frame, choose tangential directions that oppose the drift, and rank combinations by net-force alignment. Kinematic, orientation, and position candidates form separate sets. Each candidate specifies joint indices, motion directions, and a peak-correction frame.

\paragraph{Real-world primitive optimization.}
We take one candidate from each active set in recommendation order and jointly optimize their corrective primitives. Each joint's delta-action sequence typically comprises three Gaussian pulses with learnable centers, widths, and peak magnitudes, signed by the proposed motion direction, capturing timing, duration, and strength without per-timestep optimization. CMA-ES~\cite{Hansen2001CompletelyDS} optimizes these parameters using a human-assigned score in $[-1,1]$ for each real-world intervention. The highest-ranked contact-operation proposals usually sufficed.

\paragraph{Learning the correction policy.}
Once the primitives enable full-episode success, we train a correction policy on real-world rollouts that maps the base policy's observation and predicted action to an additive correction. The base and correction policies together form the first-stage controller (details in Supplementary Section~S1).

\subsection*{Model-Based Policy Refinement via Hybrid Simulation}
\label{sec:method_model_based}

Although failure recovery enables initial success, the policy remains vulnerable to variable execution conditions and cumulative drift, and the number and trial-to-trial variability of remaining failures make further corrective optimization difficult to scale. Because first-stage trajectories cover all task phases, we use them to learn task-relevant dynamics from limited real-world data for policy refinement.
The central challenge in making such model-based refinement effective is balancing expressivity and generalization of the learned dynamics model~\cite{Liu2025DexNDMCT,Li2025RoboticWM,Xu2025NeuralRD,Shi2024AnEM,Fey2025BridgingTS}: the model must capture contact dynamics accurately and generalize to behaviors explored during policy updates. 
Full neural models can struggle to generalize, and reduced-order models have limited fidelity and cannot model contact dynamics accurately~\cite{Liu2025DexNDMCT}.
Hybrid simulation~\cite{Xu2025NeuralRD,Ataei2026FlashNeRDPC,Heiden2020NeuralSimAD} offers an alternative whose effectiveness depends on how analytical and learned components share modeling responsibility and on the granularity of neural compensation.

Pursuing full-order dynamics modeling while supporting generalization, we adopt a hybrid approach. Our key observation is that contact dynamics are central to manipulation yet difficult to model accurately. The central idea is to embed a shared per-contact neural compensator within the physics loop of MuJoCo Warp~\cite{mujocowarp2025}, a GPU-accelerated implementation of MuJoCo~\cite{todorov2012mujoco}, correcting its analytical contact model while retaining the solver, with joint-wise delta-action models at the action interface compensating for remaining discrepancies.

\paragraph{Constraint-based contact formulation.}
The analytical simulator models normal contact and friction with convex soft constraints. Each contact's response is shaped by its reference acceleration $\mathbf{a}_{\mathrm{ref},c}\in\mathbb{R}^{n_c}$, a corrective motion target in which a virtual spring counters penetration and damping opposes relative contact motion, and by inverse regularization weights $\mathbf{D}_c\in\mathbb{R}_{>0}^{n_c}$, which set how strictly the soft response is driven toward this target (larger weights enforce it more strongly, smaller weights allow compliance); $n_c$ is the contact constraint dimensionality. Contact constraints enter a coupled solve with all other active constraints to determine forces and accelerations.

\paragraph{Per-contact neural compensation.}
We learn a shared network that corrects contact responses by modulating $\mathbf{a}_{\mathrm{ref},c}$ and $\mathbf{D}_c$ before the coupled solve, while retaining the analytical solver and its contact feasibility conditions.
At every physics step, it predicts $(\mathbf{u}_{D,c},\mathbf{u}_{a,c})=g_\theta(\mathcal{C}_c,\mathbf{a}_{\mathrm{ref},c},\mathbf{D}_c)$ from local-frame features $\mathcal{C}_c$ encoding geometry, contact parameters, kinematics, and constraint quantities, which scale $\mathbf{D}_c$ multiplicatively and shift $\mathbf{a}_{\mathrm{ref},c}$ additively within bounds:
\begin{equation}
    \begin{aligned}
    \widetilde{\mathbf{D}}_c &= \eta\,\exp\!\left(\alpha_c\tanh(\mathbf{u}_{D,c})\right)\odot\mathbf{D}_c,\\
    \widetilde{\mathbf{a}}_{\mathrm{ref},c} &= \mathbf{a}_{\mathrm{ref},c}+\beta_c\tanh(\mathbf{u}_{a,c}),
    \end{aligned}
    \label{eq:bounded_contact_corrections}
\end{equation}
where operations are element-wise, $\alpha_c$ and $\beta_c$ bound the corrections, and $\eta>0$ is a per-environment renormalization factor relaxed to $1$ during training (Supplementary Section~S1).
Because ground-truth compensation targets are unavailable, we train $g_\theta$ by RL trajectory tracking~\cite{Hwangbo2019LearningAA,He2025ASAPAS}: real-world trajectories from the combined base and correction policies are replayed open loop from their recorded initial states, and a weighted reward measures agreement of simulated hand and object states with the recorded ones.

\paragraph{Joint-wise delta-action compensation.}
To compensate for remaining discrepancies in the contact formulation and other analytical components, we follow prior work~\cite{Liu2025DexNDMCT,Fey2025BridgingTS} and learn joint-wise delta-action models that operate at the action interface. Each model predicts an additive correction from its joint's state--action history, limiting input dimensionality to encourage generalization. The models are trained to align simulated transitions under corrected actions with real-world transitions under the original actions.
Without ground-truth action corrections, we train the delta-action models through trajectory matching~\cite{Liu2025DexNDMCT}. We first learn a differentiable dynamics model that approximates the contact-compensated simulator, predicting the next state from a short history of states and actions.
We then freeze this model and train the delta-action model by backpropagating multi-step hand-joint and object-orientation errors through it, using real-world trajectories collected by the policy refined in that simulator.

\paragraph{Policy refinement in hybrid simulation.}
We fine-tune the base policy in the contact-compensated simulator with proximal policy optimization (PPO)~\cite{Schulman2017ProximalPO}, keeping the first-stage correction policy frozen and excluded from simulation. 
Using the refined base policy and the correction policy, we collect rollouts in the contact-compensated simulator (base policy only) and on hardware. We use these data to train the delta-action models, then further fine-tune the base policy with both compensators enabled.
Each stage warm-starts the critic from the preceding one (feature definitions, rewards, and training details in Supplementary Sections~S1 and~S2).

\subsection*{Improving Reliability via Structured Residual Real-World RL}
\label{sec:method_realworld_rl}

In our experiments, further rounds of real-world data collection and model-based refinement yield diminishing improvements (Fig.~\ref{fig:framework_stages}(B)), possibly reflecting residual prediction errors as updated policies encounter states and contact transitions poorly covered by the available data. 
Because model-based refinement yields a stronger base policy that reduces exploration cost, we then introduce real-world RL to improve reliability through direct interaction.

Effective real-world RL for dynamic, forceful, contact-sensitive manipulation requires sample-efficient, coordinated exploration, responsive feedback, and precise commands. Poorly coordinated joint perturbations can disrupt contacts. Generating coordinated actions from learned latent representations~\cite{Yang2026LAMPLM,Zhu2026TowardsHD,Yu2026BeyondAR} can improve sample efficiency and is widely adopted, but routing complete commands through these representations can introduce reconstruction errors, constrain corrective exploration, or, when actions are chunked, delay feedback.

Building on these works, we construct a spatially and temporally structured latent action space for task-relevant exploration. However, we use it only for low-frequency residual corrections to the base policy, which retains high-frequency closed-loop execution, separating exploration from feedback control.

\paragraph{Learning a structured latent action space.}
From real-world and hybrid-simulator rollouts of the refined policy, we pair each observation $\mathbf{o}_t$ with its subsequent action chunk $\mathbf{A}_t=[\mathbf{a}_t,\ldots,\mathbf{a}_{t+T_c-1}]^\top\in\mathbb{R}^{T_c\times d_a}$, where $T_c$ is the chunk length and $d_a$ the action dimension. A multilayer perceptron predictor $\pi_\omega^{T_c}$, trained by mean-squared-error behavior cloning, outputs a reference chunk $\mathbf{A}^{\mathrm{ref}}_t=\pi_\omega^{T_c}(\mathbf{o}_t)$. Separately, a variational encoder $q_\eta$ and a conditional flow-matching decoder $\mathbf{D}_\psi$ are jointly trained on standardized action chunks with a flow-matching loss~\cite{Lipman2022FlowMF} and Kullback--Leibler (KL) regularization toward a standard Gaussian prior.

\paragraph{Latent modulation and residual execution.}
Using these models, we augment the second-stage refined base policy with a latent modulator $\pi_{\mathrm{mod},\vartheta}$, trained via model-free real-world RL to further improve performance. At each chunk boundary, we encode the reference action chunk and sample a latent correction from the modulator:
\begin{equation}
    \mathbf{z}_t \sim q_\eta(\cdot\mid\mathbf{A}^{\mathrm{ref}}_t), \qquad \Delta\mathbf{z}_t \sim\pi_{\mathrm{mod},\vartheta}(\cdot\mid\mathbf{o}_t,\mathbf{z}_t).
\end{equation}
Decoding the reference and modulated latents with the same Gaussian initialization $\boldsymbol{\epsilon}_t$ gives the residual chunk
\begin{equation}
    \Delta\mathbf{A}_t = \mathbf{D}_\psi(\mathbf{z}_t+\beta_r\Delta\mathbf{z}_t;\boldsymbol{\epsilon}_t) - \mathbf{D}_\psi(\mathbf{z}_t;\boldsymbol{\epsilon}_t),
\end{equation}
where $\beta_r=0.25$ scales the latent modulation. At every control step, we recompute the base-policy action from the latest observation and add the corresponding residual:
\begin{equation}
    \mathbf{a}^{\mathrm{exec}}_{t+k} = \pi_{\mathrm{base}}(\mathbf{o}_{t+k}) + \alpha_r\,\Delta\mathbf{A}_t[k], \qquad k=0,\ldots,T_c-1,
\end{equation}
where $\alpha_r$ scales the residual and $\pi_{\mathrm{base}}$ includes the second-stage refined base and first-stage correction policies. Thus, $\Delta\mathbf{z}_t=\mathbf{0}$ recovers the base policy's action, and nonzero modulation, applied once per chunk, enables structured exploration while the base policy retains its fine-grained commands and original feedback frequency.

\paragraph{Simulation warm-start and real-world training.}
We train the modulator with soft actor-critic (SAC)~\cite{Haarnoja2018SoftAO}, pretraining its actor and critics in simulation before continuing on the real robot, where motion capture supplies object poses for task-specific rewards and termination checks (details in Supplementary Section~S1).

\subsection*{Statistical Analysis}
Success rate is the fraction of successful single-episode trials. For sustained rotation, average episodes is the mean number of task-specific rotation cycles (episodes) completed per trial, including fractional progress.
To limit hardware wear and actuator heating, we evaluated some generalization and adaptation variants over 10--40 rather than 50 trials and report 95\% Wilson score intervals to reflect the resulting uncertainty.
Supplementary Section~S2 gives task-specific definitions, success criteria, and evaluation protocols.
Real-world results use one training run per learning configuration, with repeated evaluation of the resulting controller.
Each run requires hours of supervised hardware interaction, in our experiments (Fig.~\ref{fig:framework_stages}(C)) as in prior work~\cite{Mahmood2018SettingUA,Kalashnikov2018QTOptSD}; real-world RL studies often report a single run per configuration~\cite{Wu2022DayDreamerWM}, and even simulated benchmarks are commonly evaluated with few runs per task~\cite{Agarwal2021DeepRL}.
We therefore base comparisons on differences consistent across tasks.
An independent rater re-scored a random half of the real-world evaluation videos (5 videos, 250 trials) with these criteria and agreed with all original scores (Cohen's $\kappa=1.00$).

%% file: articles_c/acknowledgments.tex






\section*{Acknowledgments}
We thank Haochen Huang for simulation discussions and feedback on the model-based refinement description.

\paragraph*{Funding:}
Partially supported by the New Generation Artificial Intelligence-National Science and Technology Major Project (No. 2025ZD0122905; H.W.).

\paragraph*{Author contributions:}
X.L. led the project, contributing algorithms, software, experiments, and writing. H.C. contributed to Isaac Lab experiments; X.F., to a \sonic variant simulation policy. H.W. contributed discussions on methods, experiments, and research direction. L.Y. supervised.

\paragraph*{Competing interests:} 
All authors are affiliated with Galbot (X.L., H.C., and X.F., interns; L.Y., part-time; H.W., chief technology officer) and declare no other competing interests.




\noindent\textbf{Data and materials availability:}
Supporting data appear in the paper or Supplementary Materials. 
Materials are commercially available or described in the paper.

%% file: supp/supp.tex




\beginsupplement


\makesupplementtitle{%
}{}

\begin{supplementbox}
\begin{minipage}[t]{0.55\linewidth}\raggedright
\textbf{This PDF file includes:}\\[2pt]
Supplementary Sections S1 to S3 \\
Figures~\ref{fig:supp_method_failure_recovery_workflow} to~\ref{fig:baseline_failures}\\
Tables~\ref{tab:supp_evaluation_protocols} to~\ref{tab:supp_two_operator_recovery}\\
References \textit{(76--\lastrefnumber)}\\
Captions for Movies S1 to S16\\
Captions for Data S1 to S3
\end{minipage}\hfill
\begin{minipage}[t]{0.42\linewidth}\raggedright
\textbf{Other Supplementary Materials for this manuscript:}\\[2pt]
Movies S1 to S16\\
Data S1 to S3
\end{minipage}
\end{supplementbox}

\supplementcontents

\newpage



\input{supp/supp_method_3}

\input{supp/supp_expsettings_4}

\input{supp/supp_additional_res_2}

%% file: supp/supp_method_3.tex
\section{Supplementary Materials and Methods}
\label{sec:supp_method}

\subsection*{Human-Guided Corrective Optimization}
\namedlabel{sec:failure_recovery_supp}{Human-Guided Corrective Optimization}

\begin{figure}[htbp]
  \centering
  \includegraphics[width=\textwidth]{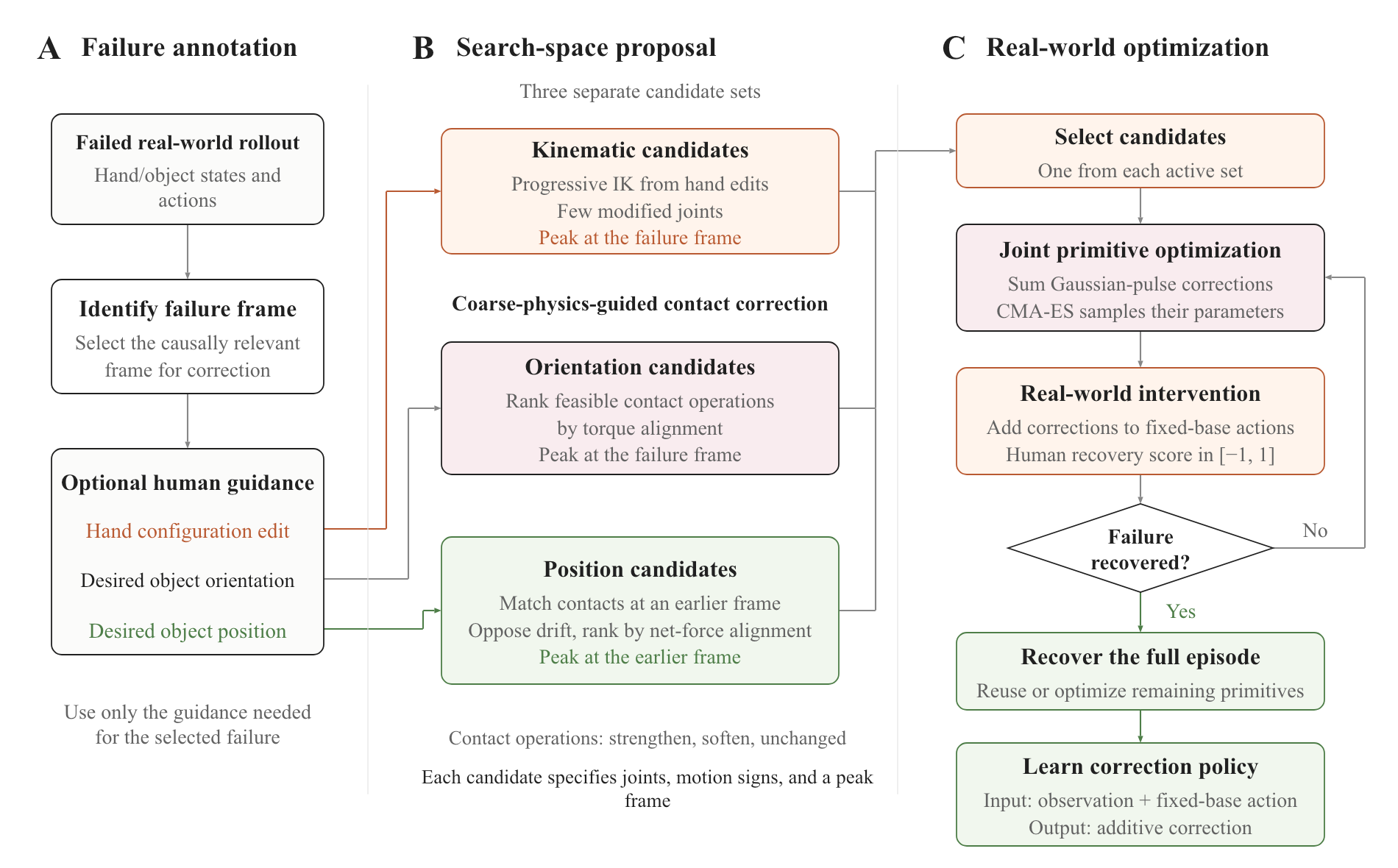}
  \caption{\footnotesize
  \textbf{Human-Guided Corrective Optimization.} 
     Failure identification and target specification, corrective search-space proposal through kinematic editing and coarse contact analysis, and real-world primitive optimization. 
  }
  \label{fig:supp_method_failure_recovery_workflow}
\end{figure}

Our coarse-to-fine workflow first identifies failures and specifies target object poses when needed, then uses human-guided kinematic correction and coarse-physics-guided contact correction to propose compact search spaces for real-world primitive optimization (Figs.~\ref{fig:method} and~\ref{fig:supp_method_failure_recovery_workflow}).

\subsubsection*{Failure identification and target specification}

\begin{figure}[htbp]
  \centering
  \includegraphics[width=\textwidth]{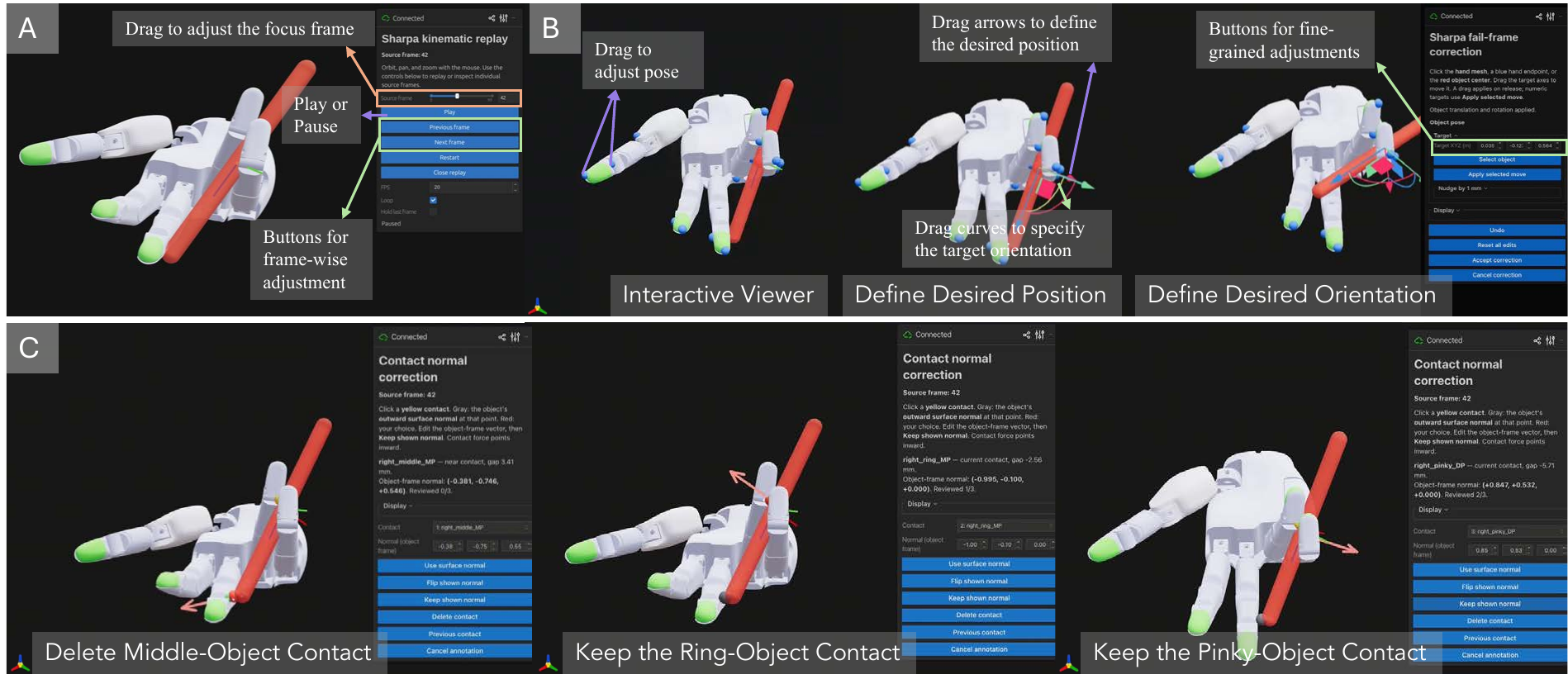}
  \caption{\footnotesize
  \textbf{Interactive Kinematic Editing and Contact Annotation.} \textbf{(A)} The kinematic replay viewer. \textbf{(B)} Hand-kinematic editing and target object-pose specification. \textbf{(C)} Contact selection and normal-direction adjustment. The interfaces are built with Viser~\cite{yi2026viser}. 
  }
  \label{fig:interactive_viewer_for_kinematic_correction_and_annotations}
\end{figure}

An annotator replays recorded hand states, object states, and actions, selecting a frame where the object falls, jams, or fails to move as intended. Dragging or buttons specify a desired object position, orientation, or both when needed (Fig.~\ref{fig:interactive_viewer_for_kinematic_correction_and_annotations}).

We anchor peak correction to visible failure because pose, velocity, and contact deviations are easier to annotate than their underlying acceleration errors. Primitives may start earlier. This assumes a short error-to-observation delay without fundamental contact-topology changes, applicable to nine tasks, including the \pass and \trianglepass families and \fourfingerpassleapns. Underactuated \thumbaround is the exception: launch errors can become visible after launch contacts disappear. We therefore anchor launch/flight corrections immediately before launch, using human judgment to identify the causally relevant grasp; catching failures use the subsequent contact frame.

\subsubsection*{Corrective search-space proposal}

Human-guided kinematic correction and coarse-physics-guided contact correction propose candidate joints and motion directions. 

\paragraph{Human-guided kinematic correction.}
The annotator edits hand bodies in the same viewer. Progressive inverse kinematics rejects joint-limit violations and first frees the edited finger's most proximal joint, adding distal joints on request. The annotator accepts a solution defining the kinematic joint subset and motion signs:
\begin{equation}
\begin{aligned}
\Delta\mathbf{q}_h^{\mathrm{kin}}&=\mathbf{q}_h^{\mathrm{edit}}-\mathbf{q}_h(t_{\mathrm{fail}}),\\
S_{\mathrm{kin}}&=\{j:\Delta q_{h,j}^{\mathrm{kin}}\neq0\},
\qquad \sigma_j^{\mathrm{kin}}=\operatorname{sign}(\Delta q_{h,j}^{\mathrm{kin}}).
\end{aligned}
\label{eq:supp_recovery_kinematic_candidate}
\end{equation}
Equation~\eqref{eq:supp_recovery_kinematic_candidate} defines the kinematic candidate; object-pose targets guide contact analysis below.

\paragraph{Coarse-physics-guided contact correction.}
The coarse model proposes force-sensitive contact operations, joint subsets, and directions; real-world optimization determines timing and magnitude. The viewer also permits contact removal/retention and normal-direction edits.

\noindent\textit{Contacts and frames.}
With the object at its original pose and the hand after any kinematic edits, we retain active contacts ($d_i\leq0$) and near contacts ($0<d_i\leq d_{\max}=0.01\,\mathrm m$), at most one per finger ($m\leq5$). For object pose $(\mathbf p,R)$ and world contact position/outward normal $(\mathbf c_i^W,\mathbf n_i^W)$, define object-frame quantities $\mathbf r_i=R^\top(\mathbf c_i^W-\mathbf p)$ and $\mathbf n_i=R^\top\mathbf n_i^W$. All following directions and wrenches use this frame, with moments about $\mathbf p$.

\noindent\textit{Orientation-guided operations.}
Given target $R^\star$, define $\mathbf e_R=\operatorname{Log}(R^\top R^\star)^\vee$ and desired torque direction $\widehat{\boldsymbol\tau}_d=\mathbf e_R/\|\mathbf e_R\|_2$ for nonzero axis-angle error. With $P_i=I-\mathbf n_i\mathbf n_i^\top$ and $\mathbf g_i=P_i(\widehat{\boldsymbol\tau}_d\times\mathbf r_i)$, set $\mathbf t_{i,1}=\mathbf g_i/\|\mathbf g_i\|_2$ when nonzero, otherwise $\mathbf0$. The nonzero direction maximizes desired torque from a unit tangential force; $\mathbf t_{i,2}=\mathbf n_i\times\mathbf t_{i,1}$ completes $C_i=[\mathbf t_{i,1}\;\mathbf t_{i,2}\;\mathbf n_i]$. A zero projection omits the tangential frame and friction component.

Operations $o_i\in\{+,-,0\}$ strengthen, soften, or leave a contact unchanged. Strengthening requires inward normal motion and positive tangential alignment when available; softening requires outward motion:
\[
\mathcal V_i^+=\{\mathbf v:-\mathbf n_i^\top\mathbf v>0,\quad
\mathbf t_{i,1}^\top\mathbf v>0\ \text{if }\|\mathbf g_i\|_2>0\},
\qquad \mathcal V_i^-=\{\mathbf v:\mathbf n_i^\top\mathbf v>0\}.
\]
An unchanged contact adds no motion.

\noindent\textit{Joint feasibility and force ranking.}
Holding object pose fixed, the hand-link translational Jacobian becomes $J_i=R^\top J_i^W(\mathbf q_h)\in\mathbb R^{3\times n_{\mathrm{dof}}}$. For joint subset $S_i$ and signs $\boldsymbol\sigma_i\in\{-1,+1\}^{|S_i|}$, let $V_i=J_i[:,S_i]\operatorname{diag}(\boldsymbol\sigma_i)$. Retain a realization if $\exists\boldsymbol\alpha_i\geq0$ with $\mathbf1^\top\boldsymbol\alpha_i=1$ and $V_i\boldsymbol\alpha_i\in\mathcal V_i^{o_i}$; normalization removes arbitrary motion scale.

We enumerate at most $3^5=243$ operation combinations, discarding infeasible ones. Directional force-increment proxies are
\[
\Delta\widehat{\mathbf f}_i(+)= -a_i\mathbf n_i+b_i\mathbf t_{i,1},\qquad
\Delta\widehat{\mathbf f}_i(-)=c_i\mathbf n_i,\qquad
\Delta\widehat{\mathbf f}_i(0)=\mathbf0,
\]
with nonnegative $a_i=b_i=c_i=1.0$. Softening reduces an existing inward force rather than pulling outward; these proxies do not predict full dynamics. A zero tangential projection leaves only the normal component.

Sum $\Delta\widehat{\mathbf F}=\sum_i\Delta\widehat{\mathbf f}_i$ and $\Delta\widehat{\boldsymbol\tau}=\sum_i\mathbf r_i\times\Delta\widehat{\mathbf f}_i$. Rank combinations with $s_R(\mathbf o)=\widehat{\boldsymbol\tau}_d^\top\Delta\widehat{\boldsymbol\tau}>0$ by decreasing $s_R$; within each operation, prefer feasible subsets with fewer joints.

\noindent\textit{Earlier position correction.}
A position annotation activates an earlier anchor within preceding frames $\mathcal T_{\mathrm{stable}}$ whose contacting links remain unchanged:
\[
t_{\mathrm{pos}}=\underset{t\in\mathcal T_{\mathrm{stable}}}{\arg\min}
\sum_i\|\mathbf r_i(t)-\mathbf r_i^{\mathrm{edit}}\|_2^2.
\]
This compares object-relative contacts, not world positions. Recompute contacts there and oppose drift toward failure using $\mathbf u_{\mathrm{pos}}=R_{OH}(t_{\mathrm{pos}})({}^{H}\mathbf p(t_{\mathrm{pos}})-{}^{H}\mathbf p(t_{\mathrm{fail}}))$, where ${}^{H}\mathbf p$ is hand-frame object position and $R_{OH}$ maps hand to object coordinates. Annotations guarantee nonempty $\mathcal T_{\mathrm{stable}}$ and nonzero $\mathbf u_{\mathrm{pos}}$. Normalize it to $\widehat{\mathbf u}_{\mathrm{pos}}$ and normalize $P_i\widehat{\mathbf u}_{\mathrm{pos}}$ for the tangential direction, using zero when its projection vanishes. Repeat feasibility and enumeration; rank positive $s_T(\mathbf o)=\widehat{\mathbf u}_{\mathrm{pos}}^\top\Delta\widehat{\mathbf F}$ as above. A zero projection again removes tangential force/alignment requirements.

\paragraph{Separate correction candidate sets.}
Kinematic, orientation, and position corrections retain separate candidate sets. Each contact-guided candidate records $\mathcal C=(t_{\mathrm{peak}},\mathbf o,\{S_i,\boldsymbol\sigma_i\}_{i:o_i\neq0})$. Position correction uses $t_{\mathrm{pos}}$; orientation and kinematic corrections use $t_{\mathrm{fail}}$. Pose targets activate their respective branches; only contact-removal edits allow choosing kinematic-only or kinematic-plus-contact search spaces. Without pose targets, only kinematic candidates apply.

\subsubsection*{Real-world primitive optimization}

We select one ranked candidate from each active type and jointly optimize their primitives under one recovery score. The kinematic-only optimization ablation restricts this procedure to human-guided kinematic candidates, omitting contact-guided orientation and position candidates.

For joint $j$ and component $u\in\{\mathrm{kin},\mathrm{pos},\mathrm{ori}\}$,
\begin{equation}
    \delta a_j^{(u)}(t)
    = \sigma_j^{(u)} \sum_{k=1}^{K_{\mathrm{prim}}}
    A_{jk}^{(u)}
    \exp\!\left[
        -\frac{\left(t-\mu_{jk}^{(u)}\right)^2}
        {2\left(w_{jk}^{(u)}\right)^2}
    \right],
    \label{eq:supp_recovery_gaussian_primitives}
\end{equation}
Here $\sigma_j^{(u)}$ is the proposed sign, $A\geq0$ the peak magnitude, and $\mu$ and $w>0$ the center and standard deviation in control steps. We use $K_{\mathrm{prim}}=3$, typically at most four joints, and $3K_{\mathrm{prim}}n_j$ parameters per $n_j$-joint candidate. Active components' parameters are concatenated and corrections summed. For \passns, \sonicdownns, \trianglepassns, \fourfingerpassleapns, and \fourfingerpassns, centers lie in $[t_{\mathrm{peak}}-5,t_{\mathrm{peak}}+5]$ and widths in $[5,10]$. For \thumbaroundns, centers equal $t_{\mathrm{peak}}$, widths lie in $[2,5]$, and only $2K_{\mathrm{prim}}n_j$ parameters remain. Magnitudes lie in $[0.035,0.35]$ rad (approximately $[2,20]$ degrees).

CMA-ES~\cite{Hansen2001CompletelyDS} evaluates eight proposals per generation, each receiving one human recovery score in $[-1,1]$. Optimization allows 30 generations (240 trials) and stops as soon as a sampled proposal recovers the failure, possibly mid-generation; it usually succeeds within 10 generations. Highest-ranked contact proposals sufficed in our experiments. For the \passvone example (Fig.~\ref{fig:interactive_viewer_for_kinematic_correction_and_annotations}), the hand is unchanged, both pose targets are specified, and only the middle-finger contact is excluded. Both branches prioritize strengthening ring/little contacts ($s_T=+2.1681$, $s_R=+0.0270311$). Note that these directional scores are neither probabilities nor mutually comparable physical measurements.

\subsubsection*{Learning the correction policy}
\label{sec:corrective_policy_details}
After full-episode recovery, record observations $\mathbf o_t$, fixed-base actions $\mathbf a_t^{\mathrm{base}}$, and applied corrections $\delta\mathbf a_t^{\mathrm{rec}}$ to train
\begin{equation}
    \mathcal L_{\mathrm{rec}}(\xi)
    =\mathbb E_{\mathcal D_{\mathrm{rec}}}
    \left[
    w_t\left\|\pi_{\mathrm{rec},\xi}(\mathbf o_t,\mathbf a_t^{\mathrm{base}})-\mathbf y_t\right\|_2^2
    +\lambda_{\mathrm{rec}}\left\|\pi_{\mathrm{rec},\xi}(\mathbf o_t,\mathbf a_t^{\mathrm{base}})\right\|_2^2
    \right],
    \label{eq:supp_recovery_bc_objective}
\end{equation}
Targets $\mathbf y_t$ are clipped recorded corrections; $w_t$ weights intervention/non-intervention samples and $\lambda_{\mathrm{rec}}$ regularizes magnitude. Adding this learned correction to the fixed base yields the first-stage controller for full-episode data collection.

\subsection*{Model-based Policy Refinement}
\label{sec:model_based_refinement_supp}

We learn contact compensation inside MuJoCo Warp and joint-wise input corrections for remaining discrepancies, retaining its analytical backbone.

In this stage, we use the same contact-compensation and delta-action model architectures and training hyperparameters across tasks, with input and output dimensions adjusted for the robot embodiment. Policy refinement follows a common training schedule while retaining each task’s base-policy reward and simulation settings.

\subsubsection*{Contact compensation and trajectory tracking}

Contact models close the rigid-body equations of motion through a constitutive law $f_n=F(\delta,\dot{\delta})$, where $\delta$ and $\dot{\delta}$ denote penetration and penetration rate. A simple spring--damper model is
\[
f_n=[K\delta+B\dot{\delta}]_+,
\qquad [x]_+=\max(0,x),
\]
where $K$ and $B$ are force-law stiffness and damping coefficients, and clipping excludes tensile forces. More elaborate models incorporate Hertzian elasticity and Hunt--Crossley dissipation.

MuJoCo couples contact forces to the resulting contact accelerations through regularized soft constraints, solved jointly with other active system constraints. Geometry and friction parameters determine contact kinematics and admissible forces. For frictionless normal contact,
\[
f_n=\left[D_n(a_{\mathrm{ref},n}-a_n)\right]_+,
\qquad D_n=R_n^{-1}>0.
\]
Here, $a_n$ is the resulting normal acceleration. The relation is implicit because this acceleration depends on contact forces throughout the coupled system. The reference acceleration retains a spring--damper structure,
\[
a_{\mathrm{ref},c,i}=-k^{\mathrm{eff}}_{c,i}r_{c,i}-b_{c,i}v_{c,i},
\]
where $c$ and $i$ index contacts and constraint rows, and $r_{c,i}$ and $v_{c,i}$ are position residuals and relative velocities in constraint coordinates. The coefficients $k^{\mathrm{eff}}_{c,i}$ and $b_{c,i}$ are derived from user-specified solver parameters and specify restoring and dissipative responses at the acceleration level. The effective stiffness includes impedance weighting; frictional directions of elliptic contacts have $r_{c,i}=0$, giving a damping-only reference.

Positive diagonal regularization $\mathbf R\succ0$ makes the force-space objective strongly convex: its Hessian is $\mathbf A+\mathbf R\succ0$, with $\mathbf A=\mathbf J\mathbf M^{-1}\mathbf J^\top\succeq0$, where $\mathbf J$ is the constraint Jacobian and $\mathbf M$ is the system inertia matrix. This regularization also changes the constitutive response. On the compressive frictionless branch, $a_{\mathrm{ref},n}-a_n=R_nf_n$, so $R_n$ controls the acceleration mismatch associated with a contact force, and its inverse $D_n$ determines enforcement strength toward the reference. For contact constraint rows, $D_{c,i}=R_{c,i}^{-1}>0$.

If the unconstrained force minimizer of the coupled problem satisfies all force-admissibility constraints, including the normal-force and friction-cone constraints,
\begin{equation}
\mathbf f_c=\operatorname{diag}(\mathbf D_c)
(\mathbf a_{\mathrm{ref},c}-\mathbf a_c),
\label{eq:supp_model_contact_interface}
\end{equation}
where $\mathbf a_c$ is the resulting acceleration in the solver's contact coordinates. When the candidate force is inadmissible, the full constrained solve determines the response. For an isolated scalar constraint in the feasible regime, $a=w a_{\mathrm{ref}}+(1-w)a_{\mathrm{free}}$, with $w=AD/(1+AD)$, inverse effective mass $A>0$, and unconstrained acceleration $a_{\mathrm{free}}$. At fixed $A$ and $a_{\mathrm{ref}}$, increasing $D$ strengthens the response toward the reference; decreasing $D$ permits softer behavior. We learn corrections to both $\mathbf a_{\mathrm{ref},c}$ and $\mathbf D_c$ while retaining the analytical solver.

A shared network $(\mathbf u_{D,c},\mathbf u_{a,c})=g_\theta(\mathcal C_c,\mathbf a_{\mathrm{ref},c},\mathbf D_c)$ uses geometry, contact parameters, kinematics, and row features. Geometry is object-relative; solver features use constraint coordinates. It modifies
\begin{equation}
\begin{aligned}
\mathbf s_c&=\exp\!\left(\alpha_c\tanh(\mathbf u_{D,c})\right),\\
\widetilde{\mathbf D}_c&=\eta\,\mathbf s_c\odot\mathbf D_c,\\
\widetilde{\mathbf a}_{\mathrm{ref},c}
&=\mathbf a_{\mathrm{ref},c}+\beta_c\tanh(\mathbf u_{a,c}).
\end{aligned}
\label{eq:supp_model_contact_modulation}
\end{equation}
Element-wise nonlinearities preserve positive weights and bound acceleration residuals. Per-environment weight normalization over valid modified rows $\mathcal V$ is optional. In our experiments, normalization improves training stability, whereas unnormalized weights better fit contact responses. We therefore interpolate the normalization denominator:
\begin{equation}
\eta(\lambda)=\frac{\sum_{(c,i)\in\mathcal V}D_{c,i}}
{(1-\lambda)\sum_{(c,i)\in\mathcal V}s_{c,i}D_{c,i}
+\lambda\sum_{(c,i)\in\mathcal V}D_{c,i}},
\qquad \lambda\in[0,1].
\label{eq:supp_model_contact_normalization}
\end{equation}
During training, we gradually increase $\lambda$ from $0$ (full normalization) to $1$ ($\eta=1$, no normalization), relaxing the fixed-total-weight restriction. Empty selected sets receive no update. Corrections enter every physics-step constraint solve, which couples contact-wise predictions with all other constraints. The nominal bounds are $\alpha_c=0.7$ and $\beta_c=10.0$.

For real trajectories $\mathcal D_0^{\mathrm{real}}=\{\mathbf T_i\}$ with $\mathbf T_i=\{(\mathbf s_t^h,\mathbf s_t^o,\mathbf a_t)\}_t$, initialize recorded hand/object states and replay actions at control frequency $f_c$. Compensation updates at physics frequency $f_s$ and learns through RL tracking without correction labels~\cite{Hwangbo2019LearningAA,He2025ASAPAS}. Physics step $t_s$ uses $t_r=\lfloor t_sf_c/f_s\rfloor$:
\begin{equation}
r_{t_s}=w_h r_h(\hat{\mathbf s}_{t_s}^h,\mathbf s_{t_r}^h)
+w_o r_o(\hat{\mathbf s}_{t_s}^o,\mathbf s_{t_r}^o).
\label{eq:supp_model_tracking_reward}
\end{equation}
Hats denote simulation.
Reference time advances independently of tracking success and clamps at available frames. Reward/reset/termination details appear in Sec.~\nameref{sec:workflow_per_nerual_contact}.

\subsubsection*{Surrogate-based joint-wise delta-action learning}

After contact-based policy refinement, simulated trajectories $\mathcal D_1^{\mathrm{sim}}$ from the refined base alone train a differentiable surrogate; real trajectories $\mathcal D_1^{\mathrm{real}}$ from that base plus stage-one correction train delta actions. Real data are therefore not strictly on-policy for base-only simulation.

Normalize joints and targets by joint limits. State $\mathbf x_t=(\bar{\mathbf q}_t,\rho(\mathbf s_t^o))$ uses the unit object longitudinal axis $\boldsymbol\ell_t$:
\begin{equation}
\rho(\mathbf s_t^o)
=\left(\ell_{t,x}^2,\ell_{t,y}^2,\ell_{t,z}^2,
\ell_{t,x}\ell_{t,y},\ell_{t,x}\ell_{t,z},\ell_{t,y}\ell_{t,z}\right)^\top.
\label{eq:supp_model_object_encoding}
\end{equation}
This encoding is invariant to axis reversal/axial rotation and excludes translation and axial spin. Surrogate $F_\psi$ receives state/previous-command history $\mathcal H_t$ and command $\bar{\mathbf a}_t$ driving $t\to t+1$. Recurrent state resets per training window and persists through the $P=16$ recursively predicted transitions, $\hat{\mathbf x}_{t+k+1}=F_\psi(\hat{\mathcal H}_{t+k},\bar{\mathbf a}_{t+k})$. Histories use predicted states and applied commands; measured future states only supervise:
\begin{equation}
\begin{aligned}
\mathcal L_{\mathrm{wm}}(\psi)
&=\mathbb E_{\mathcal D_1^{\mathrm{sim}}}
\left[\sum_{k=1}^{P}d(\hat{\mathbf x}_{t+k},\mathbf x_{t+k})\right],\\
d(\hat{\mathbf x},\mathbf x)
&=\lambda_h\|\hat{\bar{\mathbf q}}-\bar{\mathbf q}\|_2^2
+\lambda_o\|\hat{\mathbf x}^{o}-\mathbf x^{o}\|_2^2.
\end{aligned}
\label{eq:supp_model_surrogate_loss}
\end{equation}

Separate joint models $J^i_{\phi_i}$ use local state--action history and nominal commands to predict bounded $\Delta\bar a_{t,i}$, converted by $\Delta a_{t,i}=(q_i^{\max}-q_i^{\min})\Delta\bar a_{t,i}/2$. Freeze $F_{\psi^\star}$ and train all joint models jointly through it. Initial history preserves measured states but replaces commands with compensated ones; subsequent histories recursively use predicted states and recomputed compensated actions:
\[
\hat{\mathbf x}_{t+k+1}^{\phi}=F_{\psi^\star}(\hat{\mathcal H}_{t+k}^{\phi},\bar{\mathbf a}_{t+k}+\Delta\bar{\mathbf a}_{t+k}(\phi)).
\]
\begin{equation}
\mathcal L_{\mathrm{delta}}(\phi)
=\mathbb E_{\mathcal D_1^{\mathrm{real}}}
\left[\sum_{k=1}^{P}d(\hat{\mathbf x}_{t+k}^{\phi},\mathbf x_{t+k})\right],
\qquad P=16.
\label{eq:supp_model_delta_loss}
\end{equation}
Only $\phi$ updates; gradients traverse the fixed surrogate, full horizon, and corrected initial action history, requiring no delta-action labels. Architectures, histories, and training settings appear in Sec.~\nameref{sec:workflow_delta_action_model}.

\subsubsection*{Refinement schedule} \label{sec:supp_methdo_refinement_scehdule}

We collect 200 real rollouts, including successful and failed trials, with base plus correction policy, to learn contact compensation. After that, we refine the base alone using PPO~\cite{Schulman2017ProximalPO} in the contact compensator-hybrid simulator. Then, we collect another 200 real rollouts, again including both outcomes, with refined base plus correction and 100,000 simulated rollouts with refined base alone. We then train delta actions and the surrogate, respectively. A second PPO refinement fixes both simulator compensators. Each refinement uses 1,000 PPO epochs, warm-starting the critic from initial training or the preceding refinement. 
The stage-one correction policy remains frozen throughout model-based refinement. In our observations, it produces large corrections around targeted critical failures and near-zero outputs at many other steps. These corrections are learned under real-world dynamics and need not improve simulated execution, so we exclude the correction policy from simulation and PPO during this stage. We nevertheless reuse it without retraining for real-world evaluation and subsequent real-world RL, because base-policy refinement mainly improves other aspects of execution and does not reliably resolve the targeted failures on its own.

\subsection*{Latent-Space Real-World Reinforcement Learning}
\label{sec:latent_realworld_rl_supp}

Here $\pi_{\mathrm{base}}$ denotes the fixed model-based refined base plus stage-one correction, evaluated at every control step. An observation-conditioned predictor, variational encoder, and flow decoder provide the representation for a separately learned latent modulation policy.

In this stage, we use the same SAC architecture and optimization hyperparameters, latent dimension, and residual scaling across tasks unless explicitly varied in ablations. Reward definitions and rollout horizons are task-specific. Exploration chunk lengths are selected from $\{4,8,16\}$ in simulation with $d_z=16$ fixed. Simulation-based selection and real-world optimization share the per-control-step bootstrap discount $\gamma_{\mathrm{step}}=0.99^{1/8}$.

\subsubsection*{Action data and reference prediction}
\label{sec:supp_latent_data}

For each task, we collect 200 real and 100,000 hybrid-simulation trajectories of the refined policy. 
For trajectories in the real-world, we keep all rollouts but keep transitions only before the termination (\emph{e.g.,} object dropping).
For trajectories in the simulation, we collect only successful trials. 
Rollouts give observation/chunk pairs $(\mathbf o_t,\mathbf A_t)$ with
\begin{equation}
\mathbf A_t=[\mathbf a_t,\ldots,\mathbf a_{t+T_c-1}]^\top\in\mathbb R^{T_c\times d_a}.
\label{eq:supp_latent_chunk}
\end{equation}
$T_c$ counts control steps and $d_a$ hand joints. 
Denote the sources $\mathcal D^{\mathrm{real,ac}}$ and $\mathcal D^{\mathrm{sim,ac}}$.
We use them to train the reference predictor and the latent space. 
And we sample from $\mathcal D^{\mathrm{real,ac}}$ and $\mathcal D^{\mathrm{sim,ac}}$ via 1:10.

\paragraph{Reference predictor.}
\label{sec:supp_latent_predictor}
Using the base-policy BC model's five-layer MLP and supervised procedure, with $T_cd_a$ outputs, we train the reference predictor via:
\begin{equation}
    \mathcal{L}_{\mathrm{BC}}(\omega)=
    \mathbb{E}_{(\mathbf{o}_t,\mathbf{A}_t)\sim\mathcal{D}_{\mathrm{pred}}}
    \left[
    \frac{1}{T_cd_a}
    \left\|\pi_\omega^{T_c}(\mathbf{o}_t)-\mathbf{A}_t\right\|_F^2
    \right],
    \label{eq:supp_latent_bc}
\end{equation}
where $\mathcal{D}_{\mathrm{pred}} = \mathcal D^{\mathrm{real,ac}}\cup \mathcal D^{\mathrm{sim,ac}}$.
At chunk boundaries, $\mathbf A_t^{\mathrm{ref}}=\pi_\omega^{T_c}(\mathbf o_t)$ supplies the modulation reference.

\subsubsection*{Variational encoding and flow decoding}
\label{sec:supp_latent_codec}

\begin{figure}[ht]
  \centering
  \includegraphics[width=\textwidth]{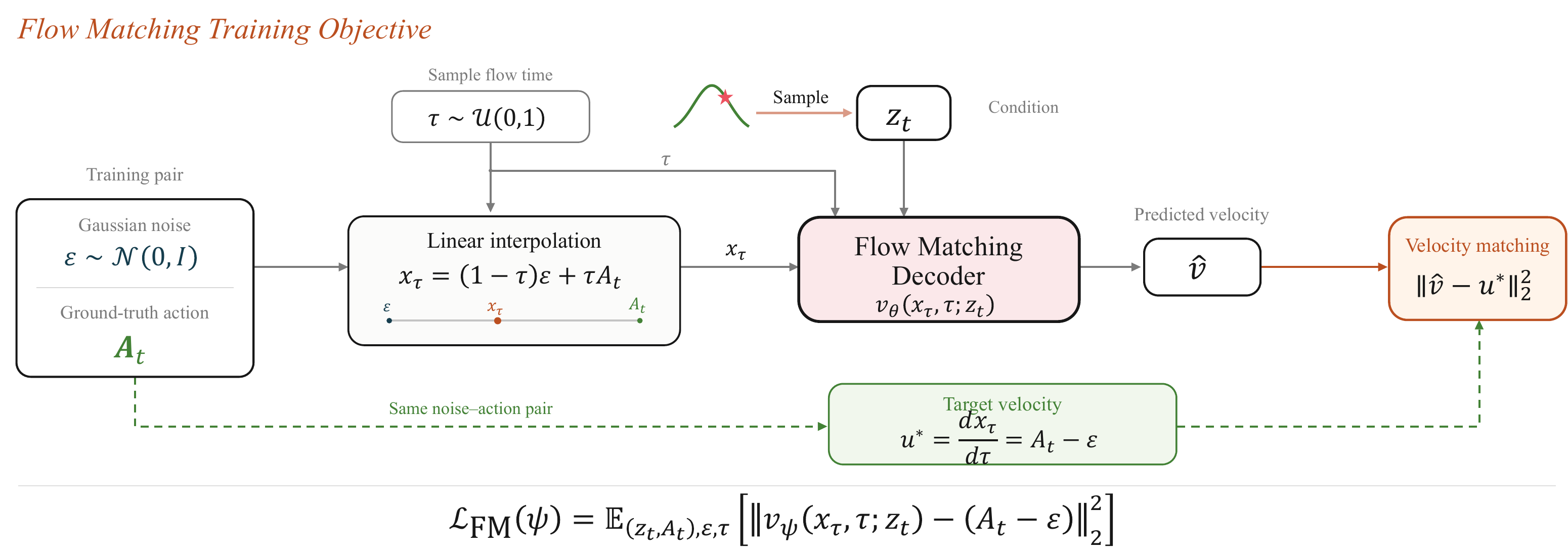}
  \caption{\footnotesize
  \textbf{Data Flow and Loss Computation for Training the Flow-Matching Decoder.} 
  }
  \label{fig:supp_method_flow_matching_training}
\end{figure}

Flattening/coordinate-wise standardization $\mathcal S$ gives $\mathbf x_1=\mathcal S(\mathbf A)\in\mathbb R^D$, $D=T_cd_a$, with $d_a=22$ (Sharpa) or $16$ (LEAP). Reference encoding uses the same statistics; decoding restores command coordinates. The encoder defines a diagonal Gaussian with log-variance $\boldsymbol\ell_\eta$ and reparameterized training samples:
\begin{equation}
    q_\eta(\mathbf{z}\mid\mathbf{x}_1)
    =\mathcal{N}\!\left(
    \boldsymbol{\mu}_\eta(\mathbf{x}_1),
    \operatorname{diag}\!\left[\exp\!\left(\boldsymbol{\ell}_\eta(\mathbf{x}_1)\right)\right]
    \right),
    \label{eq:supp_latent_posterior}
    \end{equation}
\begin{equation}
    \mathbf{z}
    =\boldsymbol{\mu}_\eta(\mathbf{x}_1)
    +\exp\!\left(\tfrac12\boldsymbol{\ell}_\eta(\mathbf{x}_1)\right)
    \odot\boldsymbol{\epsilon}_z,
    \qquad
    \boldsymbol{\epsilon}_z\sim\mathcal{N}(\mathbf{0},\mathbf{I}_{d_z}).
    \label{eq:supp_latent_reparameterization}
\end{equation}
Independently sample $\mathbf x_0\sim\mathcal N(\mathbf0,I_D)$ and $\tau\sim\mathcal U(0,1)$; let $\mathbf x_\tau=(1-\tau)\mathbf x_0+\tau\mathbf x_1$. Conditional flow matching~\cite{Lipman2022FlowMF} jointly trains encoder/decoder through reparameterization:
\begin{equation}
    \mathcal{L}_{\mathrm{FM}}(\eta,\psi)
    =\mathbb{E}
    \left[
    \frac{1}{D}
    \left\|\mathbf{v}_\psi(\mathbf{x}_\tau,\tau,\mathbf{z})
    -(\mathbf{x}_1-\mathbf{x}_0)\right\|_2^2
    \right].
    \label{eq:supp_latent_flow_matching}
\end{equation}
The expectation covers data, encoder samples, flow noise, and time. We further add prior regularization:
\begin{equation}
    \begin{aligned}
        \mathcal{L}_{\mathrm{model}}(\eta,\psi)
        &=\mathcal{L}_{\mathrm{FM}}(\eta,\psi)+\beta_{\mathrm{KL}}\mathcal{L}_{\mathrm{KL}}(\eta),\\
        \mathcal{L}_{\mathrm{KL}}(\eta)
        &=\mathbb{E}_{\mathbf{x}_1}
        \left[
        D_{\mathrm{KL}}\!\left(
        q_\eta(\mathbf{z}\mid\mathbf{x}_1)
        \,\middle\|\,\mathcal{N}(\mathbf{0},\mathbf{I}_{d_z})
        \right)
        \right],
    \end{aligned}
    \label{eq:supp_latent_codec_objective}
\end{equation}
Here $\beta_{\mathrm{KL}}=10^{-4}$.
Generation integrates
\begin{equation}
    \frac{d\mathbf{x}_\tau}{d\tau}
    =\mathbf{v}_\psi(\mathbf{x}_\tau,\tau,\mathbf{z}),
    \qquad
    \mathbf{x}_0=\boldsymbol{\epsilon},
    \quad
    \boldsymbol{\epsilon}\sim\mathcal{N}(\mathbf{0},\mathbf{I}_D).
    \label{eq:supp_latent_flow_ode}
\end{equation}
using $L=16$ uniform Euler steps, $\tau_\ell=\ell/L$ and $\mathbf x_{\tau_{\ell+1}}=\mathbf x_{\tau_\ell}+L^{-1}\mathbf v_\psi(\mathbf x_{\tau_\ell},\tau_\ell,\mathbf z)$, as in training.
$\mathbf D_\psi(\mathbf z;\boldsymbol\epsilon)$ includes integration, inverse standardization, and reshaping to $T_c\times d_a$. Flow time is distinct from control time (Fig.~\ref{fig:supp_method_flow_matching_training}).

\subsubsection*{Latent modulation and residual execution}
\label{sec:supp_latent_residual_execution}

At each training chunk boundary, we sample the reference latent and unscaled modulation:
\begin{equation}
\begin{aligned}
\mathbf{z}_t
&\sim q_\eta\!\left(\cdot\mid\mathcal{S}(\mathbf{A}^{\mathrm{ref}}_t)\right),\\
\Delta\mathbf{z}_t
&\sim\pi_{\mathrm{mod},\vartheta}(\cdot\mid\mathbf{o}_t,\mathbf{z}_t).
\end{aligned}
\label{eq:supp_latent_modulation}
\end{equation}
We then decode both with identical Gaussian initialization and integration:
\begin{equation}
    \begin{aligned}
    \widetilde{\mathbf{A}}_t
    &=\mathbf{D}_\psi(\mathbf{z}_t;\boldsymbol{\epsilon}_t),\\
    \widehat{\mathbf{A}}_t
    &=\mathbf{D}_\psi(\mathbf{z}_t+\beta_r\Delta\mathbf{z}_t;\boldsymbol{\epsilon}_t),\\
    \Delta\mathbf{A}_t
    &=\widehat{\mathbf{A}}_t-\widetilde{\mathbf{A}}_t,
    \end{aligned}
\label{eq:supp_latent_residual_chunk}
\end{equation}
We use  $\beta_r=0.25$.
Evaluation uses the posterior mean and one fixed initialization generated by \texttt{torch.randn} with a fixed seed.
Subtract decoded chunks in original command coordinates, not the predictor output. Execute
\begin{equation}
    \mathbf{a}^{\mathrm{exec}}_{t+k}
    =\pi_{\mathrm{base}}(\mathbf{o}_{t+k})
    +\alpha_r\,\Delta\mathbf{A}_t[k],
    \qquad k=0,\ldots,T_c-1,
\label{eq:supp_latent_executed_action}
\end{equation}
with $\alpha_r=0.25$ for all tasks. Modulation updates at $f_c/T_c$, consuming its precomputed residual one step at a time; the base controller recomputes from current feedback at $f_c$.

\paragraph{Reconstruction and identity.}
In Eq.~\eqref{eq:supp_latent_residual_chunk}, shared decoder noise makes zero modulation exactly recover the base controller regardless of reconstruction accuracy.

\subsubsection*{Base-controller preservation and residual action structure}
\label{sec:supp_latent_reachable_sets}

Let $\mathcal C_{\mathrm{base}}\subseteq\mathbb R^{T_cd_a}$ denote action chunks reachable by the unmodified base controller under the considered initial conditions and dynamics. For fixed $\alpha_r$ and $\beta_r$, let $\mathcal Z_{\alpha_r}$ denote chunks realizable under admissible latent offsets, including zero offsets, with base commands evaluated along the realized residual-augmented trajectories. These sets describe controller rollouts rather than the full actuator domain.

\paragraph{Preservation.}
Admissible zero offsets and shared decoder stochastic inputs imply
\begin{equation}
\mathcal C_{\mathrm{base}}\subseteq\mathcal Z_{\alpha_r}.
\label{eq:supp_latent_reachable_inclusion}
\end{equation}
\textit{Proof.} Shared decoder initialization makes every decoded difference zero when the latent offset is zero. Starting from the same state and coupling policy and environment randomness, actions and subsequent observations coincide by induction, recovering each original rollout. This holds for any fixed $\alpha_r$, regardless of reconstruction accuracy, provided that clipping and other action processing are identical. This is a representational property; it does not imply that stochastic SAC samples or learns every original rollout.

\paragraph{Residual action structure.}
Concatenate base commands evaluated along the residual-augmented trajectory into $\mathbf A_t^{\mathrm{base}}$:
\begin{equation}
\begin{aligned}
\mathbf A_t^{\mathrm{exec}}
&=\mathbf A_t^{\mathrm{base}}+\alpha_r(\widehat{\mathbf A}_t-\widetilde{\mathbf A}_t)\\
&=(\mathbf A_t^{\mathrm{base}}-\widetilde{\mathbf A}_t)
+(1-\alpha_r)\widetilde{\mathbf A}_t+\alpha_r\widehat{\mathbf A}_t.
\end{aligned}
\label{eq:supp_latent_chunk_decomposition}
\end{equation}
The feedback-driven base command enters with coefficient one and does not pass through the codec. Even $\alpha_r=1$ retains the difference between the base-command chunk and the reference decoded chunk. The decoder therefore parameterizes corrections rather than reconstructing complete control commands.

Earlier residuals alter subsequent observations and base actions, so $\mathbf A_t^{\mathrm{base}}$ is evaluated along the realized residual-augmented trajectory and need not belong to the unmodified rollout set. The available corrections depend on the reference latent, shared decoder initialization, admissible latent offsets, and action constraints. This analysis establishes exact base-controller recovery and residual composition, without asserting strict expansion of the trained decoder's output set.

\subsubsection*{Simulation warm-start and real-world optimization}
\label{sec:supp_latent_sac}

We pretrain the SAC actor and critics in simulation for 1,000 epochs using 8$\times$8,192 parallel environments with Eq.~\eqref{eq:supp_latent_executed_action}, then transfer them to hardware for continued optimization. We fix the base controller, predictor, codec, and standardization; only modulation actor/critics learn.
A chunk of $n_t\leq T_c$ executed steps receives $R_t=\sum_{k=0}^{n_t-1}r_{t+k}$ without within-chunk discounting; early termination shortens $n_t$. Its bootstrap discount is $\gamma_{\mathrm{step}}^{n_t}=0.99^{n_t/8}$; true terminal transitions disable bootstrapping. 
A motion-capture system supplies object pose for rewards/termination, not modulation-policy inputs.
Codec settings and RL protocols appear in Sec.~\nameref{sec:real_world_rl_exp_details}.

%% file: supp/supp_expsettings_4.tex
\section{Experimental Settings}
\label{sec:supp_exp_settings}

\subsection*{Hardware, Tasks, and Evaluation}
\begin{figure}[!htbp]
  \centering
  \includegraphics[width=0.65\textwidth]{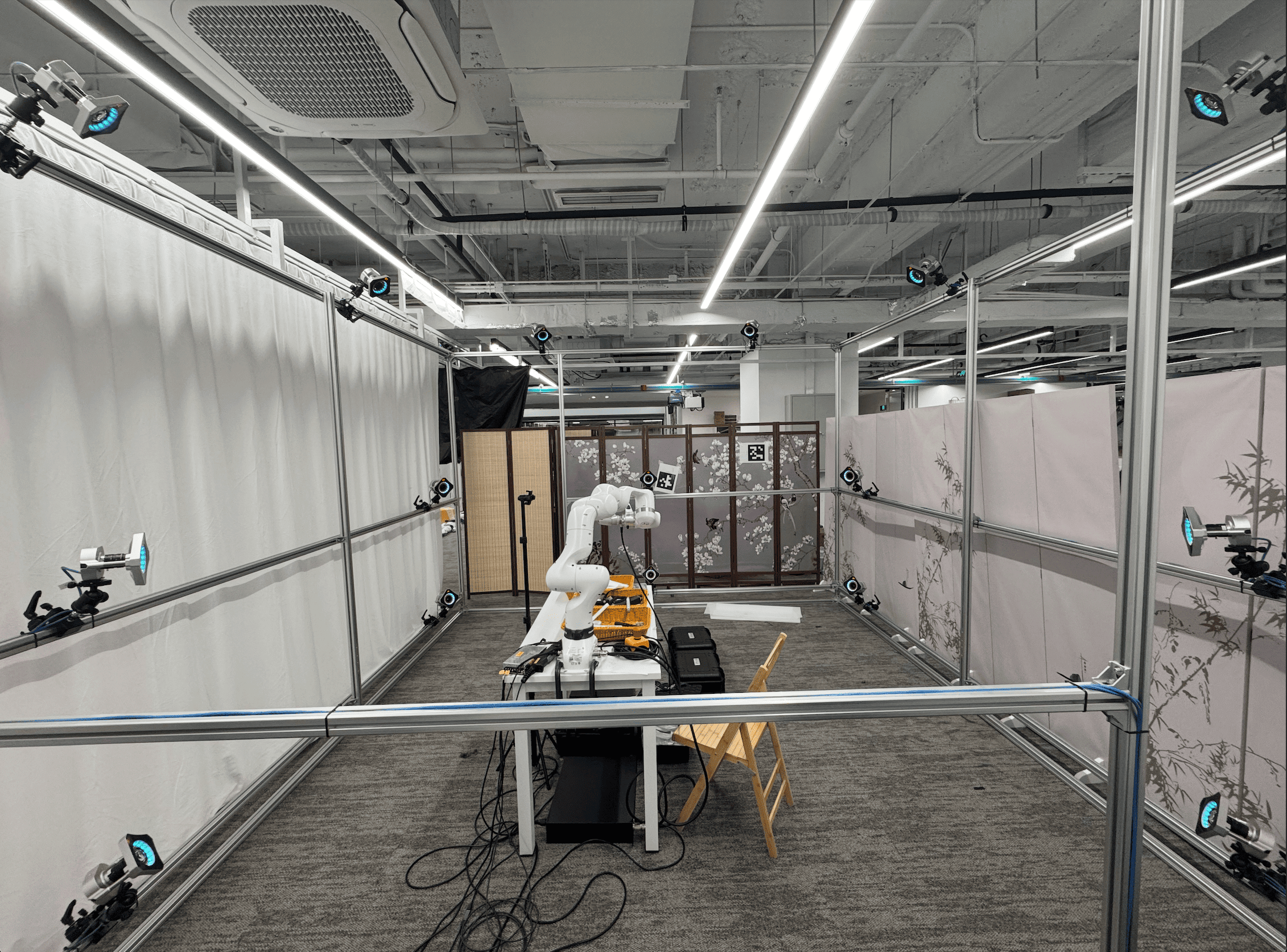}
  \caption{\footnotesize
  \textbf{Motion Capture Cage Setup}.
  }
  \label{fig:mocapcage_setup}
\end{figure}

\paragraph{Hardware and sensing.}
\label{para:supp_mocap_setup}
We use the right-handed Sharpa Wave W01 and LEAP Hand v1. A motion-capture system supplies termination signals and task-specific rewards, except that \thumbaround rewards are assigned from post-rollout human outcome labels  (Fig.~\ref{fig:mocapcage_setup}).
\textbf{Policies produced by our full pipeline do not require motion capture at deployment}.

\subsubsection*{Tasks and object configurations}
In \passvzerons, the pen starts vertically with the palm horizontal, traverses index--middle, middle--ring, and ring--little slots over approximately $2\pi$ radians, then reverses and retraces them (approximately $4\pi$ per episode). In \passvonens, the pen also starts vertically; the palm inclines $40^\circ$ to the ground and the index/middle fingers extend approximately horizontally (palm--finger opening approximately $140^\circ$); the pen returns dorsally without reversal ($4\pi$ per episode). In \passvtwons, the pen starts approximately parallel to the palm and follows the same dorsal return, requiring anti-tilting compensation on the outward transfer. \passvthree instead reverses for the return and requires tilt prevention throughout. \passvtwo uses a capsule of radius $0.7\,\mathrm{cm}$ with a $14\,\mathrm{cm}$ cylindrical section and hemispherical caps; the other three variants use a cylinder of radius $0.7\,\mathrm{cm}$ and length $20\,\mathrm{cm}$. In each variant, the simulated and physical objects have the same geometry.

\thumbaround launches and catches a pen after approximately $2\pi$ rotation around the thumb; its nominal rigid PLA cylinder has radius $0.7\,\mathrm{cm}$ and length $20\,\mathrm{cm}$, with TPU used for compliance adaptation. \sonicdown uses a radius-$0.54\,\mathrm{cm}$ pen for two downward slot transfers with conical pen-axis sweeps; reported speed measures axis motion.

\textit{Triangle Pass} uses elongated prisms with cross-sectional side lengths of $18$--$20\,\mathrm{mm}$ and length $15\,\mathrm{cm}$. For \trianglepassdownreversedns, corresponding joints of the middle, ring, and little fingers share commands.

\paragraph{Object-to-hand width ratio.}
We define $\rho=d_{\mathrm{obj}}/w_{\mathrm{index}}$, where $d_{\mathrm{obj}}$ is object diameter and $w_{\mathrm{index}}$ is the width of the index finger's first segment. The measured human finger is $11.6\,\mathrm{mm}$ wide: a $9\,\mathrm{mm}$ ballpoint pen gives $\rho=0.776$, and $10.6$--$10.8\,\mathrm{mm}$ spinning pens give $\rho=0.914$--$0.931$. For Sharpa, $w_{\mathrm{index}}=15.5\,\mathrm{mm}$; the $10.8$--$14\,\mathrm{mm}$ cylinders used in \sonicdownns, \passns, and \thumbaround yield $\rho=0.697$--$0.903$.

\subsubsection*{Evaluation protocols and metrics}
\label{sec:supp_evaluation_protocols}
\paragraph{Episode completion.}
A \pass episode traverses index--middle, middle--ring, ring--little, middle--ring, and index--middle slots, with variant-specific direction and return path. Success for \pass requires completing the cycle and returning to the initial slot at the end of the rotation episode. 
\thumbaround and \sonicdown  each use one complete spinning maneuver.
A \trianglepass or \fourfingerpass episode is a revolution that rotates the object for about $2\pi$.
Nominal-task success rates use 50 trials, with horizons set by the observed control steps for one episode.

\paragraph{Average and sustained rotation.}
Figure~\ref{fig:rollout_figure}(C, D) report mean $\pm$ SD of episode counts per rollout, including incomplete cycles. A human evaluator scores \pass progress from completed slot transitions. Table~\ref{tab:supp_evaluation_protocols} specifies counts and increments; these rollouts need not coincide with success-rate trials.

For \trianglepassdownreversedns, direct transfer, Stages 1–2, and the baselines except SERNF use 50 rollouts because these policies did not sustain more than one episode; Stage 3 uses five rollouts because it supports longer continuous rotation.
SERNF is also evaluated over five rollouts.

\begin{table}[htbp]
\centering
\small
\caption{\footnotesize
\textbf{Real-world evaluation counts and episode accounting.} All nominal tasks use 50 single-episode trials for success-rate evaluation. The counts below apply to average rotation; the two sets need not be identical. 
}
\label{tab:supp_evaluation_protocols}
\begin{tabular}{@{}lcl@{}}
\toprule
Task & Rollouts & Episode increments or values \\
\midrule
\passvzero & 20 & Increments of $0.25$ \\
\passvonens, \passvtwons, \passvthree & 50 per task & Increments of $0.25$ \\
\thumbaround & 50 & $0$ or $1$ \\
\sonicdown & 50 & $0$, $0.5$, or $1$ \\
\trianglepassupns, \trianglepassdownns & 5 per task & Increments of $0.25$ \\
\trianglepassdownreversedns & 50 / 5 & Increments of $0.25$ \\
\fourfingerpassleap & 50 & Increments of $0.25$ \\
\bottomrule
\end{tabular}
\end{table}

We report mean and best continuous runs over 20 \passvzero trials and five per \trianglepass variant, plus the longest demonstrations for other \pass variants. 
For \passvonens, \passvtwons, \passvthreens, and \fourfingerpassleapns, mean rotation is computed from 50 single-episode rollouts and is therefore at most one episode. The multi-episode maxima quoted in the main text come from separate continuous demonstrations.
Continuous runs used to calculate mean accumulated rotation continue until task failure, principally object dropping or jamming that prevents further rotation. Triangle Pass testing is limited by lengthy runs and actuator heating after 20--30 minutes, which may affect execution.
For \trianglepassupns, the policy completes $39.9\pm25.46$ episodes across five evaluation trials (mean$\pm$population SD). A separate demonstration, excluded from these statistics, reaches 91 consecutive episodes.

\paragraph{Mocap-based pen-axis angular speed.}
Reported pen-axis speeds are computed from motion-capture object trajectories. At mocap timestamp $t_k$, let $\mathbf c_k$ and $\mathbf e_k$ denote the object center and a consistently identified endpoint, defining $\mathbf u_k=(\mathbf e_k-\mathbf c_k)/\|\mathbf e_k-\mathbf c_k\|_2$. Over the evaluated motion interval, the mean angular speed in $\mathrm{rad/s}$ is
\begin{equation}
\bar\omega_{\mathrm{axis}}
=\frac{\sum_{k=1}^{N}\arccos\!\left[\operatorname{clip}(\mathbf u_{k-1}^{\top}\mathbf u_k,-1,1)\right]}{t_N-t_0}.
\end{equation}
This accumulates pen-axis angular travel through direction reversals while excluding spin about the pen's own axis.

\paragraph{Stage combinations in Fig.~\ref{fig:framework_stages}.}
S2 only follows the same two-iteration Stage~2 procedure as S1+S2 (200 real-world rollouts per iteration), but collects these rollouts with the direct-transfer policy.
Each extension of S1+S2 (to 2,400 and 4,400 rollouts) collects 2,000 further real-world rollouts, updates the contact compensator and delta-action models on them with the simulated dataset size unchanged, and then refines the policy.
S1+S3 omits the hybrid simulator: Stage~3 starts from the direct-transfer base policy with the stage-one correction policy, and its simulation warm-start uses the calibrated analytical simulator.

\paragraph{Time-cost estimation.}
For Fig.~\ref{fig:framework_stages}(C), human effort for each S2 or S3 collection phase is $H=N(t_{\mathrm{rollout}}+3)/3600$ hours per task, where $N$ is the rollout count, $t_{\mathrm{rollout}}$ is the measured rollout duration in seconds, and $3\,\mathrm{s}$ accounts for resetting. A 30-minute cooldown follows every 30 minutes of operation, with none after completion, giving hardware experiment time $T=H+0.5\max\{0,\lceil 2H\rceil-1\}$ hours. This rule is applied separately to each collection phase, including the independent 200-rollout representation-learning phase for settings with S3.

For settings including S1, we add the recorded corrective-optimization time and the full 3-hour budget for separate correction-policy data collection. The latter contributes 3 hours of human effort and 5.5 hours of hardware time per task: 3 hours of operation plus five 30-minute cooldowns. These S1 rollouts remain excluded from the rollout counts in Fig.~\ref{fig:framework_stages}. Totals sum the included phase estimates before averaging across ten tasks. The estimates are approximate upper bounds for the accounted phases and exclude computation for model training and policy optimization, including simulation training and real-world-data updates.


\paragraph{Rollout horizons for budget comparisons and sustained rotation.}
The budget comparisons in Fig.~\ref{fig:framework_stages} cap real-world training rollouts at 100 control steps (about 3.3 seconds at 30\,Hz) for \passvzero and at 400 control steps (20 seconds at 20\,Hz) for all three \trianglepass variants. This protocol already achieves 100\% single-episode success on these four tasks. Across the tasks in this comparison, rollout collection including reset takes at most 30 seconds per rollout. Training for sustained rotation instead uses the longer rollout limits (200 steps for \passvzero and 1,600 steps for \trianglepassns). These longer training rollouts require more hardware interaction time, which is not represented by the time estimates in Fig.~\ref{fig:framework_stages}(C).

\paragraph{Episode-level evaluation without resets.}
Evaluation units correspond to task-defined rotation episodes and may be assessed sequentially during continuous execution. 
For the adapted long-board condition, 45 episodes were assessed within a single uninterrupted run. This measures sustained execution rather than success across independently initialized trials; no binomial confidence interval is reported.

\paragraph{Training repetitions and evaluation.}
Each real-world learning configuration was trained once, and repeated evaluations were conducted using the resulting fixed controller. The reported success rates therefore characterize execution repeatability for these controllers; variability across independent real-world training runs was not measured. Repeated real-world training was constrained by the substantial human involvement, training time, and hardware wear associated with each run. Repetitions for simulation experiments are specified separately in the corresponding captions.

\paragraph{Independent re-scoring.}
The independent rater, a member of the authors' laboratories who was not involved in developing the method,
re-scored the single-episode evaluation videos of five of the ten tasks, selected at random: \thumbaroundns, \passvzerons, \passvtwons, \trianglepassupns, and \trianglepassdownreversedns. These comprise 250 trials, including 6 failures (one in \thumbaround and five in \passvtwons). The rater's scores matched the original scores on every trial (Cohen's $\kappa=1.00$).

\subsection*{System Identification for Direct Transfer}
\label{sec:supp_direct_transfer_sysid}
\label{sec:supp_actuator_sysid}
Our direct-transfer calibration has two scopes: task-agnostic free-hand calibration identifies actuator parameters and delays and learns an action compensator; per-task calibration identifies contact parameters from each task's own hand--object trajectories. The calibrated free-hand model is shared across tasks with matching hardware, mounting, and timing settings and held fixed during contact identification. 
The Sharpa and LEAP hands are identified separately with this protocol; the parameters and tracking results reported below are for the Sharpa hand.
This protocol is intended to reduce simulator mismatch before simulation-only policy training and deployment. Here, zero-shot policy transfer means deployment without real-world policy optimization; it does not exclude the hardware measurements and calibration data used to construct the simulator.

\paragraph{Task-agnostic physical identification.}
The actuator-calibration protocol uses free-space joint motions without an object or a task policy. Starting from a stationary hand posture, we excite individual joints with bounded, smooth bidirectional target sequences at multiple amplitudes and frequencies, while holding the remaining joint targets fixed. Each sequence includes an initial stationary interval and a recovery interval. We record the actual joint-position vectors submitted to the hardware interface and the encoder responses as separate, asynchronously timestamped streams. Host monotonic timestamps bracket every command and feedback call; repeated or out-of-order feedback frames are retained in the raw log but excluded from the fitting objective. Hardware control mode and speed/current settings are held fixed and recorded, since the identified parameters describe the effective closed-loop response under those settings.

We jointly search actuator parameters and a shared action/observation delay budget. The starting XML 
provides reference proportional gains $K_{p,j}^{0}$, derivative gains $K_{d,j}^{0}$, passive joint damping $b_j^{0}$, joint friction losses $f_j^{0}$, and equivalent rotor inertias $J_j^{0}$ (the MuJoCo \texttt{armature} field). Where the XML specifies a damping ratio, we compile the reference actuator model once to obtain $K_{d,j}^{0}$ before varying any parameters. These references come from the starting XML, including its reference actuator block, rather than from subsequently modified Python actuator presets. For each fitted joint group $g$, the candidate parameters are
\begin{equation}
\begin{split}
K_{p,j}&=s_{p,g}K_{p,j}^{0},\qquad
(K_{d,j},b_j)=s_{v,g}(K_{d,j}^{0},b_j^{0}),\\
f_j&=s_{f,g}f_j^{0},\qquad
J_j=s_{J,g}J_j^{0},\qquad s_{p,g},s_{v,g},s_{f,g},s_{J,g}\in[0.5,1.5].
\end{split}
\label{eq:supp_actuator_sysid_scales}
\end{equation}
The default grouping shares scales across corresponding joints of the four non-thumb fingers, with separate groups for the five thumb joints and pinky CMC joint. Only groups excited in the calibration data are fitted. A zero XML reference remains fixed at zero. A common viscous scale preserves the XML ratio of derivative and passive damping: encoder-position trajectories alone do not reliably distinguish these two velocity-dependent effects. Joint friction loss here is distinct from the geom contact friction fitted in the following subsection.

Small-step command-to-encoder measurements supply the response-time reference $\hat\tau = 11.27$ ms for identification. We applied positive and negative joint-position steps of 0.04 and 0.08 rad at command rates of 20, 30, and 100 Hz, with ten repetitions per condition, measured the time from command submission to the first fresh encoder sample showing the displacement, and took the arithmetic mean of the twelve condition medians.

For each candidate we choose $s_\tau\in[0.8,1.5]$ and $\rho\in[0,1]$, giving $\tau=s_\tau\hat\tau$, $\tau^a=\rho\tau$, and $\tau^o=(1-\rho)\tau$. With physics timestep $h$, we set $B=\operatorname{round}(\tau/h)$, $A=\operatorname{round}(\rho B)$, and $O=B-A$, so that action and observation delays consume $A$ and $O$ physical steps. The actuator parameters and delay pair are fixed across all trajectories evaluated for a candidate. The measured response is a prior on this effective delay budget, not a separate identification of transport latency: actuator dynamics and feedback sampling also contribute to the measured response.

For every candidate, we initialize the fixed-base simulation from the recorded stationary joint state and replay the actual submitted position targets open loop, with no policy feedback. Targets are held between their recorded submission times and applied on the physics grid after the candidate action delay. We compare simulated joint positions, shifted by the candidate observation delay, with fresh encoder samples at their host receipt times. The calibration records the hand-frame gravity vector and uses it in replay; free-space calibration disables contact. Candidate selection minimizes the mean squared position error, averaging samples and scored joints within each trajectory and then weighting trajectories equally:
\begin{equation}
\boldsymbol\theta^\star=\underset{\boldsymbol\theta\in\Theta_{\mathrm{eval}}}{\arg\min}
\frac{1}{N}\sum_{i=1}^{N}\frac{1}{M_i|\mathcal J_i|}
\sum_{k=1}^{M_i}\sum_{j\in\mathcal J_i}
\left[\hat q_{i,j}(t_{i,k}-Oh;\boldsymbol\theta)-q^{\mathrm{real}}_{i,k,j}\right]^2,
\label{eq:supp_actuator_sysid_objective}
\end{equation}
where $\mathcal J_i$ contains the excited joints, and samples from the initial stationary interval are excluded from scoring. Entire held-out trajectories can assess prediction error without contributing to candidate selection. As in contact-parameter identification, evaluated parameter combinations are ranked by replay-response error. The resulting values are effective model parameters; encoder-only data can leave multiple parameter combinations and action/observation allocations nearly indistinguishable.

\paragraph{Identified parameters and free-hand tracking accuracy.}
\textbf{Physical identification more than halved the free-hand replay trajectory error.} We fitted the physical parameters and delays using 44 trajectories and evaluated the selected model on 22 held-out trajectories, which did not contribute to parameter selection. A CMA-ES search used a budget of 1,024 candidate evaluations, including the reference candidate. Table~\ref{tab:supp_freehand_sysid_accuracy} reports the square root of the equally weighted trajectory MSE in Eq.~\ref{eq:supp_actuator_sysid_objective}, converted to degrees. On held-out trajectories, this RMSE decreased from $0.4003^\circ$ to $0.1716^\circ$, a \textbf{57.1\% reduction}. These errors compare simulated and measured joint-position responses to the same recorded commands. The improvement was obtained by fitting physical parameters and delays, before learning the action compensator described below.

\begin{table}[htbp]
\centering\small
\setlength{\tabcolsep}{4pt}
\caption{\footnotesize \textbf{Free-hand trajectory agreement improves substantially after joint identification of physical parameters and delays.} The reference uses the starting XML parameters with action and observation delays of $6\,\mathrm{ms}$ and $5\,\mathrm{ms}$, respectively; the identified model uses $6\,\mathrm{ms}$ and $3\,\mathrm{ms}$. Both models are evaluated on the same trajectories within each split, before learned actuator compensation.}
\label{tab:supp_freehand_sysid_accuracy}
\resizebox{0.4\textwidth}{!}{%
\begin{tabularx}{0.5\linewidth}{@{}Xrrr@{}}
\toprule
Trajectories & \shortstack{Reference\\RMSE ($^\circ$)} & \shortstack{Identified\\RMSE ($^\circ$)} & Reduction\\
\midrule
Training (44) & 0.3976 & \textbf{0.1685} & \textbf{57.6\%}\\
Held-out (22) & 0.4003 & \textbf{0.1716} & \textbf{57.1\%}\\
\bottomrule
\end{tabularx}
}
\end{table}

Table~\ref{tab:supp_identified_hand_parameters} lists the identified absolute parameter values. Using the cross-condition response-time reference $\hat\tau=11.27\,\mathrm{ms}$, the fitted scale $s_\tau=0.823279$ and action fraction $\rho=0.671893$ yield $A=6$ action-delay steps and $O=3$ observation-delay steps. At a $1\,\mathrm{ms}$ physics timestep, these correspond to $6\,\mathrm{ms}$ and $3\,\mathrm{ms}$, respectively, for a total delay of $9\,\mathrm{ms}$. These are effective fitted delays; their one-way split is not independently identified. Eight parameter scales lie near the search bounds, so the reported parameters describe the best evaluated model within the specified search range.
The calibrated direct-transfer tests, the 
base policies for Stages 1--3 and the hybrid-simulator analyses (Fig.~\ref{fig:model_based_abl} and Supplementary Section~S3) use the identification of the corresponding hand.
The ``Direct'' entries in Figs.~\ref{fig:rollout_figure},~\ref{fig:framework_stages}, and~\ref{fig:human_takeover_failures} report the base policies that enter the first stage.



\begin{table*}[t]
\centering\small
\caption{\footnotesize \textbf{Identified absolute free-hand parameters, rounded for display.} $K_p$ is in $\mathrm{N\,m/rad}$, $K_d$ in $\mathrm{N\,m\,s/rad}$, and joint friction loss $f$ in $\mathrm{N\,m}$. Passive damping $b$ and equivalent rotor inertia $J$ are displayed in units of $10^{-6}\,\mathrm{N\,m\,s/rad}$ and $10^{-3}\,\mathrm{kg\,m^2}$, respectively. I/M/R denotes the index, middle, and ring fingers; their corresponding parameter values coincide at the displayed precision. $K_d$ and $b$ share a fitted multiplier.}
\label{tab:supp_identified_hand_parameters}
\begin{tabular}{@{}lrrrrr@{}}
\toprule
Joint & $K_p$ & $K_d$ & $b$ ($10^{-6}$) & $f$ & $J$ ($10^{-3}$)\\
\midrule
Thumb CMC FE & 8.21345 & 0.308439 & 45.6649 & 0.0795785 & 4.43069 \\
Thumb CMC AA & 15.7657 & 0.533755 & 55.7891 & 0.153281 & 3.84568 \\
Thumb MCP FE & 7.0424 & 0.214359 & 25.0655 & 0.0447537 & 1.34228 \\
Thumb MCP AA & 7.74227 & 0.120896 & 11.9861 & 0.0565785 & 2.16578 \\
Thumb IP & 1.34983 & 0.0221548 & 2.14756 & 0.0065554 & 0.597928 \\
I/M/R MCP FE & 7.10022 & 0.106176 & 12.1625 & 0.0484308 & 3.71585 \\
I/M/R MCP AA & 8.05764 & 0.122656 & 11.9117 & 0.0604535 & 3.97208 \\
I/M/R PIP & 1.34651 & 0.0243651 & 2.33858 & 0.00654716 & 0.755515 \\
I/M/R DIP & 1.3299 & 0.0361674 & 1.24895 & 0.00217895 & 0.221032 \\
Pinky CMC & 0.690039 & 0.0199702 & 30.7957 & 0.0116033 & 0.173518 \\
Pinky MCP FE & 7.10022 & 0.106157 & 12.1625 & 0.0484308 & 3.71585 \\
Pinky MCP AA & 8.05764 & 0.122574 & 11.9117 & 0.0604535 & 3.97208 \\
Pinky PIP & 1.34651 & 0.0243647 & 2.33858 & 0.00654716 & 0.755515 \\
Pinky DIP & 1.3299 & 0.0361674 & 1.24895 & 0.00217895 & 0.221032 \\
\bottomrule
\end{tabular}
\end{table*}

\paragraph{Task-agnostic learned actuator compensation.}
After selecting the physical parameters $\boldsymbol\theta^\star$, we fix them and their fitted delays and collect a large set of task-agnostic, free-space simulated joint trajectories. We also collect new real free-hand trajectories for learning and evaluating the action compensator, separately from the physical-identification dataset. A differentiable world model $F_\psi$ predicts the next encoder-equivalent joint state from joint-position and submitted-target histories. A separate delta-action model $g_\phi$ predicts bounded joint-target corrections. Both networks are MLPs with hidden widths $(512,256,128)$ and have no object or contact inputs. Histories and targets are normalized using joint limits. Real-data inputs use only encoder feedback received before each command, with acquisition cadence matched to the model and large gaps excluded. Trajectories are split before constructing overlapping history windows.

We first train $F_\psi$ on simulated responses, then freeze its weights and train $g_\phi$ on free-hand real responses by backpropagating joint-state error through the fixed world model:
\begin{equation}
\begin{aligned}
\widetilde{\mathbf u}_t &= \mathbf u_t+g_\phi(\mathcal H_t^q,\mathcal H_t^u),\\
\widehat{\mathbf q}_{t+1} &= F_{\psi^\star}(\mathcal H_t^q,\mathcal H_t^{\widetilde u}),\\
\mathcal L_{\mathrm{act}} &= \mathbb E_{\mathcal D_{\mathrm{free}}^{\mathrm{real}}}
\!\left[\frac{1}{P}\sum_{k=1}^{P}\|\widehat{\mathbf q}_{t+k}-\mathbf q_{t+k}^{\mathrm{real}}\|_2^2\right]
+\lambda_\Delta\mathbb E\|\Delta\mathbf u\|_2^2.
\end{aligned}
\label{eq:supp_freehand_actuator_compensation}
\end{equation}
Corrections are bounded relative to each joint's range, and corrected targets respect joint limits. The world-model history contains corrected targets, including its initial action history; subsequent predictions roll forward recursively. The training horizon spans the fitted delay so that corrections receive gradients from their later effects. The frozen world model remains differentiable with respect to these targets. Held-out trajectories evaluate joint-state prediction, and direct MuJoCo replay checks compensation against the physical simulator rather than only its learned surrogate.

The resulting free-hand model consists of the identified physical parameters, fitted action/\allowbreak observation delays, and the fixed learned action compensator. During subsequent simulation, the compensator modifies absolute joint targets before the physical-step action-delay queue and actuator dynamics, with the same observation timing used for training. This simulator-side correction makes the simulated response approach the response to the recorded real command. It requires no task reward or object trajectory, and its fixed weights are shared across compatible tasks. The world model provides training gradients, while policy simulation retains the analytical dynamics and learned action compensator. Per-task contact identification below and subsequent task-dependent hybrid-simulator refinement (Sec.~\nameref{sec:workflow_delta_action_model}) are separate stages.

\paragraph{Free-hand compensation results.}
\textbf{The learned action compensator further improves free-hand tracking.} On evaluation trajectories from the newly collected real free-hand dataset, adding the learned delta-action compensator to the identified physical model reduced joint-position RMS error from $3.27\,\mathrm{mrad}$ ($0.187^\circ$) to $2.70\,\mathrm{mrad}$ ($0.155^\circ$), a \textbf{17.4\% reduction}. The $0.187^\circ$ uncompensated baseline is measured on this dataset, whereas the $0.1716^\circ$ result in Table~\ref{tab:supp_freehand_sysid_accuracy} uses the held-out physical-identification trajectories.

\paragraph{Per-task contact-parameter identification.}
\label{sec:supp_contact_sysid}
The contact-identification protocol is applied independently to each manipulation task $k$. It starts with an initial policy trained using default XML contact parameters and specifies a separate collection budget of 100 real-world rollouts of 50 control steps per task. The resulting task-specific dataset $\mathcal D_k$ records hand states, commanded actions, and object poses. 
Initial-policy training and candidate replay use the calibrated free-hand model of the corresponding hand.
We search only the first two components of geom-level \texttt{friction}, \texttt{solref}, and \texttt{solimp}. The actuator parameters, fitted delays, and compensator weights remain fixed, together with the task's geometry, other physical parameters, and unsearched contact components. Parameter randomization is disabled during candidate comparison. The common search ranges in Table~\ref{tab:supp_contact_sysid_candidates} yield 9,000 possible combinations per task.

\begin{table}[htbp]
\centering\small
\caption{\footnotesize \textbf{Candidate geom parameters searched independently for each task.} Indices are zero-based; unlisted contact components remain at their default XML values.}
\label{tab:supp_contact_sysid_candidates}
\begin{tabularx}{0.6\linewidth}{@{}lX@{}}
\toprule
Geom field & Candidate values\\
\midrule
\texttt{friction[0]} & 1.0, 1.5, 2.0, 2.5, 3.0\\
\texttt{friction[1]} & 0.001, 0.005\\
\texttt{solref[0]} & 0.001, 0.002, 0.003, 0.004, 0.005\\
\texttt{solref[1]} & 0.90, 0.92, 0.94, 0.96, 0.98, 1.00\\
\texttt{solimp[0]} & 0.95, 0.96, 0.97, 0.98, 0.99\\
\texttt{solimp[1]} & 0.99, 0.991, 0.992, 0.993, 0.994, 0.995\\
\bottomrule
\end{tabularx}
\end{table}

Within each task, we sample candidate combinations and evaluate them in parallel environments, holding one combination fixed per environment. For each candidate, we initialize from that task's recorded hand--object states and replay its recorded action sequences open loop through the fixed hand model. Recorded commands are held between submission times; the fixed compensator receives only simulated delayed joint states. Object poses are compared at their aligned capture times, independently of the hand encoder delay. Initial histories are backfilled from the recorded initial state, with a common history/delay warm-up excluded from scoring. We select a separate contact-parameter vector for task $k$ by minimizing its mean object-orientation trajectory error:
\begin{equation}
\boldsymbol\theta_{c,k}^\star
=\underset{\boldsymbol\theta_c\in\Theta_{c,k}^{\mathrm{eval}}}{\arg\min}\;
\frac{1}{N_k}\sum_{i=1}^{N_k}\frac{1}{|\mathcal T_{k,i}|}\sum_{t\in\mathcal T_{k,i}}
d_R\!\left(\hat R_{k,i,t}(\boldsymbol\theta_c),R_{k,i,t}^{\mathrm{real}}\right),
\label{eq:supp_contact_sysid_objective}
\end{equation}
where $N_k$ is the number of fitting trajectories, $\mathcal T_{k,i}$ is a fixed set of valid post-warm-up samples, and $\Theta_{c,k}^{\mathrm{eval}}$ contains the combinations evaluated for task $k$. The simulated and recorded orientations are compared with the shortest geodesic angular error $d_R$ in radians, treating quaternion signs equivalently. Trajectories receive equal weight; task success and position error do not enter candidate selection. Replays use a fixed horizon without success-based termination or measured-state correction. Entire validation trajectories are scored only after selecting the candidate from fitting trajectories. Each task retains its own dataset, search record, and selected contact parameters $\boldsymbol\theta_{c,k}^\star$; the search ranges and compatible free-hand calibration are shared.

\paragraph{Tracking performance after identification.}
To assess whether improved actuator modeling also improves object motion, we replayed recorded real-world commands open loop on 20 \passvzero trajectories with an object, starting from the measured initial hand--object states. Using whole-trajectory statistics in the identified simulator, enabling the learned free-hand compensator reduced joint-position RMSE only slightly, from $3.93^\circ$ to $3.90^\circ$ ($0.8\%$), while object-orientation error was $4.21^\circ$ without the compensator and $4.25^\circ$ with it.

On these replays, the small joint-tracking improvement was not accompanied by improved object tracking. More broadly, manipulation replays using the calibrated actuator model, fitted delays, learned action compensator, and task-specific identified contact parameters could still exhibit object sticking or drops instead of following the recorded motion. These remaining trajectory mismatches motivate the per-contact modeling described in Sec.~\nameref{sec:workflow_per_nerual_contact}.

\paragraph{Direct transfer with the calibrated simulator.}
After task-agnostic identification of actuator parameters and action/observation delays, learned free-hand compensation, and task-specific contact identification, we trained policies in the calibrated simulator using either proprioceptive histories alone or proprioceptive histories augmented with object-state histories.
\textbf{Neither input formulation achieved successful task completion on any of the ten tasks in our direct-transfer tests} (0/50 evaluation trials per task; Wilson 95\% CI 0--7.1\%). Policies trained with the default configuration, which randomizes actuator gains and delays and omits the learned free-hand compensator, also failed in direct transfer (0/50 trials per task).

\paragraph{Sim-to-real transfer from other simulators.}
\label{sec:supp_other_simulators_transfer}
Additional direct-transfer tests using Isaac Lab are reported in Supplementary Section~S3.

\paragraph{Task difficulty and direct transfer.}
Task difficulty and contact conditions offer a plausible explanation for why the public Wuji-MJLab and TeleDexter~\cite{Li2026TeleDexter} demonstrations transfer successfully while our direct-transfer tests fail. Their demonstrated rotation patterns appear qualitatively more forgiving: Wuji-MJLab's glove covering and pen surface treatment, and TeleDexter's extensive tape wrapping on the pen and finger surfaces below the fingertips, can enhance grip and reduce slipping and dropping. Our tasks repeatedly rely on precarious two-finger support and rapid contact transitions, and include underactuated object rotation in \thumbaroundns, without continuous stabilization by a multi-finger grasp. These demands can amplify the effects of errors in friction, contact response, and actuation timing, yielding a larger sim-to-real performance gap even for comparable modeling inaccuracies. This task-dependent sensitivity helps explain why direct transfer remains difficult despite the calibration above; Supplementary Section~S3 details the qualitative comparison.

\subsection*{Base-Policy Training in Simulation}
\subsubsection*{Simulator, actuation, and randomization}
We train in MuJoCo Warp on eight RTX 4090 GPUs with 8,192 environments per GPU. Table~\ref{tb:per_task_simulation_settings} gives control timing; Table~\ref{tab:simulation-settings} gives solver settings. Both hands use built-in position actuators. The fixed-base, 22-DoF Sharpa XML specifies link masses, inertias, collision geometry, and reference joint parameters; simulation actuators are constructed programmatically. 
Table~\ref{tab:sharpa_pd_gains} gives the XML proportional-gain references used by the actuator-calibration protocol, rather than fitted values.






\begin{table*}[ht]
\centering
\caption{\footnotesize \textbf{Per-task Simulation Settings. } }
\label{tb:per_task_simulation_settings}
\resizebox{1.0\textwidth}{!}{%
\begin{tabular}{l|cccccccccc}
\toprule
\multirow{1}{*}{Parameter} & \textbf{\passvzero} & \textbf{\passvone} & \textbf{\passvtwo}  & \textbf{\passvthree} & \textbf{\thumbaround} & \textbf{\sonicdown} & \textbf{\makecell[c]{\textit{Triangle Pass} \\ \texttt{(palm-up)}}} &  \textbf{\makecell[c]{\textit{Triangle Pass} \\ \texttt{(palm-down)}}} &  \textbf{\makecell[c]{\textit{Triangle Pass} \\ \texttt{(palm-down, reversed)}}} &  \textbf{\makecell[c]{\texttt{Four-Finger-Pass} \\ \texttt{(palm-down, LEAP)}}} \\
\midrule

Simulation Timestep (s) & $0.001$   &  $0.001$  & $0.001$ & $0.001$  &  $0.001$ & $0.001$ & $0.001$  & $0.001$  & $0.001$  & $0.001$   \\

Decimation  & 33 & 33 & 33 & 33 &
    50 &
    10 &
    50 & 50 & 50 & 50 \\

Control Frequency (Hz) & $\approx$30 (30.3) & $\approx$30 (30.3) & $\approx$30 (30.3) & $\approx$30 (30.3) &
    20 & 100 &
    20 & 20 & 20 & 20 \\

\bottomrule
\end{tabular}
}
\end{table*}

\begin{table}[htbp]
\centering\small
\caption{\footnotesize
\textbf{Simulation/contact-solver settings (left) and starting XML Sharpa proportional-gain references for identification (right).} The reference actuator block uses damping ratios $\zeta=0.9$; the calibration tool compiles this block once to obtain reference derivative gains before searching parameter scales. These entries are not numerical identification results.}
\label{tab:simulation-settings}\label{tab:sharpa_pd_gains}
\begin{minipage}[t]{0.46\linewidth}
\centering
\begin{tabularx}{\linewidth}{@{}Xr@{}}
\toprule
Solver parameter & Value\\\midrule
Friction cone & elliptic\\
\texttt{iterations} & 100\\
\texttt{ls\_iterations} & 50\\
\texttt{impratio} & 1\\
Gravity & $(0,0,-9.81)\,\mathrm{m\,s^{-2}}$\\
\texttt{nconmax} / \texttt{njmax} & 128 / 2000\\
\bottomrule
\end{tabularx}
\end{minipage}\hfill
\begin{minipage}[t]{0.50\linewidth}
\centering
\begin{tabularx}{\linewidth}{@{}Xr@{}}
\toprule
Sharpa joint(s) & $K_p$\\\midrule
Thumb CMC FE / CMC AA & 6.95 / 13.20\\
All MCP FE / MCP AA & 4.76 / 6.62\\
Thumb IP; other PIP/DIP & 0.90\\
Pinky CMC & 1.38\\
\bottomrule
\end{tabularx}
\end{minipage}
\end{table}

\paragraph{Default and calibrated training randomization.}
Table~\ref{tab:mjlab-randomization} lists the default randomization settings before calibration. Physical parameters are sampled per environment at startup and retained across resets. In this default configuration, proportional and derivative gains use independent $\mathcal U(0.8,1.2)$ multipliers relative to their nominal values, sampled separately for each actuator and environment.

For the calibrated policy-training runs reported above, the identified hand parameters ($K_p$, $K_d$, passive damping, joint friction loss, and armature), fitted action/observation delays, and learned free-hand compensator weights are held fixed throughout training. We do not randomize actuator parameters or delays around their identified values. Contact randomization instead uses multiplicative factors around each task's identified contact parameters: for each randomized component $j$ in environment $e$,
\[
\theta_{c,k,e,j}=\theta_{c,k,j}^{\star}s_{e,j},\qquad s_{e,j}\sim\mathcal U(0.8,1.2).
\]
This applies to sliding and torsional friction and the identified components of \texttt{solref} and \texttt{solimp} (the first two components of each); factors are sampled at startup and retained across resets. Unsearched contact components remain fixed. Thus, calibrated training does not use the default absolute sliding-friction range $\mathcal U(0.3,1.5)$. The $1.0$--$3.0$ sliding-friction candidates in Table~\ref{tab:supp_contact_sysid_candidates} are the identification search grid, not the policy-training distribution. Object mass/inertia, reset perturbations, and actor observation noise retain the settings in Table~\ref{tab:mjlab-randomization}. For \sonicdownns, training additionally randomizes the object radius within $[0.5,0.55]\,\mathrm{cm}$.

Only \pass and \sonicdown use reset position perturbations; orientation and initial velocities are fixed. LEAP uses grasp-bank initialization without object-pose perturbations.

Training adds componentwise uniform actor noise; critic observations are uncorrupted. Rollout collection for distillation, dynamics learning and latent-space training retains physical/reset randomization but disables actor noise. Formal simulation evaluation fixes physical parameters and disables reset randomization.

\begin{table}[htbp]
\centering
\small
\renewcommand{\arraystretch}{1.18}
\setlength{\tabcolsep}{5pt}
\caption{\footnotesize
\textbf{Default randomization and delay settings.} $\mathcal{U}$ and $\LogU$ denote uniform and log-uniform distributions, respectively; $\ell_i$ and $u_i$ are the soft limits of joint $i$. $\hat\tau$ denotes the measured command-to-encoder response reference. Action and sensor-observation delays share one randomized total and are jointly discretized in physics steps. The learned free-hand-compensator configuration instead retains its calibrated hand parameters and fixed delay pair.}
\label{tab:mjlab-randomization}
\resizebox{0.9\textwidth}{!}{%
\begin{tabularx}{\linewidth}{@{}p{0.29\linewidth}Xp{0.17\linewidth}@{}}
    \toprule
    Parameter & Distribution or initialization & Sampling / application time \\
    \midrule
        Object mass and inertia scale & $s\sim\LogU(0.5,2.0)$; $(m,\mathbf{I})=s(m_0,\mathbf{I}_0)$ & Once at startup \\
        Proportional gain scale & $s_{p,i}\sim\mathcal U(0.8,1.2)$; $K_{p,i}=s_{p,i}K_{p,i}^{0}$ & Once at startup \\
        Derivative gain scale & $s_{d,i}\sim\mathcal U(0.8,1.2)$; $K_{d,i}=s_{d,i}K_{d,i}^{0}$ & Once at startup \\
        Sliding friction: object and fingertips & $\mu_s\sim\mathcal{U}(0.3,1.5)$ & Once at startup \\
        Torsional friction: object and fingertips & $\mu_t\sim\LogU(10^{-4},2\times10^{-2})\,\mathrm{m}$ & Once at startup \\
        Rolling-friction field: object and fingertips$^{\dagger}$ & $\mu_r\sim\LogU(10^{-5},5\times10^{-3})\,\mathrm{m}$ & Once at startup \\
        Initial object $y$ and $z$ (\pass and \sonicdown only) & $y=y_0+\epsilon_y$, $z=z_0+\epsilon_z$ around the task's nominal pose; independent $\epsilon_y,\epsilon_z\sim\mathcal{U}(-0.01,0.01)\,\mathrm{m}$ & Every reset \\
        Default initial hand joint angles & $q_i=0.2\,\xi_i$, $\xi_i\sim\mathcal{U}(\ell_i,u_i)$; task-specific grasp initializations override this rule & Every reset \\
        Actor observation noise$^{\S}$ & $\epsilon_k\sim\mathcal{U}(-0.01,0.01)$, added independently in each observation component's units or representation & Each observation during training \\
        Total delay  & $\tau_t=s_t\hat\tau$, $s_t\sim\mathcal U(0.8,1.5)$ & Per environment, each control step \\
        Action delay & $\rho_t\tau_t$, $\rho_t\sim\mathcal U(0,1)$ & Per environment, each control step \\
        Sensor-observation delay & $(1-\rho_t)\tau_t$ & Per environment, each control step \\
    \bottomrule
\end{tabularx}
}
\vspace{4pt}
\begin{minipage}{\linewidth}
    \footnotesize
    $^{\dagger}$RigidContact uses \texttt{condim=4}: sliding and torsional friction are active, whereas the sampled rolling coefficient does not contribute rolling resistance. The object geom has higher contact priority than the hand geoms, so its coefficients determine dynamically generated hand--object contact friction.

    \smallskip
    $^{\S}$Noise affects joint positions, joint targets, object endpoints, goal endpoints, object velocity, hand geometry positions, and contact observations.
\end{minipage}
\end{table}

\subsubsection*{Behavior-cloned base policies}
\label{sec:supp_base_bc_training}
For the proprioception-only policies used in our subsequent pipeline, we distill simulated controller trajectories into a deterministic ResMLP using ten-frame histories of joint positions and issued joint targets, flattened separately in chronological order. All controlled joints, including the thumb, are retained without masking. No object observations are provided: the $3H$ object-motion conditioning inputs are set to zero.
For $n_{\mathrm{dof}}$ joints and prediction horizon $H$, the input and output dimensions are $20n_{\mathrm{dof}}+3H$ and $Hn_{\mathrm{dof}}$, respectively ($H=1$).
Joint positions and absolute-target actions remain in radians, without input/output normalization. 

The network has an input projection to width 512, five width-512 residual blocks, and a linear output head. Each block contains two main-path linear layers and a learned linear shortcut, with ReLU after the first main-path layer and after residual addition; no layer normalization or output squashing is used. We minimize the batch-mean summed squared error over predicted targets,
\begin{equation}
    \mathcal L_{\mathrm{BC,base}}
    =\frac{1}{B}\sum_{b=1}^{B}
    \left\|\widehat{\mathbf A}_b-\mathbf A_b\right\|_F^2.
    \label{eq:supp_base_bc_loss}
\end{equation}
For the trajectory-index data loader used here, each trajectory contributes 390 indexed windows per epoch; indices wrap within short trajectories, and boundary histories repeat the first frame. The loaded trajectories have no held-out split, and training uses no validation-based checkpoint selection. Table~\ref{tab:supp_base_bc_training} summarizes the training configuration.

\begin{table}[htbp]
\centering\small
\caption{\footnotesize \textbf{Architecture and training settings for the behavior-cloned base policy.} 
}
\label{tab:supp_base_bc_training}
\resizebox{0.9\textwidth}{!}{%
\begin{tabularx}{\linewidth}{@{}Xl@{}}
\toprule
Setting & Value\\
\midrule
Observation history / default prediction horizon & 10 / 1 control steps\\
Hidden width / residual blocks & 512 / 5\\
Residual main-path layers / shortcut & 2 linear / learned linear\\
Activation / hidden normalization & ReLU / none\\
Output & Linear absolute joint targets (radians)\\
Input/output normalization & None\\
Indexed training windows per trajectory per epoch & 390\\
Optimizer / learning rate & Adam / $2\times10^{-5}$\\
Adam moments / $\epsilon$ & $(0.9,0.999)$ / $10^{-8}$\\
Weight decay / learning-rate schedule & 0 / none\\
Parallel training processes & 8\\
Batch size per process / nominal global batch & 128 / 1,024\\
Training epochs (default configuration) & 10,000\\
EMA decay & 0.999\\
Checkpoint saving & Every 5,000 updates and each epoch\\
Held-out validation / validation-based selection & None / none\\
Random seed & 1234\\
\bottomrule
\end{tabularx}
}
\end{table}

\paragraph{Policy-input variants and motion-capture latency.}
\label{para:supp_policy_input_transfer}
In direct-transfer tests using the calibrated simulator described above, we compared proprioceptive histories alone with proprioceptive histories augmented by object-state histories. The latter contain the object's position relative to the hand and the six-dimensional sign-invariant longitudinal-axis encoding in Eq.~\ref{eq:delta_object_encoding}, giving three position and six axis features per object-history frame. We measured the motion-capture object-state latency and, during policy distillation, sampled object-observation delays as the measured latency scaled by a factor drawn from $\mathcal{U}(0.8,1.5)$.
This randomization concerns the object observations and is separate from the hand-encoder delay model. Even after task-agnostic actuator/delay identification, learned free-hand compensation, and task-specific contact identification, neither input formulation achieved successful task completion on any of the ten tasks in these direct-transfer tests (0/50 evaluation trials per task), including the object-augmented policies distilled with motion-capture delay randomization. The subsequent pipeline retains proprioception-only policy inputs.

\subsubsection*{\pass family}
\paragraph{Initialization.}
\passvtwo uses a rigid capsule with radius $7$~mm, a $140$~mm cylindrical section, and hemispherical caps; \passvzerons, \passvonens, and \passvthree use a rigid cylinder with radius $7$~mm and length $200$~mm. These simulated objects have the same geometry as the physical ones. Table~\ref{tab:pass-initial-poses} uses metres and XYZW quaternions, with independent $\varepsilon_y,\varepsilon_z\sim\mathcal U(-0.01,0.01)$.

\begin{table}[htbp]
    \centering
    \small
    \caption{
    \footnotesize \textbf{Initial object root poses for the four pass styles.}
    Quaternion values are rounded.}
    \label{tab:pass-initial-poses}
    \begin{tabular}{@{}lll@{}}
    \toprule
    Style & Initial position $p_0$ & Initial quaternion $q_0$ \\
    \midrule
    \textit{Pass}-V-R
    & $(-0.02,-0.16+\varepsilon_y,0.52+\varepsilon_z)$
    & $(0,0,0,1)$ \\
    \textit{Pass}-V-NR
    & $(-0.02,-0.16+\varepsilon_y,0.52+\varepsilon_z)$
    & $(0,0,0,1)$ \\
    \textit{Pass}-H-NR
    & $(-0.02,-0.09+\varepsilon_y,0.57+\varepsilon_z)$
    & $(-0.7071,0,0,0.7071)$ \\
    \textit{Pass}-H-R
    & $(-0.02,-0.09+\varepsilon_y,0.57+\varepsilon_z)$
    & $(-0.7071,0,0,0.7071)$ \\
    \bottomrule
    \end{tabular}
\end{table}

\paragraph{Targets and reward.}
The first target is $+30^\circ$ about the object's local $y$ axis from reset orientation. Each reached target increments $s$ and advances at the next control step. Reversal variants use
\begin{equation}
 q_g^{\mathrm{next}}=q_g\otimes Q_y\!\left((-1)^{\lfloor s/N_{\mathrm{rev}}\rfloor}\pi/6\right),
\end{equation}
where $Q_y$ is a local-$y$ rotation, $N_{\mathrm{rev}}=11$ for \passvzero and $12$ for \passvthree (reversals at $11,22,33,\ldots$ and $12,24,36,\ldots$). For \passvzerons, although the target schedule reverses after 11 targets (330$^\circ$), the pen overshoots at each reversal, so the measured rotation per episode is close to $4\pi$.
\passvone and \passvtwo always advance $+30^\circ$. The counter counts targets, not revolutions.

For unit object/target quaternions, define 
\begin{align}
    \Delta q_t &=q_t\otimes q_{g,t}^{-1}, \\
    e_t&=\|(\Delta q_t)_{xyz}\|_2^2, \\
    \theta_t&=2\arcsin(\min\{1,\sqrt{e_t}\}),
\end{align}
so $e_t=\sin^2(\theta_t/2)$. The reward is
\begin{equation}
 r_t^{\mathrm{env}}=6e^{-e_t/0.01}+1.5\mathbf1[e_t\leq0.1]+250\mathbf1[\theta_t\leq0.2].
 \label{eq:passv0-reward}
\end{equation}
PPO uses $r_t^{\mathrm{scaled}}=0.01r_t^{\mathrm{env}}$, with separate timeout bootstrapping. Near-target tolerance $e_t\leq0.1$ corresponds to $0.6435$ rad ($36.87^\circ$); success, advancement and the bonus use $0.2$ rad ($11.46^\circ$), twice the configured pen tolerance $0.1$. Termination occurs at $\|p_t-p_{g,t}\|_2\geq0.24$ m or timeout.

Table~\ref{tab:passv0-reward} summarizes the reward components.

\begin{table}[htbp]
\centering
\small
\caption{\footnotesize
\textbf{Reward components for \pass style rotation.}}
\label{tab:passv0-reward}
\begin{tabular}{@{}lll@{}}
\toprule
Component & Contribution to $r_t^{\mathrm{env}}$ & Purpose \\
    \midrule
    Orientation proximity
    & $6\exp(-e_t/0.01)$
    & Dense alignment shaping \\
    Near-target reward
    & $1.5\,\mathbf{1}[e_t\leq0.1]$
    & Broad proximity reward \\
    Target-success bonus
    & $250\,\mathbf{1}[\theta_t\leq0.2]$
    & Rewards target completion \\
\bottomrule
\end{tabular}
\end{table}

\subsubsection*{\trianglepass family}
Full-sequence training stalled near $90^\circ$--$180^\circ$, motivating sequential stage policies for reorientation and grasp reconfiguration. The designated task-frame rotation axis is $z$.

\paragraph{Stage sequences.}
\trianglepassup uses four stages: (1) rotate $+90^\circ$ while moving the index finger to the opposite side; (2) rotate another $+90^\circ$ (cumulative $180^\circ$) while moving the thumb across; (3) complete thumb gaiting; (4) move the middle finger across to restore the initial grasp. Stages 3--4 use reference-joint and fingertip--object relation rewards. \trianglepassdown uses the same decomposition with the palm downward.

\trianglepassdownreversed uses three stages: (1) rotate $-180^\circ$ with thumb--object relation guidance; (2) complete thumb gaiting using the same reference/relation guidance; (3) restore the grasp by moving the middle, ring and little fingers across under reference/relation rewards. During the final coordinated-gait stage, corresponding MCP FE, MCP AA, PIP and DIP joints share commands, $a_{\mathrm{middle},j}=a_{\mathrm{ring},j}=a_{\mathrm{little},j}$, while contacts remain separate. MCP FE/AA denote metacarpophalangeal flexion--extension/abduction--adduction.

\paragraph{Stage rewards.}
For stage $s$, use the quaternion orientation error defined above and
\begin{align}
 r_{t,\mathrm{task}}^{(s)}&=\lambda_g^{(s)}r_{t,\mathrm{goal}}^{(s)}+\lambda_f^{(s)}r_{t,\mathrm{rel}}^{(s)}+\lambda_h^{(s)}r_{t,\mathrm{ref}}^{(s)},\\
 r_{t,\mathrm{goal}}^{(s)}&=\exp[-\theta_t^{(s)}/\tau_{\mathrm{rot}}-\|\mathbf p_t-\mathbf p_g^{(s)}\|_2/\tau_{\mathrm{pos}}],\\
 r_{t,\mathrm{rel}}^{(s)}&=\chi_t^{(s)}\sum_{\ell\in\mathcal C_s}\exp[-E_{\ell,t}^{(s)}/\tau_f],\\
 r_{t,\mathrm{ref}}^{(s)}&=\exp\left[-\frac{\sum_{j\in\mathcal J_s}(h_{t,j}-h_{\mathrm{ref},j}^{(s)})^2}{2\sigma_h^2|\mathcal J_s|}\right].
\end{align}
Weights are $\lambda_g=0.5$ and $\lambda_f=10^{-4}$; $\lambda_h=0.1$ with reference guidance and zero otherwise. Temperatures are $\tau_{\mathrm{rot}}=0.5\,\mathrm{rad}$, $\tau_{\mathrm{pos}}=0.05\,\mathrm m$ and $\tau_f=0.01\,\mathrm m$. Current/target positions are $\mathbf p_t,\mathbf p_g^{(s)}$; $\mathcal C_s$ contains the stage relations above, summed without averaging or per-relation weights. Geometric errors are
\begin{equation}
 E_{\ell,t}^{(s)}=\begin{cases}
 [m_\ell^{(s)}-c_{\ell,t}^{(s)}]_+ & \text{target-side placement},\\
 |c_{\ell,t}^{(s)}-m_\ell^{(s)}| & \text{target surface offset},\\
 |\Delta y_{i,t}-\Delta y_{i,0}| & \text{relation preservation}.
 \end{cases}
\end{equation}
Here $[x]_+=\max(x,0)$, $c$ is signed clearance along the relevant direction or object-attached plane (positive on the desired side), and $m=0.005\,\mathrm m$. Target-side error vanishes at $c\geq m$; offset error additionally penalizes overshoot. For preservation, $\Delta y_{i,t}=f^y_{i,t}-p^y_t$ is the world-$y$ fingertip--object offset and $\Delta y_{i,0}$ its stage-initial value. Orientation-gated surface offsets use $\chi_t^{(s)}=\mathbb I[\theta_t^{(s)}\leq5^\circ=\pi/36\,\mathrm{rad}]$; otherwise $\chi=1$. These terms use fingertip geometry, not measured forces.

$h_{t,j}$ is the current joint angle; references $h_{\mathrm{ref},j}^{(s)}$ are task/stage-specific human heuristics; angles are in radians and $\sigma_h=10^\circ=\pi/18\,\mathrm{rad}$. $\mathcal J_s$ selects thumb joints for thumb-gait completion, middle-finger joints for the final palm-up/down stage, and middle/ring/little joints for the reversed final stage. Division by $|\mathcal J_s|$ normalizes across guided-joint counts.

\paragraph{Composition and distillation.}
We execute stage policies sequentially and behavior-clone successful simulation trajectories into one policy. The $180^\circ$ decomposition imposes no rollout termination; the distilled policy can continue, with $2\pi$ counted per evaluation episode, although distillation alone does not guarantee reliable continuous execution.

\subsubsection*{\thumbaroundns}
We initialize from a grasp calibrated on the physical hand. With object angular velocity $\boldsymbol\omega_t$, $\mathbf e_z=(0,0,1)^\top$ and signed accumulated rotation $\phi_t=\sum_{k=1}^t\boldsymbol\omega_k^\top\mathbf e_z\Delta t$, the reward switches permanently to recovery after $270^\circ$, using $G_t=\mathbb I[\max_{1\leq k\leq t}\phi_k>3\pi/2]$:
\begin{align}
 r_t^{\mathrm{rot}}&=1.25\operatorname{clip}(\boldsymbol\omega_t^\top\mathbf e_z,-0.5,0.5),\\
 r_t^{\mathrm{rec}}&=\lambda_d e^{-d_t^2/(2\sigma_d^2)}
 -\lambda_v[1-e^{-\|\mathbf v_t\|^2/(2\sigma_v^2)}]\nonumber\\
 &\quad-\lambda_\omega[1-e^{-\|\boldsymbol\omega_t\|^2/(2\sigma_\omega^2)}],\\
 r_t&=(1-G_t)r_t^{\mathrm{rot}}+G_t(1-D_t)r_t^{\mathrm{rec}}-50D_t.
\end{align}
Here $d_t=\|\mathbf f_{\mathrm{index},t}-\mathbf p_t\|_2$ is index-tip/object-center distance, $\mathbf v_t$ is linear velocity, and $D_t=\mathbb I[p_t^z<0.4\,\mathrm m]$ detects drops. We use $\lambda_d=0.5$, $\lambda_v=\lambda_\omega=0.0001$ and $\sigma_d=\sigma_v=\sigma_\omega=1.0$ in their respective units. Recovery rewards returning toward the hand and penalizes residual motion; drops disable recovery and incur $50$.

\subsubsection*{\sonicdownns}
We reward directed rotation of virtual endpoints $\mathbf k_{1,2}=(0,0,\mp0.075)^\top\,\mathrm m$ projected onto world $yz$, preservation of initial projected radii, downward motion and action regularization. Define $\mathbf u_{i,t}=P_{yz}R(\mathbf q_t)\mathbf k_i$, $\rho_{i,t}=\|\mathbf u_{i,t}\|_2$, where $R$ maps object to world coordinates and $P_{yz}$ selects $y,z$; record $\rho_{i,0}$ at reset. Then
\begin{align}
 Q_t&=\tfrac12\sum_{i=1}^2\exp[-((\rho_{i,t}-\rho_{i,0})/0.01)^2],\\
 \delta\phi_{i,t}&=\operatorname{atan2}(u^y_{i,t-1}u^z_{i,t}-u^z_{i,t-1}u^y_{i,t},\mathbf u_{i,t-1}^\top\mathbf u_{i,t}),\\
 \Omega_t&=-\frac{\sum_i m_{i,t}\delta\phi_{i,t}}{\Delta t_c\max(1,\sum_i m_{i,t})},\qquad\Delta t_c=0.01\,\mathrm s,\\
 r_t&=Q_t\operatorname{clip}(\Omega_t,-\pi,\pi)+0.25Q_t
       +0.03r_t^{\mathrm{down}}-2\times10^{-4}\|\mathbf a_t\|_2^2.
\end{align}
The validity mask $m_{i,t}$ excludes the first control step and endpoints whose current or previous radius is at most $10^{-4}\,\mathrm m$. The minus sign selects rotation about negative world $x$. For normalized actions $\mathbf a_t$, $r_t^{\mathrm{down}}=\operatorname{clip}(-v_t^z,0,0.5\,\mathrm{m\,s^{-1}})$ rewards bounded downward speed without penalizing translational-speed magnitude.

\subsubsection*{\fourfingerpassleapns}
Following HORA~\cite{Qi2022InHandOR}, we initialize from $5\times10^5$ stable palm-down grasps and reward positive-$z$ rotation:
\begin{align}
 r_t={}&1.25\operatorname{clip}(\boldsymbol\omega_t^\top\mathbf e_z,-0.5,0.5)+r_t^{\mathrm{guide}}
 -0.3\|\mathbf v_t\|_1-0.3\|\mathbf h_t-\mathbf h_0\|_2^2\nonumber\\
 &-0.1\|\boldsymbol\tau_t\|_2^2-2.0(\boldsymbol\tau_t^\top\dot{\mathbf h}_t^{\mathrm{FD}})^2,\\
 r_t^{\mathrm{guide}}={}&\operatorname{clip}(0.03/(\theta_t+10^{-6}),0,0.3)+0.3\mathbb I[\theta_t<\pi/12].
\end{align}
$\mathbf h_0$ is the reset grasp configuration; $\mathbf h_t$, $\boldsymbol\tau_t$ and $\dot{\mathbf h}_t^{\mathrm{FD}}$ are joint angles, applied torques and low-level-controller finite-difference velocities. The final term penalizes squared net mechanical power. $\theta_t$ uses the quaternion error defined above in radians; below $15^\circ$, the guiding target advances by $90^\circ$ about the desired axis. Object velocities use consecutive control-step poses with an approximately $0.05\,\mathrm s$ interval.

\subsection*{Isaac Lab Training Settings}
For the direct sim-to-real transfer experiments in Supplementary Section~S3, we trained PPO policies on eight GPUs with 32,768 environments per GPU, using approximately 12.44 billion environment steps for \passvzero and 15 billion for \fourfingerpassleapns.

\paragraph{\passvzerons.}
The entropy coefficient was $5\times10^{-4}$. The pinky finger remained locked until the first $180^\circ$ of net forward rotation to encourage ring-finger use. With orientation error $\theta_t$ and $e_t=\sin^2(\theta_t/2)$, the reward included $6\exp(-e_t/0.01)$, a $1.5$ bonus for $e_t\leq0.1$, and a $250$ target-success bonus for $\theta_t\leq0.2$, together with a small action penalty. Object displacement of at least $0.24\,\mathrm{m}$ triggered termination and a $250$ penalty. One-shot bonuses rewarded qualifying little-finger dorsal contact within approximately $0.2\,\mathrm{s}$ of odd-numbered reversals ($50$) and subsequent completion of the first reverse-direction target ($100$), without an intervening drop. 

\paragraph{\fourfingerpassleapns.}
Gravity was fixed at $9.81\,\mathrm{m/s^2}$. Targets advanced in $90^\circ$ increments when the cuboid's projected long-axis heading error in the world $xy$ plane fell below $15^\circ$, while reward shaping and its success bonus used full three-dimensional orientation error. The reward combined clipped $z$-axis angular velocity, orientation shaping, and penalties on object translation, joint displacement from the initial pose, torque, and squared total joint power. The episode limit was 400 control steps.

\subsection*{Stage 1: Human-Guided Corrective Optimization}
\namedlabel{sec:corrective_policy_details_in_epxsettings}{Correction Policy}
We behavior-clone a full-hand correction policy that adds joint-target offsets to the fixed base policy. 

\paragraph{Input and architecture.}
For $n_q$ joints, measured positions $\mathbf q_t$, issued commands $\mathbf a_t^{\mathrm{cmd}}$ and base targets $\mathbf a_t^{\mathrm{base}}$, the input is

\begin{equation}
    \mathbf{o}_t^{\mathrm{rec}}
    =\operatorname{concat}\!\left(
    \mathbf{q}_{t-H_{\mathrm{rec}}+1:t},
    \mathbf{a}_{t-H_{\mathrm{rec}}:t-1}^{\mathrm{cmd}},
    \mathbf{a}_t^{\mathrm{base}}
    \right)
    \in\mathbb{R}^{(2H_{\mathrm{rec}}+1)n_q}.
    \label{eq:recovery_bc_observation}
\end{equation}

Histories are flattened chronologically and initialized by repeating the initial joint configuration; no object observations are included. Fixed training-data population statistics standardize features as $\bar o_k=\operatorname{clip}((o_k-\mu_k)/(\max(\sigma_k,\epsilon)+\epsilon),-10,10)$, $\epsilon=10^{-6}$. Table~\ref{tab:recovery_bc_hyperparameters} lists dimensions and settings. The deterministic, single-step residual MLP is

\begin{equation}
    \begin{aligned}
        \mathbf{h}_0
        &=\operatorname{ELU}\!\left(W_{\mathrm{in}}\bar{\mathbf{o}}_t^{\mathrm{rec}}+\mathbf{b}_{\mathrm{in}}\right),\\
        \mathbf{h}_{\ell}
        &=\operatorname{ELU}\!\left(
        \operatorname{LN}\!\left[
        \mathbf{h}_{\ell-1}
        +W_{\ell,2}\operatorname{ELU}\!\left(W_{\ell,1}\mathbf{h}_{\ell-1}+\mathbf{b}_{\ell,1}\right)
        +\mathbf{b}_{\ell,2}
        \right]\right),\quad \ell=1,2,3,\\
        \delta\mathbf{a}_t^{\mathrm{rec}}
        &=\pi_{\mathrm{rec},\xi}\!\left(\bar{\mathbf{o}}_t^{\mathrm{rec}}\right)
        =r_{\max}\tanh\!\left(W_{\mathrm{out}}\mathbf{h}_3+\mathbf{b}_{\mathrm{out}}\right),
        \qquad r_{\max}=0.5\,\mathrm{rad}.
    \end{aligned}
    \label{eq:recovery_bc_architecture}
\end{equation}

It has no recurrence/chunking. Deployment commands are $\mathbf a_t^{\mathrm{cmd}}=\operatorname{clip}(\mathbf a_t^{\mathrm{base}}+\delta\mathbf a_t^{\mathrm{rec}},\mathbf q^{\min},\mathbf q^{\max})$.

\paragraph{Supervision and optimization.}
After corrective optimization enables full-episode success, we perform a separate real-world data-collection phase using the optimized corrective primitives and retain successful trajectories to train the correction policy. Hardware operation during this additional collection is capped at 3 hours per task. We do not impose a common trajectory count across tasks: the number of retained successful trajectories depends on the task-specific success rate and rollout duration. This collection is separate from corrective-optimization trials and excluded from the rollout counts in Fig.~\ref{fig:framework_stages}, but its full budget is included in the human-effort and hardware-time estimates for settings with S1. 
Targets $\mathbf y_t$ are recorded residuals.
The loss is

\begin{equation}
    \begin{aligned}
        \mathcal{L}_{\mathrm{rec}}(\xi)
        &=\frac{1}{Bn_q}\sum_{b=1}^{B}
        \left[
        w_b\left\|\pi_{\mathrm{rec},\xi}(\bar{\mathbf{o}}_b^{\mathrm{rec}})-\mathbf{y}_b\right\|_2^2
        +\lambda_{\mathrm{out}}\left\|\pi_{\mathrm{rec},\xi}(\bar{\mathbf{o}}_b^{\mathrm{rec}})\right\|_2^2
        \right],\\
        w_b&=10I_b+0.05(1-I_b),
    \end{aligned}
    \label{eq:recovery_bc_loss}
\end{equation}

$\lambda_{\mathrm{out}}=10^{-4}$ penalizes correction magnitude. $I_b=1$ when the recorded gate exceeds $0.5$, or is positive with residual norm above $10^{-5}$. Non-intervention targets are not replaced by zero.
Table~\ref{tab:recovery_bc_hyperparameters} specifies optimization. AdamW decay is distinct from output regularization.

\begin{table}[htbp]
\centering\small
\caption{\footnotesize
\textbf{Architecture and training hyperparameters for the correction policy.} The input and output dimensions depend on the number of controlled joints, $n_q$.}
\label{tab:recovery_bc_hyperparameters}
\begin{tabularx}{0.9\linewidth}{@{}lX@{}}
\toprule
Parameter & Value\\
\midrule
History length $H_{\mathrm{rec}}$ & 10 control steps\\
Input dimension & $21n_q$ (462 for Sharpa; 336 for LEAP)\\
Output dimension & $n_q$ (22 for Sharpa; 16 for LEAP)\\
Hidden width / residual blocks & 512 / 3\\
Linear layers per residual block & 2\\
Activation / block normalization & ELU / LayerNorm\\
Output parameterization & $0.5\tanh(\cdot)$, in radians\\
Action prediction horizon & 1 control step\\
Observation normalization & Fixed dataset mean and standard deviation\\
Normalization $\epsilon$ / clipping range & $10^{-6}$ / $[-10,10]$\\
Optimizer & AdamW\\
Learning rate & $3\times10^{-4}$, constant\\
AdamW moments / $\epsilon$ & $(0.9,0.999)$ / $10^{-8}$\\
AdamW weight decay & $10^{-2}$ (library default)\\
Gradient-norm limit & 10\\
Training updates / batch size & 20,000 / 256\\
Sampling allocation & 128 intervention frames + 128 from all frames\\
Intervention / non-intervention loss weight & 10 / 0.05\\
Output-magnitude penalty $\lambda_{\mathrm{out}}$ & $10^{-4}$\\
Reward-based loss weighting & Disabled\\
Checkpoint interval & 2,000 updates\\
\bottomrule
\end{tabularx}
\end{table}

\subsection*{Stage 2: Model-Based Policy Refinement}

\subsubsection*{Per-Contact Neural Compensation} \label{sec:workflow_per_nerual_contact}
The Sharpa and LEAP hands share this workflow and its training settings, with $n_{\mathrm{dof}}=22$ and $16$ controlled joints, respectively. Joint-dependent input and output dimensions are adjusted accordingly.

\paragraph{Data and replay.}
The reference dataset is $\mathcal D^{\mathrm{real}}=\{\mathbf T_i^{\mathrm{real}}\}_{i=1}^N$, with $\mathbf T_i^{\mathrm{real}}=\{(\mathbf s_t^h,\mathbf s_t^o,\mathbf a_t)\}_{t=0}^{L_i}$: hand joint positions, object position and WXYZ quaternion, and applied joint-position targets. $L_i$ is the final frame index; hats denote simulation states. 
Resets sample trajectories uniformly and initialize their first-frame configurations with zero velocities. The simulated object has the same geometry as the physical object that produced the recorded trajectories.

At each physics step, analytical kinematics and collision detection construct constraints; the shared network modifies valid contact rows before actuation, the coupled solve, and integration under the replayed targets. The counter then advances before reward/termination evaluation, followed by resets, target updates, and reconstruction of next-step observations.

With physics frequency $f_s$ and recorded control frequency $f_c$, reward reference index $t_r(t_s)=\min(\lfloor t_sf_c/f_s\rfloor+t_{r,0},L_i)$ uses $t_{r,0}=0$; the next target uses $\lfloor(t_s+1)f_c/f_s\rfloor+t_{r,0}$. Reference progression follows time.

\paragraph{Contact selection and observations.}

Contacts must belong to a valid environment, generate constraint rows, and satisfy $m_c-d_c>0$ (inclusion margin $m_c$, signed distance $d_c$). We retain up to 32 contacts in simulator order, without depth sorting or hand--object-only filtering. Each stores up to six row addresses; excess rows raise an error. Missing rows and slots are invalid and zero-padded before observation noise.

For simulated object pose $(\hat{\mathbf p},\hat R)$, positions and contact-frame axes use \emph{object-body} coordinates: $\mathbf x_c^o=\hat R^\top(\mathbf x_c^w-\hat{\mathbf p})$ and $[\mathbf n_c^o\;\mathbf t_{c,1}^o\;\mathbf t_{c,2}^o]=\hat R^\top[\mathbf n_c^w\;\mathbf t_{c,1}^w\;\mathbf t_{c,2}^w]$. With $m_c^+=\max(m_c,10^{-6})$, normalization is
\[
\delta_c=\clip\!\left(\frac{\max(m_c-d_c,0)}{m_c^+},0,5\right),\quad
\bar d_c=\clip(d_c/m_c^+,-5,5),\quad
\bar{\mathbf x}_c=\clip(\mathbf x_c^o/(0.10\,\mathrm m),-5,5).
\]
Position clipping is componentwise; zero margins use the denominator floor and can saturate distance features. Table~\ref{tab:hybrid_actor_obs} lists 40 contact descriptors plus six row velocities and six type codes ($\mathcal C_c$: 52 features), completed by six accelerations and six inverse weights ($\mathbf o_c$: 64). Geometry identifiers and hand-body contact positions support compensation for body-/region-specific convex-decomposition errors; this grouping does not alter storage order.

\begin{table}[htbp]
\centering\small
\caption{\footnotesize
\textbf{Per-contact actor observation $\mathbf{o}_c$.} The 40 contact descriptors and 24 constraint-row descriptors form a 64-dimensional input, partitioned into $\mathcal{C}_c$, $\mathbf{a}_{\mathrm{ref},c}$, and $\mathbf{D}_c$ as described in the text.}
\label{tab:hybrid_actor_obs}
\begin{tabularx}{0.9\linewidth}{@{}Xr@{}}
    \toprule
    Component & Dimension\\
    \midrule
    Active-contact indicator & 1\\
    Normalized penetration and signed distance $(\delta_c,\bar d_c)$ & 2\\
    Normalized contact position in the object body frame & 3\\
    Normalized contact position in the colliding hand body frame & 3\\
    Contact normal and two tangent axes in the object body frame & 9\\
    Contact friction coefficients & 5\\
    Normal and friction reference parameters (\texttt{solref}, \texttt{solreffriction}) & 4\\
    Impedance parameters (\texttt{solimp}) & 5\\
    One-hot contact dimension over $\{1,3,4,6\}$ & 4\\
    Two geometry identifiers, normalized by the model's geometry count minus one & 2\\
    Contact type and collision identifier & 2\\
    Constraint-row velocities & 6\\
    Analytical constraint-row reference accelerations $\mathbf{a}_{\mathrm{ref},c}$ & 6\\
    Analytical constraint-row inverse regularization weights $\mathbf{D}_c$ & 6\\
    Constraint-row type codes & 6\\
    \midrule
    Total per contact & 64\\
    Total across 32 slots & 2048\\
    \bottomrule
\end{tabularx}
\end{table}

Running mean/std uses valid contacts only, with denominator $\mathrm{std}+10^{-2}$. Explicit geometry/collision identifiers mean that shared weights and relative features do not imply geometry-identity independence.

\begin{table}[htbp]
\centering\small
    \caption{\footnotesize
    \textbf{Critic observation for the default rigid-contact Sharpa model.}}
\label{tab:hybrid_critic_obs}
    \begin{tabularx}{0.9\linewidth}{@{}Xr@{}}
        \toprule
        Component & Dimension\\
        \midrule
        Hand joint positions, normalized using joint limits & 22\\
        Previous raw contact-compensation action & 384\\
        Object position & 3\\
        Sign-invariant representation of the object's longitudinal axis & 6\\
        Reference object position & 3 \\
        Sign-invariant representation of the reference object's longitudinal axis & 6\\
        Object world-frame linear and angular velocities & 6\\
        World-frame pose and linear/angular velocity of five fingertip geometries & 65\\
        Contact-found indicator and contact-frame force for 14 hand-subtree sensor slots & 56\\
        \midrule
        Total & 551\\
        \bottomrule
    \end{tabularx}
\end{table}

For Table~\ref{tab:hybrid_critic_obs}, object-frame keypoints $(0,0,\pm0.075)\,\mathrm m$ define unit axis $\boldsymbol\ell$. Its sign-invariant features are $[\ell_x^2,\ell_y^2,\ell_z^2,\ell_x\ell_y,\ell_x\ell_z,\ell_y\ell_z]$. Each fingertip contributes 7 pose and 6 velocity features; each of 14 phalangeal sensor slots contributes a found flag and three force components. Only current sensor values are read despite a four-step allocation. The uncorrupted critic observes reference position/axis; the actor observes contacts only. Neither receives a full reference quaternion or phase variable; reference information also enters replay, reset, and reward.

\paragraph{Networks.}
The shared actor processes each contact independently through a normalized-input $64\rightarrow64\rightarrow32\rightarrow16$ trunk (ELU after the first two layers), then two linear six-output mean heads for the weight and acceleration branches. 
Zero head weights/biases yield identity correction under deterministic mean evaluation. 
Concatenating 32 contacts produces 384 correction coefficients, distinct from hand commands.
The separate critic is $d_{\mathrm{critic}}\rightarrow512\rightarrow256\rightarrow128\rightarrow1$ ($d_{\mathrm{critic}}=551$ for Sharpa), with hidden ELUs and linear output.

\paragraph{Tracking reward.}

We use both object and hand tracking rewards in Eq.~\eqref{eq:supp_model_tracking_reward}, with $w_o=w_h=1$. 
For simulated/reference quaternions, let $\mathbf e=\hat{\mathbf Q}_{t_s}\otimes\mathbf Q_{t_r}^{*}=(e_w,e_x,e_y,e_z)$, where $*$ is conjugation and $\otimes$ quaternion multiplication. For unit quaternions, $E_R=e_x^2+e_y^2+e_z^2=\sin^2(\varphi/2)$ for relative angle $\varphi$. Let $d_p=\|\hat{\mathbf p}_{t_s}-\mathbf p_{t_r}\|_2$ denote position error in meters. The object-tracking term is

\begin{align}
    r_o\!\left(\hat{\mathbf{s}}_{t_s}^o,\mathbf{s}_{t_r}^o\right)
    ={}&6\exp\!\left(-\frac{E_R}{0.01}\right)
     +6\exp\!\left(-\frac{d_p}{0.01}\right)
    \notag\\
    &+0.8\|\hat{\boldsymbol\omega}_{t_s}\|_2\exp\!\left(-\frac{E_R}{0.05}\right)
     +1.5\ind\{E_R\leq0.1\}
    \notag\\
    &+250\ind\{d_R\leq0.1,\ d_p\leq0.005\},
    \label{eq:hybrid_tracking_reward}\\
    d_R={}&2\arcsin\!\left(\min(\sqrt{E_R},1)\right),
\end{align}
where $\hat{\boldsymbol\omega}_{t_s}$ is world-frame object angular velocity and $d_R$ is the geodesic angle between the simulated and reference orientations, computed from all three vector components of $\mathbf e$. The speed term is gated by orientation error, not reference-speed matching. The final bonus applies at every step satisfying both $d_R\leq0.1$ and $d_p\leq0.005\,\mathrm m$ (5 mm). 

The hand-tracking term is
\begin{equation}
    r_h\!\left(\hat{\mathbf s}_{t_s}^h,\mathbf s_{t_r}^h\right)
    =\exp\!\left(-\frac{\|\hat{\mathbf q}_{t_s}-\mathbf q_{t_r}\|_2}{0.01}\right),
    \label{eq:hybrid_hand_tracking_reward}
\end{equation}
where $\hat{\mathbf q}_{t_s}$ and $\mathbf q_{t_r}$ contain the simulated and recorded hand joint angles in radians; the norm spans all 22 Sharpa joints without joint-limit normalization.

\paragraph{Optimization and simulation settings.}

Table~\ref{tab:hybrid_ppo} applies \emph{only} to contact-model PPO training. PPO uses generalized advantage estimation and sampled-coefficient likelihoods, without differentiating through physics. Contact training and both policy-refinement phases use elliptic friction cones. Refinement instead shares the base-policy PPO configuration.

\begin{table}[htbp]
    \centering\small
    \caption{\footnotesize
    \textbf{PPO and simulation settings for training the contact compensator.}}
    \label{tab:hybrid_ppo}
    \begin{tabularx}{0.8\linewidth}{@{}Xl@{}}
    \toprule
    Setting & Value\\
    \midrule
    Optimizer & Adam\\
    Initial learning rate & $10^{-3}$\\
    Learning-rate schedule / desired KL & Adaptive / $0.01$\\
    Discount factor / GAE parameter & $0.99$ / $0.95$\\
    PPO clipping parameter & $0.2$\\
    Value-loss coefficient / entropy coefficient & $1.0$ / $0.005$\\
    Clipped value loss & Enabled\\
    Optimization epochs / minibatches per update & $5$ / $4$\\
    Collected steps per environment per update & $24$\\
    Parallel environments & $8\times16{,}384$\\
    Maximum training iterations / checkpoint interval & $5000$ / $100$\\
    Maximum gradient norm & $1.0$\\
    Initial output standard deviation & $1.0$ per channel\\
    Simulation timestep / decimation & Per task (Table~\ref{tb:per_task_simulation_settings}) / $1$\\
    Maximum episode duration & Duration of the sampled reference trajectory \\ 
    Integrator / friction-cone representation & Implicit-fast / elliptic\\
    Solver iterations / line-search iterations & $100$ / $50$\\
    Impedance ratio & $1$\\
    Simulator contact capacity / constraint capacity & $128$ / $2000$\\
    Actor contact capacity / rows per contact & $32$ / $6$\\
    \bottomrule
    \end{tabularx}
\end{table}

Episodes terminate on timeout, object height $z\leq0.45\,\mathrm m$, or nonfinite physics/states. 


\paragraph{Policy refinement with contact compensation.}
We train contact compensation on 200 real-world rollouts, including successful and failed trials, from the base plus stage-one correction policy, then refine the distilled base alone with PPO. 
The reference trajectories are collected with the correction policy, whereas simulated refinement uses the base alone; the rationale is given under Sec.~\nameref{sec:supp_methdo_refinement_scehdule}.

The critic is warm-started from base training and uses privileged observations; the policy receives $20n_{\mathrm{dof}}$ state--action-history features. We use 1,000 PPO epochs and $8\times4{,}096$ environments, select the highest-training-reward checkpoint, and otherwise retain base-training rewards, PPO, and simulator settings.

\subsubsection*{Joint-Wise Delta-Action Compensation} \label{sec:workflow_delta_action_model}

\paragraph{Workflow.}
We fit differentiable surrogate $F_\psi$ to the contact-compensated simulator dataset $\mathcal D^{\mathrm{mj}+g_\theta}$, then fix it and train joint-wise delta actions $\phi=\{\phi_i\}_{i=1}^{n_{\mathrm{dof}}}$ on $\mathcal D^{\mathrm{real}}$. Below, delta/residual actions are synonymous; the recurrent implementation extends the main-text transition model to 16 recursive predictions.

\paragraph{State encoding and temporal alignment.}

Joint positions and joint-position targets are normalized using the same joint limits:
\begin{equation}
        \mathcal U_i(v)=\frac{2v-q_i^{\min}-q_i^{\max}}{q_i^{\max}-q_i^{\min}},
        \qquad
        \bar q_{t,i}=\mathcal U_i(q_{t,i}),
        \qquad
        \bar a_{t,i}=\mathcal U_i(a_{t,i}).
    \label{eq:delta_joint_normalization}
\end{equation}
Here $\mathbf a_t$ drives transition $t\rightarrow t+1$; this fixed dataset normalization applies even with learned normalization disabled. Only normalized joint positions and object-axis features are predicted, excluding velocities, forces, and other simulator states.

For the default cylindrical object, define two local keypoints $\mathbf{k}_A=(0,0,-0.075)^\top\,\mathrm{m}$ and $\mathbf{k}_B=(0,0,0.075)^\top\,\mathrm{m}$. With object rotation matrix $R_t$, the unit longitudinal axis and the six-dimensional object encoding are
\begin{align}
    \boldsymbol\ell_t
    &=\frac{R_t(\mathbf{k}_B-\mathbf{k}_A)}{\|R_t(\mathbf{k}_B-\mathbf{k}_A)\|_2},\\
    \mathbf{x}_t^o=\rho(\mathbf{s}_t^o)
    &=\bigl(\ell_{t,x}^2,\ell_{t,y}^2,\ell_{t,z}^2,
    \ell_{t,x}\ell_{t,y},\ell_{t,x}\ell_{t,z},\ell_{t,y}\ell_{t,z}\bigr)^\top.
    \label{eq:delta_object_encoding}
\end{align}
These six unconstrained outputs encode $\boldsymbol\ell_t\boldsymbol\ell_t^\top$, invariant to axial rotation and endpoint exchange. They encode an unsigned axis, not endpoints; object translation is excluded from inputs and loss.

The state is $\mathbf x_t=(\bar{\mathbf q}_t,\mathbf x_t^o)\in\mathbb R^{n_{\mathrm{dof}}+6}$. Each history frame additionally stores the normalized target held at that state, $\bar{\mathbf v}_t=\bar{\mathbf a}_{t-1}$:

\begin{equation}
    \mathcal H_t=
    \left[\bigl(\bar{\mathbf{q}}_\tau,\bar{\mathbf{v}}_\tau,\mathbf{x}_\tau^o\bigr)\right]_{\tau=t-H_{\mathrm{ctx}}+1}^{t},
    \qquad H_{\mathrm{ctx}}=8.
    \label{eq:delta_wm_history}
\end{equation}
Each of eight frames has $2n_{\mathrm{dof}}+6$ features (Sharpa: 50; LEAP: 38).

\paragraph{Surrogate world-model training.}

A two-layer width-256 GRU encodes history; its final upper-layer output concatenated with the command has $256+n_{\mathrm{dof}}$ features (Sharpa: 278; LEAP: 272). A width-256 encoder feeds separate joint/object heads. Each encoder/head has input/output linear layers and five width-256 residual blocks with learned linear shortcuts:

\begin{equation}
    \mathcal B(\mathbf{z})=
    \operatorname{ReLU}\!\left(
    W_2\operatorname{ReLU}(W_1\mathbf{z}+\mathbf{b}_1)
    +\mathbf{b}_2+W_{\mathrm{skip}}\mathbf{z}+\mathbf{b}_{\mathrm{skip}}
    \right),
    \label{eq:delta_residual_block}
\end{equation}
Linear heads predict absolute normalized states, not increments. Optional policy-in-the-loop and contact-feature branches are disabled.

\textit{Recursive prediction and loss.}
With recurrent state $\mathbf z_k$ for both GRU layers,

\begin{equation}
    \left(\hat{\mathbf{x}}_{k+1},\mathbf{z}_{k+1}\right)
    =F_\psi\!\left(\mathcal H_k,\bar{\mathbf{a}}_k;\mathbf{z}_k\right).
    \label{eq:delta_recurrent_wm}
\end{equation}
Each updated history appends $(\hat{\bar{\mathbf q}}_{k+1},\bar{\mathbf a}_k,\hat{\mathbf x}_{k+1}^o)$. Recorded commands drive all $P=16$ steps; predictions replace future states without teacher forcing. Hidden states start at zero and persist while the \emph{entire} updated eight-frame window is processed each step. Gradients span all steps. The discrepancy and objective are

\begin{equation}
    d(\hat{\mathbf{x}},\mathbf{x})=
    \lambda_h\|\hat{\bar{\mathbf{q}}}-\bar{\mathbf{q}}\|_2^2
    +\lambda_o\|\hat{\mathbf{x}}^o-\mathbf{x}^o\|_2^2,
    \qquad \lambda_h=\lambda_o=1.
    \label{eq:delta_state_discrepancy}
\end{equation}

\begin{equation}
    \psi^\star=\arg\min_\psi\;
    \mathbb E_{(\mathcal H_t,\mathbf{a}_{t:t+P-1},\mathbf{x}_{t+1:t+P})\sim\mathcal D^{\text{mj}+g_\theta}}
    \left[\sum_{k=t}^{t+P-1}d\!\left(\hat{\mathbf{x}}_{k+1},\mathbf{x}_{k+1}\right)\right].
    \label{eq:delta_wm_objective}
\end{equation}
Errors sum over features and horizon and average over the batch.

\paragraph{Joint-wise delta-action compensator training.}

\textit{Independent joint networks.}
Each joint has a separately parameterized network with two-frame local input:

\begin{equation}
    \mathbf{h}_{t,i}=\left(
    \bar q_{t-1,i},\bar q_{t,i},
    \bar v_{t-1,i},\bar v_{t,i},\bar a_{t,i}
    \right)^\top\in\mathbb R^5.
    \label{eq:delta_joint_history}
\end{equation}
Each network consists of a $5\rightarrow128$ linear layer, two width-128 residual blocks of the form in Eq.~\eqref{eq:delta_residual_block}, and a $128\rightarrow1$ linear layer followed by a sigmoid. The normalized correction is
\begin{equation}
    \Delta\bar a_{t,i}(\phi_i)
    =\epsilon_a\left[2\sigma\!\left(f_{\phi_i}(\mathbf{h}_{t,i})\right)-1\right],
    \qquad \epsilon_a=0.1,
    \label{eq:delta_bounded_action}
\end{equation}
Here $f_{\phi_i}$ ends at the linear output. Corrections lie in $(-0.1,0.1)$; inputs exclude recurrence, object features, and other joints. Concatenation gives $\bar{\mathbf a}_t^\phi=\bar{\mathbf a}_t+\Delta\bar{\mathbf a}_t(\phi)$, equivalent in physical coordinates to

\begin{equation}
    \Delta a_{t,i}(\phi_i)=
    \frac{q_i^{\max}-q_i^{\min}}{2}\,\Delta\bar a_{t,i}(\phi_i),
    \qquad
    \mathbf{a}_t^\phi=\mathbf{a}_t+\Delta\mathbf{a}_t(\phi).
    \label{eq:delta_physical_action}
\end{equation}
Compensated commands are not clipped to $[-1,1]$. Joint-wise updates remain coupled through the full-hand surrogate and shared objective.

\textit{Compensating the initial action history.}
For an interior sample, nine overlapping eight-frame recorded windows and their next commands yield preliminary corrections. The first eight modify held targets in the next eight measured frames:

\begin{equation}
    \bar{\mathbf{v}}_\tau^\phi
    =\bar{\mathbf{v}}_\tau
    +\Delta\bar{\mathbf{a}}_{\tau-1}^{\mathrm{init}}(\phi),
    \qquad
    \mathcal H_t^\phi=
    \left[\bigl(\bar{\mathbf{q}}_\tau,\bar{\mathbf{v}}_\tau^\phi,\mathbf{x}_\tau^o\bigr)\right]_{\tau=t-7}^{t}.
    \label{eq:delta_corrected_history}
\end{equation}
The $\mathrm{init}$ corrections use uncompensated recorded windows; measured states remain unchanged. Gradients also pass through these initial-history corrections.

\textit{Transition matching through the fixed surrogate.}
For $P=16$ transitions from $\mathcal H_t^\phi$, corrections are recomputed from the evolving history and next recorded command:

\begin{equation}
    \left(\hat{\mathbf{x}}_{k+1}^\phi,\mathbf{z}_{k+1}^\phi\right)
    =F_{\psi^\star}\!\left(
    \mathcal H_k^\phi,
    \bar{\mathbf{a}}_k+\Delta\bar{\mathbf{a}}_k(\phi);
    \mathbf{z}_k^\phi
    \right).
    \label{eq:delta_compensated_transition}
\end{equation}
Histories append $(\hat{\bar{\mathbf q}}_{k+1}^\phi,\bar{\mathbf a}_k^\phi,\hat{\mathbf x}_{k+1}^{o,\phi})$; measured future states serve only as targets:
\begin{equation}
    \phi^\star=\arg\min_\phi\;
    \mathbb E_{(\mathcal H_t,\mathbf{a}_{t:t+P-1},\mathbf{x}_{t+1:t+P})\sim\mathcal D^{\text{real}}}
    \left[\sum_{k=t}^{t+P-1}
    d\!\left(\hat{\mathbf{x}}_{k+1}^\phi,\mathbf{x}_{k+1}\right)\right],
    \label{eq:delta_residual_objective}
\end{equation}
The same $d$ and weights apply, without correction labels or a residual-magnitude penalty. Only $\phi$ is optimized; gradients propagate through the fixed surrogate. Its parameter gradients are cleared each batch and recurrent state reset between batches.

\paragraph{Datasets and optimization settings.}
After contact-based refinement, we collect 200 real-world rollouts, including successful and failed trials, with refined base plus stage-one correction policy and 100,000 contact-compensated simulation rollouts with the base alone. Joint order and XYZW quaternion conversion precede caching.

Surrogate trajectories are split 99\%/1\% for training/validation in their existing dictionary order, without shuffling. Delta-action models use all 200 real-world rollouts for training, with no validation set. Each trajectory supplies 400 logical indices per epoch; deterministic window starts wrap to the usable range. Short histories repeat the first frame; the world-model loader clamps future indices to the final frame. Delta-action starts additionally respect prediction horizon and recorded progress. Padded boundary frames contribute to loss without per-step validity masks.

Both models train for 100 epochs. The surrogate selects the minimum-validation-loss checkpoint; the delta-action models retain the final training checkpoint.
Table~\ref{tab:delta_training_settings} summarizes the settings; Adam additionally uses moments $(0.9,0.999)$, $\epsilon=10^{-8}$, and zero weight decay.

\begin{table}[htbp]
\centering\small
\caption{\footnotesize
\textbf{Training settings for the surrogate world model and joint-wise delta-action compensator.}}
\label{tab:delta_training_settings}
\resizebox{0.8\linewidth}{!}{
    \begin{tabularx}{\linewidth}{@{}Xll@{}}
    \toprule
    Setting & Surrogate $F_\psi$ & Residual models $\{f_{\phi_i}\}$\\
    \midrule
        Training dataset & $\mathcal D^{\text{mj}+g_\theta}$ & $\mathcal D^{\text{real}}$\\
        Training / validation split & 99\% / 1\% & All data / no validation\\
        Training epochs & 100 & 100\\
        Checkpoint selection & Minimum validation loss & Final\\
        Measured initial context & 8 frames & 8 frames\\
        Features per world-model history frame & $2n_{\mathrm{dof}}+6$ & $2n_{\mathrm{dof}}+6$ (fixed surrogate)\\
        Local residual-network input & --- & 5 features per joint\\
        History used by each residual network & --- & 2 frames\\
        Prediction horizon $P$ & 16 & 16\\
        Recurrent encoder & 2-layer GRU, width 256 & Fixed surrogate only\\
        Residual-block width & 256 & 128\\
        Residual blocks & 5 per encoder/head & 2 per joint network\\
        Predicted state dimension & $n_{\mathrm{dof}}+6$ & $n_{\mathrm{dof}}+6$ (fixed surrogate)\\
        Action correction dimension & --- & $n_{\mathrm{dof}}$\\
        Normalized correction bound & --- & $0.1$ per joint\\
        Loss weights $(\lambda_h,\lambda_o)$ & $(1,1)$ & $(1,1)$\\
        Batch size per process & 128 & 128\\
        Parallel training processes & 8 & 8\\
        Optimizer / learning rate & Adam / $2\times10^{-5}$ & Adam / $2\times10^{-5}$\\
        Parameter EMA coefficient & 0.999 & 0.999\\
    \bottomrule
    \end{tabularx}
}
\end{table}

\paragraph{Policy refinement with both compensators.}
With both compensators, we repeat base-only refinement for 1,000 PPO epochs using $8\times4{,}096$ environments, base-training configuration, and highest-training-reward checkpoint selection. The stage-one correction policy remains excluded in simulation; real-world evaluation adds it to the refined base output.


\subsection*{Stage 3: Real-World Reinforcement Learning}
\label{sec:real_world_rl_exp_details}

\paragraph{Simulation warm-start.}
Before real-world training, we warm-start the latent modulation actor and critics in simulation for 1,000 epochs using $8\times8{,}192$ parallel environments. The base controller, chunk predictor, and action representation remain fixed.

\paragraph{SAC architecture.}
The modulation actor receives the hand-observation history $\mathbf o_t$ concatenated with the encoded reference latent $\mathbf z_t$. Its MLP has four hidden layers of widths $512$, $512$, $256$, and $128$, with ELU activations. The output parameterizes a diagonal Gaussian over the $d_z$ latent offsets; log standard deviations are clipped to $[-5,2]$, and a $\tanh$ transformation bounds the sampled offsets to $[-1,1]$ before the latent-offset scaling described below. Two independent scalar Q networks use the same hidden-layer widths and activations, taking the concatenated hand-observation history, reference latent, and latent offset $(\mathbf o_t,\mathbf z_t,\Delta\mathbf z_t)$ as input. The critics evaluate the latent action, rather than the decoded joint-action chunk. Object poses are used for rewards and termination, not as actor or critic inputs during real-world training.

\paragraph{Chunk-level transitions and SAC updates.}
Each executed action chunk forms one replay transition. For $n_t\leq T_c$ executed control steps, its reward is $R_t=\sum_{k=0}^{n_t-1}r_{t+k}$, with no discounting within the chunk. The learner applies a reward multiplier of $0.01$. Simulation-based chunk-length selection and real-world optimization use the same per-control-step bootstrap discount $\gamma_{\mathrm{step}}=0.99^{1/8}$. A chunk of $n_t$ executed steps uses $\gamma_{\mathrm{step}}^{n_t}=0.99^{n_t/8}$, preserving the eight-step reference discount $\gamma_8=0.99$. Early termination shortens the executed chunk and its accumulated reward. True terminal transitions disable bootstrapping; time-limit truncations retain it.

After each collection phase, we perform $86$ critic updates using replay mini-batches of $670$ transitions. The actor and entropy temperature update every third critic update, starting with the first update of each phase, giving $29$ actor and temperature updates per phase. Target critics are softly updated after every critic update, with $\bar\theta\leftarrow(1-\tau)\bar\theta+\tau\theta$ and $\tau=0.005$. We optimize the actor, critics, and entropy temperature with Adam. The target entropy is $\mathcal H_{\mathrm{target}}=-0.25d_z$, equal to $-4$ for the default $d_z=16$; the temperature is restored from the simulation warm-start checkpoint and remains learnable. Table~\ref{tab:supp_latent_sac_training} summarizes the settings.

\begin{table}[htbp]
\centering\small
\caption{\footnotesize \textbf{SAC settings for real-world latent-residual optimization.} Update counts refer to one collection--update phase; replay entries are chunk-level transitions.}
\label{tab:supp_latent_sac_training}
\resizebox{0.8\textwidth}{!}{%
\begin{tabularx}{\linewidth}{@{}Xl@{}}
\toprule
Setting & Value\\
\midrule
Actor / critic hidden widths & $512$--$512$--$256$--$128$\\
Hidden activation & ELU\\
Critic ensemble size & $2$\\
Actor distribution & $\tanh$-squashed diagonal Gaussian\\
Actor log-standard-deviation bounds & $[-5,2]$\\
Reference bootstrap discount ($T_c=8$) & $0.99$\\
Bootstrap discount for $n_t$ executed steps & $0.99^{n_t/8}$\\
Reward multiplier for SAC & $0.01$\\
Optimizer / Adam moments & Adam / $(0.9,0.999)$\\
Actor / critic learning rate & $3\times10^{-5}$ / $10^{-4}$\\
Entropy-temperature learning rate & $3\times10^{-5}$\\
Replay mini-batch size & $670$ chunk transitions\\
Critic updates per collection--update phase & $86$\\
Actor / temperature updates per phase & $29$ each\\
Target-critic update interval & Every critic update\\
Target-critic soft-update coefficient $\tau$ & $0.005$\\
Target entropy & $-0.25d_z$ ($-4$ for $d_z=16$)\\
Temperature initialization & Simulation warm-start checkpoint\\
Evaluation checkpoint & Final training update\\
\bottomrule
\end{tabularx}
}
\end{table}

\paragraph{Real-world training protocol and termination.}
The fixed high-frequency controller sums the model-based refined base and stage-one correction policies, then adds the decoded residual action. Each task uses 1,000 collection--update epochs with two rollouts each (2,000 total); these epochs differ from PPO/supervised epochs. Rollouts fail at tracked object height $z<0.35\,\mathrm m$, a low threshold that limits mocap-jitter false detections. Task-specific step limits appear in Table~\ref{tb:max_episodes_per_rollout_in_training}; the 200-step limit for \passvzero and the 1,600-step limits for \trianglepass apply to training for sustained rotation. The budget comparisons in Fig.~\ref{fig:framework_stages} instead use a 100-step limit for \passvzero and a 400-step limit for all three \trianglepass variants.

Learning curves use exploratory training rollouts because separate hardware evaluations are costly. Noise can reduce performance below deterministic evaluation, including an initial fine-tuning dip. Final evaluation uses the checkpoint after the last training update and follows the protocols above.

\begin{table*}[ht]
\centering
\caption{\footnotesize \textbf{Maximum Environment Steps per Training Rollout in the Real-world RL Stage.} The 200-step limit for \passvzero and the 1,600-step limits for \trianglepass are used when training sustained rotation. For the budget comparisons in Fig.~\ref{fig:framework_stages}, \passvzero instead uses a 100-step limit and all three \trianglepass variants use a 400-step limit.}
\label{tb:max_episodes_per_rollout_in_training}
\resizebox{1.0\textwidth}{!}{%
\begin{tabular}{l|cccccccccc}
\toprule
\multirow{1}{*}{Task} & \textbf{\passvzero} & \textbf{\passvone} & \textbf{\passvtwo}  & \textbf{\passvthree} & \textbf{\thumbaround} & \textbf{\sonicdown} & \textbf{\makecell[c]{\textit{Triangle Pass} \\ \texttt{(palm-up)}}} &  \textbf{\makecell[c]{\textit{Triangle Pass} \\ \texttt{(palm-down)}}} &  \textbf{\makecell[c]{\textit{Triangle Pass} \\ \texttt{(palm-down, reversed)}}} &  \textbf{\makecell[c]{\texttt{Four-Finger-Pass} \\ \texttt{(palm-down, LEAP)}}} \\
\midrule

\makecell[c]{\#Maximum Environment Steps \\ per Training Rollout} & 200 & 78  & 78 & 60  & 20 & 400  & 1600  & 1600 & 1600 & 400 \\

\bottomrule
\end{tabular}
}
\end{table*}

\paragraph{Residual scaling and evaluation.}
Unless varied in ablations, all tasks use empirically chosen defaults: latent dimension $d_z=16$, action-residual scale $\alpha_r=0.25$, and pre-decoder latent-offset scale $\beta_r=0.25$ (Eq.~\eqref{eq:supp_latent_residual_chunk}). Evaluation uses posterior means, a shared fixed \texttt{torch.randn} initialization from a fixed seed for reference/modulated decoding, and 16 uniform Euler steps matching training.

\paragraph{Action data for the exploration space.}
For each second-stage policy, the chunk predictor and latent action representation use 200 real-world and 100,000 hybrid-simulation trajectories (Sec.~\nameref{sec:supp_latent_data}). Real-world data retain all rollouts but only pre-termination transitions; simulated trajectories are successful rollouts.

\paragraph{Action-chunk encoder--decoder configuration.}
Table~\ref{tab:supp_action_codec_training} specifies the full-hand Gaussian codec (Eq.~\eqref{eq:supp_latent_posterior}). Encoder and flow-network residual blocks use two linear layers with intermediate ELU, skip addition, then LayerNorm and ELU. The flow network concatenates noisy chunks, sinusoidal time embeddings, and latents. The separate chunk predictor uses the same five-block, width-512 ResMLP and optimization settings as the base BC model (Table~\ref{tab:supp_base_bc_training}), with $T_cd_a$ outputs and the loss in Eq.~\eqref{eq:supp_latent_bc}.

Streaming windows use unique trajectories shuffled with seed $7$ and split 95\%/5\% for training/validation. Fixed coordinate-wise means/population std are estimated from up to 50,000 randomly sampled training windows; std is floored at $10^{-6}$. Real-world and simulated training data are sampled at a $1{:}10$ ratio. Joint encoder/decoder training synchronizes gradients across eight GPUs, validates each epoch, and selects minimum validation loss. Epochs end at the update limit or loader exhaustion.

\begin{table}[htbp]
\centering\small
\caption{\footnotesize \textbf{Reference architecture and training configuration for the action-chunk encoder--decoder.} Deployment retains the same decoding configuration.}
\label{tab:supp_action_codec_training}
\resizebox{0.8\linewidth}{!}{
\begin{tabularx}{\linewidth}{@{}Xl@{}}
\toprule
Setting & Value\\
\midrule
Action representation & Full hand (Sharpa: $22$; LEAP: $16$ joints)\\
Latent encoder & Diagonal Gaussian variational encoder\\
Encoder hidden width / residual blocks & $128$ / $5$\\
Flow-network hidden width / residual blocks & $256$ / $6$\\
Activation / block normalization & ELU / LayerNorm\\
Latent dimension / flow-time embedding dimension & $16$ / $64$\\
KL weight $\beta_{\mathrm{KL}}$ & $10^{-4}$\\
Flow integration at deployment & $16$ uniform Euler steps\\
Optimizer / learning rate & Adam / $3\times10^{-4}$\\
Learning-rate schedule / weight decay & None / $0$\\
Parallel training processes & $8$\\
Batch size per process / nominal global batch & $4096$ / $32{,}768$\\
Training epochs / maximum updates per epoch & $2000$ / $2000$\\
Maximum validation batches per process per epoch & $50$\\
Trajectory-level training / validation split & $95\%$ / $5\%$\\
Normalization sample size & Up to $50{,}000$ training windows\\
Data-loading workers per process & $64$\\
Checkpoint selection & Lowest validation loss\\
\bottomrule
\end{tabularx}
}
\end{table}

\paragraph{Selecting the exploration chunk size.}
We compare chunk lengths $T_c\in\{4,8,16\}$ with $d_z=16$ fixed, training residual latent policies in 2 simulation environments for up to 1,000 epochs with actuator P-gains scaled by $0.8$. As in real-world optimization, the bootstrap discount is $\gamma_{T_c}=0.99^{T_c/8}$, anchored at $\gamma_8=0.99$. Best final performance selects: 8 for the \pass family, \sonicdownns, \trianglepassdownreversedns, and \fourfingerpassleapns; 4 for \thumbaroundns; 16 for \trianglepassupns, \trianglepassdownns, and Sharpa board rotation.

\paragraph{Task rewards.}
\noindent\textit{Pass tasks.}
The \pass family (\passvzerons, \passvonens, \passvtwons, \passvthreens) retains simulation goal updates but adopts a denser goal arrangement (15-degree spacing) and a lower goal-reaching bonus (125 points). It also uses sparse keypoint-orientation rewards to avoid noisy mocap angular velocities. For local keypoints $\mathbf k_A,\mathbf k_B$, unprojected coordinates are $\widetilde{\mathbf x}_j(q)=R(q)\mathbf k_j$. When using an $xy$, $yz$, or $xz$ projection $P$, set $\mathbf x_j(q)=PR(q)\mathbf k_j$, midpoint $\mathbf m=(\mathbf x_A+\mathbf x_B)/2$, segment $\mathbf v=\mathbf x_B-\mathbf x_A$, and original length $L=\|\mathbf k_B-\mathbf k_A\|_2$. Rescale about the midpoint:
\[
\begin{aligned}
\widetilde{\mathbf x}_{A,B}(q)&=\mathbf m(q)\mp\frac{L}{2}\frac{\mathbf v(q)}{\max(\|\mathbf v(q)\|_2,10^{-6})},\\
 d_{\mathrm{kp}}(q,q^\star)&=10\sum_{j\in\{A,B\}}\|\widetilde{\mathbf x}_j(q)-\widetilde{\mathbf x}_j(q^\star)\|_2.
\end{aligned}
\]
The empirical threshold $d_{\mathrm{kp}}\leq0.3$ triggers goal advancement and reduces missed approximate matches. Let $q_t^\star$ denote the current target before advancement. The per-control-step reward is
\[
    G_t=\mathbb I\!\left[d_{\mathrm{kp}}(q_t,q_t^\star)\leq0.3\right],
    \qquad r_t=125G_t.
\]
Each reached target receives the bonus once, after which the target advances; other control steps receive zero reward. For a chunk with $n_t$ executed control steps, the reward is the undiscounted sum $R_t=\sum_{k=0}^{n_t-1}r_{t+k}=125\sum_{k=0}^{n_t-1}G_{t+k}$. This is the task reward before the SAC reward multiplier specified in Table~\ref{tab:supp_latent_sac_training}. Longitudinal-axis keypoints ignore axial spin, accommodating capsule/cylinder symmetry without canonical axial initialization.

\noindent\textit{Triangle Pass and four-finger passing.}
For \trianglepass and Sharpa/LEAP four-finger passing, successive targets are $30^\circ$ apart about the prescribed $z$ direction. With XYZW quaternions, $s\in\{-1,+1\}$ and $Q_z(\theta)$ denoting a $z$ rotation, initialize $\mathbf q_{g,0}=Q_z(s\pi/6)\otimes(0,0,0,1)$. Advance by $Q_z(s\pi/6)$ after each success. The same longitudinal-endpoint criterion gives $S_t=\mathbb I[d_{\mathrm{end},t}<\epsilon_{\mathrm{goal}}]$ and sparse reward $r_t^{\mathrm{goal}}=6S_t$.

\noindent\textit{\thumbaround and \sonicdownns.}
For \thumbaroundns, a human assigns one success/failure label after each rollout; all executed chunks are then retrospectively assigned rewards of 50/0, respectively. A 20-step rollout with four-step chunks thus has an undiscounted task return of 250/0 before reward scaling.
For \sonicdownns, reward is the simulation's signed conical-rotation measure $\Omega_t$.
For \sonicdownns, per-rollout episode-completion values were not logged during real-world training. We therefore visualize training progress using an approximate reward-derived proxy, \(p=\min(R/500,1)\), where \(R\) denotes the return value plotted in the learning curve and 500 is an empirically chosen reference return associated with one complete maneuver. Because \sonicdownns is a noncyclic task containing at most one maneuver per rollout, this proxy is capped at 1. It is intended only as an approximate indicator of training progress and should not be interpreted as a directly measured episode count or success rate. This proxy is distinct from the separately scored evaluation results.

For each epoch, the reward and episode values shown in the real-world training curves in the main text and the Supplementary Section~S3 are taken from the $(2\times\text{epoch})$-th rollout during training.
Dashed curves show a reward-derived progress proxy, as actual episode counts were not recorded during the experiments. For the \pass and \trianglepass families and \fourfingerpassleapns, the converted values provide reasonably accurate estimates of episode counts because their rewards depend on the number of goals reached. For \sonicdownns, the conversion is more approximate, using an empirically observed reward from a separate, fully completed episode as a reference. Since this task is non-cyclic, we cap the resulting value at 1.

\subsection*{Baseline Configurations}
\label{sec:supp_baselines}

\noindent\textit{Common protocol.}
NDM starts from our first-stage controller (base plus stage-one correction); LR-TD3, BPG-SAC, SERNF, and RFS use our second-stage controller or its trajectories as their starting point. All four RL baselines undergo simulation training before real-world adaptation with 2,000 hardware rollouts per task. LR-TD3, BPG-SAC, and SERNF update online during hardware training; RFS uses a fixed exploration dataset for offline TD3+BC. All four RL baselines match our large-scale simulation pretraining setting of $8\times8{,}192$ parallel environments and 1,000 epochs; additional simulation-only parameter searches are described below. LR-TD3, SERNF, and RFS use the same datasets as our third-stage action models, training separate reference flows or policies on simulated and real-world data for use in the corresponding domain. BPG-SAC uses the second-stage controller for initialization and exploration guidance. Their hand-only observation contains ten frames of measured joints and commanded targets ($20n_{\mathrm{dof}}$ dimensions).
Beyond the shared S1 or S1+S2 starting controller and specified budgets, we follow each paper's data-collection, modeling, training, and reward settings as closely as possible. Method-specific adaptations are documented below.

\noindent\textit{Simulation-based conservative tuning.}
For BPG-SAC, RFS, and the joint-space residual baseline in Fig.~\ref{fig:structural_and_residual_exploration}(A), we first calibrate exploration noise in simulation before large-scale pretraining. With the starting controller fixed, we select a noise scale that retains at least 90\% of its noise-free mean return. After the $8\times8{,}192$-environment pretraining, each additional parameter-search run starts from the pretrained policy and critic checkpoints. BPG-SAC and joint-space residuals use SAC with two simulation environments for 1,000 epochs. For RFS, two environments collect a fixed simulated exploration dataset, followed by 1,000 offline TD3+BC training epochs, matching its subsequent real-world adaptation algorithm.

We sweep critic LayerNorm and actor-update frequency, together with method-specific exploration and update parameters. For SAC, we also sweep target entropy and the onset of joint-space exploration noise at epoch $\{0,50,100,150,200\}$ of the additional simulation run. An added actor penalty discourages deviation from the corresponding fixed base controller's commands, following the conservative regularization principle of TD3+BC; this is distinct from RFS's BC term on recorded residual actions. The selected hyperparameters are then used for real-world adaptation, with method-specific initialization as described below. These additional searches aim to limit contact-disrupting exploration and abrupt policy changes; more aggressive settings caused collapse in our experiments. Figures~\ref{fig:rollout_figure}(D) and~\ref{fig:structural_and_residual_exploration}(A) report hardware results using simulation-selected settings.

Our latent-residual method uses shared empirical defaults for latent dimension and residual scales, with task-specific chunk lengths selected in simulation. The comparisons evaluate the specified implementations under matched real-world rollout budgets, while leaving open potential improvements from alternative implementations or tuning strategies.

\paragraph{Joint-wise neural dynamics (NDM).}
Starting from our first-stage controller, we apply two DexNDM~\cite{Liu2025DexNDMCT} refinement iterations. Each iteration uses 200 real-world rollouts and 100,000 simulated trajectories to train its joint-wise dynamics and transition-matching residual models. Following DexNDM, we collect real-world hand transitions in a Chaos Box with randomized object loads, recording hand proprioception and commanded actions without motion-capture object states. Each iteration adapts the current controller before the next round of collection. We retain joint-wise modeling and joint-transition alignment to preserve DexNDM's proprioceptive formulation, which captures coupling and external-load effects indirectly through local joint histories.

The ``w/ delta-action model'' variant in Fig.~\ref{fig:model_based_abl}(A) additionally evaluates NDM-style joint-wise compensation using our mocap-annotated task trajectories. Its corrections retain local joint inputs but are trained through a differentiable surrogate with joint-position and unsigned object-axis supervision (Sec.~\nameref{sec:workflow_delta_action_model}). It follows NDM's joint-wise modeling principle while using our simulator-side delta-action training and policy-refinement procedure, with learned contact compensation disabled.

\paragraph{Latent-Residual TD3 (LR-TD3).}
We evaluate a state-based adaptation of LR-TD3~\cite{Zhu2026TowardsHD} that replaces its vision--language backbone with a proprioceptive flow-matching reference, so that it receives the same proprioceptive inputs and task datasets as our method and the other baselines.
The actor additionally observes current joints ($n_{\mathrm{dof}}$ dimensions) and reference latents; critics use the same hand inputs and corrected latents. Hand inputs use fixed offline normalization.

\noindent\textit{Offline training.}
The codec uses the matched datasets described above. We train two reference flow policies separately: one on simulated data only and one on real-world data only, to improve reference-policy performance in the corresponding domain. A temporal variational codec represents target offsets from initial joints, which condition both encoder and decoder. Joint-weighted Huber reconstruction penalizes offsets and first/second temporal differences with weights $(1,0.2,0.05)$; ring and little joints have weight $0.5$, others $1$. KL increases linearly from zero to $3\times10^{-4}$ over updates $4{,}000$--$8{,}000$. With the codec frozen, each reference flow learns posterior means through latent-velocity matching and decoded-command reconstruction (weights $0.25$ and $0.75$). Training and validation use a 90/10 trajectory split; checkpoints minimize held-out joint-command error, and inference uses EMA flow parameters.

\noindent\textit{Residual optimization.}
With codec and flow frozen, a zero-initialized actor bounds corrections to the first $C$ latent frames, leaving $H-C$ unchanged. Brownian exploration noise is added after bounding. Corrected latents are clipped by coordinate-wise 99.5th-percentile absolute training posterior means, decoded to absolute targets, and executed for $C$ frames before replanning, subject to joint and per-frame change limits. The actor minimizes
$
    \mathcal{L}_{\mathrm{actor}}
    =
    \mathbb{E}\left[
        -0.5\,Q_1(\mathbf{s}_t,\mathbf{Z}_t)
        +3\|\Delta\mathbf{Z}_t\|_F^2
        +0.5\,\mathcal{S}(\mathbf{Z}_t)
    \right],
$
where $\mathbf Z_t=\mathbf Z_t^{\mathrm{ref}}+\Delta\mathbf Z_t$ precedes execution clipping, $\mathbf s_t$ contains hand inputs, and $\mathcal S$ averages squared $\ell_2$ norms of second temporal differences over the full horizon. Gradients pass through latent-space critics. Reference dropout zeros only actor reference conditioning. Minimum twin-target values bootstrap with $\gamma^n$ for $n$ executed frames, excluding terminal transitions; actor updates accompany every joint-training critic update, with zero target-policy smoothing.

\noindent\textit{Simulation warm-start and real-world fine-tuning.}
We warm-start the residual latent actor and critics in simulation using the fixed simulation-trained reference flow, then transfer them to hardware and switch to the fixed real-data-trained reference flow for continued optimization. The codec remains frozen throughout. On hardware, each labeled episode triggers 200 updates. Replay draws from the latest five episodes with newest-first weights $6{:}5{:}4{:}3{:}2$ and at least one newest sample per batch. Intermediate rewards are zero; terminal success and failure rewards are $0$ and $-1$. Both outcomes are retained with manual labels. Horizons are $(H,C)=(32,25)$ except $(5,5)$ for \thumbaroundns; remaining settings appear in Table~\ref{tb:lr_td3_hyperparameters}.

\begin{table*}[htbp]
    \centering
    \small
    \setlength{\tabcolsep}{6pt}
    \caption{ \footnotesize \textbf{Default LR-TD3 hyperparameters.} Prediction and execution horizons are configurable and inherited from the reference checkpoint. Offline batch sizes are global defaults.}
    \label{tb:lr_td3_hyperparameters}
\resizebox{0.9\linewidth}{!}{
    \begin{tabular}{@{}lll@{}}
        \toprule
        Component & Parameter & Value \\
        \midrule
        Representation & Prediction / execution horizon $(H,C)$ & $32 / 25$ or $5 / 5$ \\
                       & Latent dimension per frame & $9$ \\
        Codec          & Encoder and decoder: layers / width / heads & $3 / 128 / 4$ \\
                       & Feed-forward width & $256$ \\
                       & Optimizer / learning rate & AdamW / $3\times10^{-4}$ \\
                       & Batch size / training updates & $4096 / 30{,}000$ \\
        Reference flow & Layers / width / heads & $4 / 256 / 8$ \\
                       & Feed-forward width & $1024$ \\
                       & Optimizer / learning rate & AdamW / $5\times10^{-5}$ \\
                       & Batch size / training updates & $64 / 30{,}000$ \\
                       & EMA decay / Euler integration steps & $0.99 / 10$ \\
        Residual RL    & Simulation warm-start: environments / epochs & $(8\times8{,}192) / 1{,}000$ \\
                       & Input projections / trunk & Three $256$-D projections; two $256$-unit layers \\
                       & Hidden-layer normalization / activation & LayerNorm / SiLU \\
                       & Optimizer & Adam, zero weight decay \\
                       & Actor / critic learning rate & $10^{-4} / 5\times10^{-5}$ \\
                       & Batch size & $128$ \\
                       & Per-frame discount / target update rate & $0.99 / 0.005$ \\
                       & Deterministic residual bound / noise scale & $0.10 / 0.05$ \\
                       & Reference-conditioning dropout & $0.5$ \\
                       & Replay capacity / learning-start threshold & $200{,}000 / 500$ chunk transitions \\
                       & Maximum joint-target change per frame & $\pi/18$ rad \\
        \bottomrule
    \end{tabular}
}
\end{table*}

\noindent\textit{Observed performance.}
The reference flow exhibits execution drift on hardware before RL, despite sweeps of prediction and execution horizons; subsequent real-world fine-tuning does not recover successful execution. Reduced feedback responsiveness and distortion of fine-grained commands may contribute, consistent with our action-interface and full-action exploration comparisons (Fig.~\ref{fig:structural_and_residual_exploration}).

\paragraph{Base-Policy-Guided SAC (BPG-SAC).}
We adapt controller-guided exploration~\cite{Kktabak2026RealWorldRL} using our second-stage controller as the guide. The central principle is assistance from a capable controller, which can be an MPC controller or a learned policy. Designing a separate MPC controller for our dynamic, contact-sensitive tasks would require substantial additional modeling and control design, while direct hardware transfer would still face simulation mismatch. Our second-stage controller already provides useful real-world behavior.

\noindent\textit{Initialization and guidance.}
In simulation, both the fixed guide and the initial SAC actor copy the post-S2 base policy, without the stage-one correction policy. Before online simulation training, we warm up the critic ensemble for 50,000 offline updates on simulated data. On hardware, the fixed guide uses the post-S2 base policy together with the stage-one correction policy; the trainable controller is initialized as a copy of this guide. Only its base-policy component is optimized, with the correction policy fixed. The critic ensemble is transferred from simulation. The trainable SAC base actor predicts joint targets at every step through a residual-MLP diagonal Gaussian, $\tanh$, and affine joint-bound mapping. The reference critic architecture uses two 256-unit layers with LayerNorm and $\tanh$; normalization is selected through the simulation sweep.

Minibatches mix approximately 30\% retained-prior and 70\% online transitions from either controller. Targets use the minimum of two randomly selected target critics; the actor optimizes the full-ensemble mean with learned entropy temperature. Its learning rate ramps linearly. SAC selection probability is

\begin{equation}
    p_{\mathrm{SAC}}(t)
    =
    0.5 + 0.5\min\left(\frac{t}{6000},1\right).
\end{equation}
Selection lasts 40 steps with fresh inference each step. Guide actions receive no extra noise; SAC samples its actor. Hardware evaluation uses the deterministic SAC base actor plus the fixed correction policy, without guide selection.

On hardware, the reward is $20\Delta\theta_t-0.3$, where $\Delta\theta_t$ is the mocap-derived increment about the desired signed axis, with terminal success bonus $200$ or final reward replaced by $-50$ on failure. An operator labels outcomes; invalid or stale tracking terminates as failure. Settings are in Table~\ref{tb:bpg_sac_hyperparameters}.

\begin{table*}[htbp]
    \centering
    \small
    \setlength{\tabcolsep}{6pt}
    \caption{
    \footnotesize
    \textbf{Reference BPG-SAC settings.} Hardware training uses values selected by the simulation-only sweeps described in the text.}
    \label{tb:bpg_sac_hyperparameters}
    \begin{tabular}{@{}lc@{}}
        \toprule
        Parameter & Value \\
        \midrule
        Simulation warm-start: environments / epochs & $(8\times8{,}192) / 1{,}000$ \\
        Actor mean network & ResMLP \\
        Critic ensemble size & $10$  \\
        Critic hidden layers & $256$--$256$  \\
        Initial pre-squash action std. & $0.3$\\
        Discount $\gamma$ & $0.995$  \\
        Target update rate $\tau$ & $0.01$  \\
        Optimizer & Adam \\
        Actor / critic learning rate & $3\times10^{-4}$ / $3\times10^{-5}$ \\
        Temperature learning rate & $10^{-4}$ \\
        Initial temperature & $0.05$ \\
        Target entropy & $-22$ (Sharpa), $-16$ (LEAP) \\
        Soft-SARSA temperature & $0.02$  \\
        Actor learning-rate ramp & $200{,}000$ updates  \\
        Batch size & $256$ \\
        Prior / online replay fraction & $30\%$ / $70\%$  \\
        Actor initialization in each domain & Copy of guidance policy \\
        Trainable controller component & Base policy only \\
        Critic warmup (simulated data) & $50{,}000$ offline updates
                       \\
        SAC updates per interaction step & $8$ per control step  \\
        Guidance decay duration & $6{,}000$ control steps\\
        Retained-prior capacity & All successful demonstrations  \\
        Online replay capacity & $1{,}000{,}000$ transitions  \\
        \bottomrule
    \end{tabular}
\end{table*}

\noindent\textit{Parameter sweeps.}
Following the simulation-only tuning protocol above, we test initial pre-squash action standard deviations $\{0.15,0.10,0.05,0.005,0.0005\}$, initial SAC-selection probabilities $\{0.25,0.10,0.05,0.01,0.001\}$, guidance durations $\{12{,}000,24{,}000\}$ control steps, maximum actor learning rates $\{10^{-4},\allowbreak 3\times10^{-5},\allowbreak 3\times10^{-6},\allowbreak 10^{-6}\}$, and $\{4,2,1\}$ updates per interaction step. These supplement the target-entropy, noise-onset, critic-LayerNorm, and actor-update-frequency sweeps. Hardware training uses the simulation-selected values.

\noindent\textit{Observed performance.}
Training succeeds in simulation. Exploration can still disrupt contacts and cause drops despite controller guidance, potentially limiting useful learning signals and further improvement.

\paragraph{SERNF.}
Our adaptation of SERNF~\cite{Yang2026SERNFSR} uses object poses only for rewards/termination. Its conditional RealNVP~\cite{Dinh2016DensityEU} maps an $n_{\mathrm{dof}}H$-dimensional Gaussian to $H$ absolute-target actions, using attention-conditioned affine couplings and final $\tanh$ bounds. Observation statistics and action bounds are fitted on BC training data. Synchronous execution applies the first $E\le H$ actions before replanning.

\noindent\textit{Initialization and optimization.}
We train separate BC policies on simulated data only and real-world data only. Starting from the simulation-trained policy, we perform 3,000 offline critic-warmup updates followed by 1,000 offline RL updates on simulated data before online simulation training. BC maximizes exact action-chunk likelihood, reserving 10\% of trajectories for validation; offline RL uses both outcomes. During critic warmup, the actor is frozen while two HL-Gauss critics learn discounted $E$-step returns, using distribution expectations as $Q$-values. Terminal chunks use their actual durations without bootstrapping. The actor then minimizes
$
    \mathcal{L}_{\pi}
    =
    -\mathbb{E}_{\mathbf{o}\sim\mathcal{D}}
    \left[
        \min_{j\in\{1,2\}}
        Q_{\phi_j}
        \bigl(\mathbf{o},\hat{\mathbf{A}}_{0:E-1}\bigr)
    \right]
    -
    \lambda
    \mathbb{E}_{(\mathbf{o},\mathbf{A})\sim\mathcal{D}_{\mathrm{off}}}
    \left[
        \log \pi_{\theta}(\mathbf{A}\mid\mathbf{o})
    \right].
$
where $\hat{\mathbf A}$ is the best of 24 samples under the minimum critic value, with gradients through that sample. Pure BC instead selects maximum likelihood among 64 samples. Candidate noise has standard deviation $0.7$; Bellman targets use one standard-Gaussian actor sample.

\noindent\textit{Simulation warm-start and real-world fine-tuning.}
After offline warmup, we train the simulation policy and critics through online RL under the common warm-start budget. Hardware fine-tuning starts from the real-data-trained policy and the transferred simulation critics; we first warm up these critics offline on real-world data with the actor fixed, then jointly fine-tune the policy and critics. Collection and updates alternate; overlapping replay windows never cross episode boundaries. Offline sampling probability is $p_{\mathrm{off}}=wN_{\mathrm{off}}/(wN_{\mathrm{off}}+N_{\mathrm{on}})$ for replay sizes $N_{\mathrm{off}},N_{\mathrm{on}}$ and $w=0.5$. Imitation uses only offline data. On hardware, drops or invalid tracking terminate execution even within chunks.

Sparse rewards use our goal initialization/advancement and $+1$ reaching bonus for \passns, \trianglepassns, and \fourfingerpassns. For \thumbaroundns, each reached goal advances $30^\circ$ about $(0,0,1)$. In each 400-step \sonicdown episode with $s\in\{0,1,2\}$ completed slot transitions, a chunk receives $+1$ only when fully contained in the first $200s$ steps, otherwise zero; all rewards are zero for $s=0$.

Horizon selection starts at $H=E=10$, reducing it for poor distillation: final $(H,E)=(2,1)$ for \thumbaround (single-step feedback) and $(10,10)$ otherwise. Real-world fine-tuning uses 20 rounds of 100 rollouts, grouping collection to approach the original 15-minute duration. Maximum lengths match Table~\ref{tb:max_episodes_per_rollout_in_training}; settings appear in Table~\ref{tb:sernf_hyperparameters}.

\begin{table}[htbp]
    \centering
    \small
    \setlength{\tabcolsep}{4pt}
    \caption{\footnotesize
    \textbf{Default hyperparameters of our SERNF implementation.}}
    \label{tb:sernf_hyperparameters}
    \begin{tabular}{@{}p{0.58\linewidth}p{0.22\linewidth}@{}}
        \toprule
        Parameter & Value \\
        \midrule
        Prediction / execution horizon $(H,E)$ & $(10,10)$ or $(2,1)$ \\
        Flow coupling layers / width / attention heads & $16 / 256 / 8$ \\
        Number of critics & $2$ \\
        Critic layers / width / attention heads & $3 / 512 / 8$ \\
        Critic feed-forward width & $256$ \\
        HL-Gauss bins / support & $101 / [0,100]$ \\
        HL-Gauss smoothing standard deviation & $0.75\times$ bin width \\
        Discount $\gamma$ / target update rate $\tau$ & $0.997 / 0.05$ \\
        Optimizer / weight decay & AdamW / $10^{-4}$ \\
        BC / RL learning rate & $10^{-4} / 2\times10^{-5}$ \\
        Batch size / gradient norm limit & $512 / 10$ \\
        Offline / online imitation weight $\lambda$ & $0.1 / 0.1$ \\
        Conditioning dropout: BC / RL & $0.2 / 0.1$ \\
        Critic dropout & $0.1$ \\
        Demonstration action-noise standard deviation & $0.01$ (normalized) \\
        BC updates & $200{,}000$ \\
        Simulation critic warmup / offline RL updates & $3{,}000 / 1{,}000$ \\
        Simulation warm-start: environments / epochs & $(8\times8{,}192) / 1{,}000$ \\
        Real-world collection--update rounds & $20$ \\
        Collection / updates per real-world round & 100 rollouts / $1{,}000$ \\
        Online replay capacity & $200{,}000$ entries \\
        \bottomrule
    \end{tabular}
\end{table}

\noindent\textit{Observed performance.}
Simulation training produces useful behavior, and the simulation-trained policy with best-of-$N$ selection yields some successful trials. 
After real-world fine-tuning, success exceeds the second-stage controller's on \passvzero and the four slower tasks but stays below it on the other five dynamic tasks. The $n_{\mathrm{dof}}H$-dimensional action-chunk space may make exploration inefficient under the hardware budget, contributing to this gap.

\paragraph{Residual Flow Steering (RFS).}
We adapt RFS~\cite{Su2026RFS} to our hand-only observations and joint-target actions. We train two single-step flow-matching policies separately on the post-S2 datasets: one on 100,000 hybrid-simulation trajectories and the other on 200 real-world trajectories per task. With the simulation-trained flow fixed, we jointly train the noise-steering and action-residual outputs using online PPO for 1,000 epochs in $8\times8{,}192$ parallel simulation environments. The subsequent two-environment parameter search uses offline TD3+BC on simulated exploration data to match the real-world adaptation algorithm. RFS uses the same task-specific rewards as our method in each domain; real-world rewards are specified under \textit{Task rewards} above.

At each control step, one output steers the flow policy through its initial noise, following the latent-noise steering principle of DSRL~\cite{Wagenmaker2025SteeringYD}, while another supplies a joint-action residual. We evaluate their combination under the RFS PPO/TD3+BC protocol, without isolating the noise branch. Jointly predicting these two action-dimensional vectors requires $2n_{\mathrm{dof}}$ outputs on the $n_{\mathrm{dof}}$-DoF hand. Our implementation operates at each control step without an explicit temporal action representation; the additive action-residual branch has neither a learned spatial nor temporal action prior.

\noindent\textit{Real-world data collection and offline adaptation.}
For hardware adaptation, we switch to the fixed real-data-trained flow policy and initialize the residual model and critic from simulation training. Human takeover does not achieve effective recovery in our setup (Fig.~\ref{fig:human_takeover_failures}), so we collect 2,000 exploratory real-world rollouts per task using this initialized controller, then perform offline TD3+BC on the fixed dataset. Transitions record the initial noise, policy-generated action residual, executed action, reward, and next observation. The critic is conditioned on the final action $a_b+a_r$, while the actor optimizes its Q-value with BC regularization on the recorded policy residual $a_r$. The initial-noise output has no separate BC target. Policy exploration thus supplies the adaptation data in place of expert corrections.

The reference offline schedule comprises 1,000 update rounds, each with 86 critic gradient updates, for 86,000 critic updates in total, with one policy update every ten critic updates. We evaluate the checkpoint after the final update. Table~\ref{tb:rfs_offline_hyperparameters} lists reference settings following RFS~\cite{Su2026RFS}; the simulation-selected learning rate, residual scale, actor-update interval, and target-smoothing parameters are used for hardware adaptation.

\begin{table}[htbp]
    \centering\small
    \caption{\footnotesize\textbf{Reference offline TD3+BC settings for RFS.} Hardware adaptation uses simulation-selected hyperparameters and the same fixed dataset of 2,000 real-world rollouts per task. Update rounds do not collect additional rollouts.}
    \label{tb:rfs_offline_hyperparameters}
    \begin{tabularx}{0.6\linewidth}{@{}lX@{}}
        \toprule
        Parameter & Value \\
        \midrule
        Actor / critic learning rate & $3\times10^{-4}$ / $3\times10^{-4}$ \\
        Policy update delay & 10 critic updates \\
        Target policy noise standard deviation & 0.2 \\
        Target noise clip & 0.5 \\
        BC weight $\lambda_{\mathrm{BC}}$ & 0.2 \\
        Batch size & 512 transitions \\
        Replay buffer & Fixed offline dataset \\
        Offline update rounds & 1,000 \\
        Critic gradient updates per round & 86 \\
        Total critic gradient updates & 86,000 \\
        Evaluation checkpoint & Final training update \\
        \bottomrule
    \end{tabularx}
\end{table}

\noindent\textit{Parameter sweeps.}
In the additional simulation-only TD3+BC stage, we test residual-scale multipliers $\{0.5,\allowbreak 0.25,\allowbreak 0.1,\allowbreak 0.01\}$ relative to the reference scale, actor learning rates $\{10^{-4},\allowbreak 3\times10^{-5},\allowbreak 3\times10^{-6},\allowbreak 10^{-6}\}$, and actor-update intervals of $\{20,40,80\}$ critic updates, together with the critic-LayerNorm sweep and base-distance penalty described above. Target-smoothing standard deviation/clip pairs are $\{(0.1,0.25),\allowbreak (0.05,0.1),\allowbreak (0.01,0.02),\allowbreak (0.005,0.01)\}$; these affect TD3 critic targets and are separate from execution noise. Alongside the reference endpoint of 86,000 critic updates, we examine intermediate offline-adaptation checkpoints after $10$, $100$, $1{,}000$, $10{,}000$, $20{,}000$, and $40{,}000$ updates for deterioration. Hardware adaptation uses the simulation-selected hyperparameters.

\noindent\textit{Observed performance.}
Residual training succeeds in large-scale parallel simulation. Limited coverage of useful corrections in the fixed real-world exploration dataset may still hinder offline policy improvement.

\subsection*{Generalization, Adaptation, and Human Disturbances}
\label{sec:supp_generalization_adaptation}

Figure~\ref{fig:few_shot_zero_shot_tasks} summarizes these tests.

\paragraph{Zero-shot generalization.}
Table~\ref{tab:supp_zero_shot_generalization} distinguishes single-episode success evaluations from successful demonstrations without formal repeated-trial assessment and groups changes by contact geometry, compliance, friction, mass distribution, curvature, and weight.

\begin{table*}[!htbp]
\centering
\small
\caption{\footnotesize
\textbf{Zero-shot generalization across physical variations.} Groups highlight the primary properties changed by each test. Success rates are measured over the listed numbers of evaluation trials. 95\% CI: Wilson score interval for the success rate. ``Successful demo'' denotes successful execution demonstrated without formal repeated-trial success-rate evaluation; no success rate is assigned to these entries.}
\label{tab:supp_zero_shot_generalization}
\resizebox{0.75\textwidth}{!}{%
\begin{tabularx}{\textwidth}{@{}>{\raggedright\arraybackslash}p{0.24\textwidth}>{\raggedright\arraybackslash}Xllc@{}}
\toprule
Task & Test condition & Result & 95\% CI & Trials \\
\midrule
\multicolumn{5}{@{}p{\textwidth}@{}}{\textbf{Surface geometry:} Changes in surface flatness and contact patch size.} \\
\passvzerons & Cylinder $\rightarrow$ octagonal prism & $100\%$ & $72$\%--$100\%$ & 10 \\
\passvzerons & Cylinder $\rightarrow$ hexagonal prism & $70\%$ & $48$\%--$85\%$ & 20 \\
\passvonens & Cylinder $\rightarrow$ octagonal prism & $50\%$ & $30$\%--$70\%$ & 20 \\
\passvtwons & Capsule $\rightarrow$ hexagonal prism & $60\%$ & $39$\%--$78\%$ & 20 \\
\passvtwons & Capsule $\rightarrow$ octagonal prism & Successful demo & --- & --- \\
\passvthreens & Cylinder $\rightarrow$ hexagonal prism & $100\%$ & $72$\%--$100\%$ & 10 \\
\passvthreens & Cylinder $\rightarrow$ octagonal prism & $80\%$ & $58$\%--$92\%$ & 20 \\
\midrule
\multicolumn{5}{@{}p{\textwidth}@{}}{\textbf{Compliance:} Increased object compliance and altered contact response.} \\
\passvzerons & Rigid cylinder $\rightarrow$ compliant TPU cylinder & $90\%$ & $70$\%--$97\%$ & 20 \\
\passvonens & Rigid cylinder $\rightarrow$ compliant TPU cylinder & Successful demo & --- & --- \\
\passvtwons & Rigid capsule $\rightarrow$ compliant TPU cylinder & Successful demo & --- & --- \\
\passvthreens & Rigid cylinder $\rightarrow$ compliant TPU cylinder & $55\%$ & $34$\%--$74\%$ & 20 \\
\midrule
\multicolumn{5}{@{}p{\textwidth}@{}}{\textbf{Friction:} Reduced contact friction and increased susceptibility to slipping.} \\
\trianglepassupns & Lower-friction object & $94\%$ (47/50) & $84$\%--$98\%$ & 50 \\
\trianglepassupns & Friction-enhancing fingertip wrap removed & $60\%$ & $42$\%--$75\%$ & 30 \\
\midrule
\multicolumn{5}{@{}p{\textwidth}@{}}{\textbf{Mass distribution:} Shifted center of mass and altered gravitational torques.} \\
\trianglepassupns & L-shaped object, T-shaped object, and long board & Successful demo & --- & --- \\
\trianglepassdownns & L-shaped and T-shaped objects & Successful demo & --- & --- \\
\fourfingerpassleapns & L-shaped and T-shaped objects & Successful demo & --- & --- \\
\midrule
\multicolumn{5}{@{}p{\textwidth}@{}}{\textbf{Surface curvature:} Varying local contact curvature.} \\
\fourfingerpassleapns & C-shaped object & Successful demo & --- & --- \\
\midrule
\multicolumn{5}{@{}p{\textwidth}@{}}{\textbf{Mass:} Substantially increased object weight.} \\
\trianglepassdownns & $100\,\mathrm{g}$ book-like object; training range: $10$--$20\,\mathrm{g}$ & Successful demo & --- & --- \\
\bottomrule
\end{tabularx}
}
\end{table*}

The nominal \passvzero cylinder has radius $0.7\,\mathrm{cm}$ and length $20\,\mathrm{cm}$; prism opposite-face widths are $1.4\,\mathrm{cm}$. Separate continuous demonstrations reach four episodes on hexagonal prisms, five on octagonal prisms, and four on TPU cylinders. Wrap removal exposes lower-friction nails. The heavy \trianglepassdown book measures $15\times2\times20\,\mathrm{cm}$. Palm-up board result is zero-shot demonstration. 

\paragraph{Adaptation with additional real-world rollouts.}
We adapt policies to eight variants of object dimensions, shape, and compliance using 250--500 additional real-world rollouts per variant. 
Table~\ref{tab:supp_few_shot_adaptation} reports pre- and post-adaptation success rates for seven variants and sustained execution for the long-board variant; initial rates range from below $10\%$ to approximately $50\%$.

\begin{table*}[htbp]
\centering
\small
\setlength{\tabcolsep}{4pt}
\renewcommand{\arraystretch}{1.12}
\caption{
\footnotesize\textbf{Adaptation to object variations.} Each variant uses 250--500 additional real-world training rollouts. Success rates are shown before and after adaptation; $\approx$ marks approximate initial rates. Successful/evaluated episode counts refer to post-adaptation evaluations. 95\% CI: Wilson score interval for the post-adaptation success rate.
For the long-board condition, 45 evaluated episodes were completed consecutively within a single uninterrupted run. This result measures sustained execution rather than success across independently initialized trials; no binomial confidence interval is reported.
}
\label{tab:supp_few_shot_adaptation}
\resizebox{0.9\textwidth}{!}{%
\begin{tabularx}{\textwidth}{@{}>{\raggedright\arraybackslash}p{0.22\textwidth}>{\raggedright\arraybackslash}Xrccc@{}}
\toprule
Task & Object change & \makecell[c]{Additional\\rollouts} & \makecell[c]{Before\\adaptation} & \makecell[c]{After\\adaptation} & \makecell[c]{95\%\\CI} \\
\midrule
\passvzerons & Thicker cylinder (diameter $1.4\rightarrow1.6\,\mathrm{cm}$) & 400 & $\approx50\%$ & $100\%$ (50/50) & $93$\%--$100\%$ \\
\thumbaroundns & Rigid object $\rightarrow$ compliant TPU object & 300 & $<10\%$ & $96\%$ (48/50) & $87$\%--$99\%$ \\
\fourfingerpassleapns & Cuboid $\rightarrow$ elliptical cylinder & 500 & $\approx40\%$ & $100\%$ (50/50) & $93$\%--$100\%$ \\
\trianglepassdownns & Long board ($15\times2\times5\,\mathrm{cm}$) & 500 & up to 1 episode & 45 episodes & - \\
\trianglepassupns & Thinner object (width $2.0\rightarrow1.5\,\mathrm{cm}$) & 250 & $<10\%$ & $100\%$ (25/25) & $87$\%--$100\%$ \\
\trianglepassupns & Cuboid $\rightarrow$ cylinder & 500 & $\approx30\%$ & $100\%$ (40/40) & 91\%--100\% \\
\trianglepassupns & Cuboid $\rightarrow$ octagonal prism & 250 & $\approx40\%$ & $100\%$ (30/30) & 89\%--100\%  \\
\trianglepassdownns & Cuboid $\rightarrow$ octagonal prism & 500 & $\approx50\%$ & $100\%$ (15/15) & $80$\%--$100\%$ \\
\bottomrule
\end{tabularx}
}
\end{table*}

The adapted \trianglepassdown policy rotates the $15\times2\times5\,\mathrm{cm}$ board for over 16 minutes, completing more than 45 consecutive episodes without reset. We also test $15\times2\times7$ and $15\times2\times10\,\mathrm{cm}$ boards, not included among the eight evaluated.

\paragraph{Human disturbances.}
Uncalibrated pushes perturb pose and balance; drags shift contacts toward object ends (Fig.~\ref{fig:few_shot_zero_shot_tasks}(P--R), movie S15). These can disrupt rapid \passvzero transitions, destabilize contact-sensitive \trianglepassup and \trianglepassdown fingertip grasps, or prevent the next cycle. As an open-loop baseline, we record action sequences from successful executions and replay them on the real robot without feedback-dependent action updates, using the same manual-disturbance protocol. Table~\ref{tab:supp_human_disturbances} compares completion rates at matched task horizons; the sole successful palm-down replay ends in a poor hand--object configuration. Representative sustained demonstrations are reported separately.

\begin{table*}[htbp]
\centering
\small
\setlength{\tabcolsep}{4pt}
\renewcommand{\arraystretch}{1.12}
\caption{
\footnotesize
\textbf{Robustness to human-applied disturbances.} Closed-loop policies and open-loop replay are evaluated under manually applied pushes and drags at the specified episode horizons. Counts denote successful/total trials. Sustained runs are separate closed-loop demonstrations. Disturbance forces and displacements are uncalibrated.}
\label{tab:supp_human_disturbances}
\resizebox{0.8\textwidth}{!}{%
\begin{tabularx}{\textwidth}{@{}p{0.18\textwidth}Xcp{0.13\textwidth}p{0.13\textwidth}p{0.20\textwidth}@{}}
\toprule
Task & Main effect of disturbances & \makecell[c]{Episodes\\per trial} & \makecell[l]{Closed-loop\\completion} & \makecell[l]{Open-loop\\replay} & Representative sustained run \\
\midrule
\trianglepassupns & Object tilt disrupts balance and fingertip support. & 1 & $100\%$ (50/50) & $16\%$ (8/50) & 2 episodes under pushes and drags \\
\trianglepassdownns & Object tilt destabilizes contact-sensitive fingertip grasps. & 1 & $100\%$ (10/10) & $10\%$ (1/10) & 2.5 episodes under three consecutive pushes \\
\passvzerons & Pose and contact shifts disrupt rapid transitions and subsequent rotation cycles. & 2 & $100\%$ (10/10) & $0\%$ (0/10) & 3.5 episodes under three disturbances (pushes/drags) \\
\bottomrule
\end{tabularx}
}
\end{table*}

\subsection*{Ablation Protocols and Analysis}

\subsubsection*{Stage 1: Human-Guided Corrective Optimization}
\label{sec:supp_failure_recovery_ablations}

We compare full first-stage optimization with human takeover, two fixed-pulse variants, and real-world optimization restricted to human-guided kinematic proposals. For each task, Fig.~\ref{fig:human_takeover_failures}(B) reports the maximum progress through a single episode achieved during failure recovery. One operator judges correction by avoidance of the identified failure with normal continuation, not necessarily episode completion. Case counts are 3, 2, 2, and 1 for \passvzerons, \passvonens, \passvtwons, and \passvthreens; 3 for \thumbaroundns, 2 for \sonicdownns, 6 each for \trianglepassupns, \trianglepassdownns, and \trianglepassdownreversedns, and 2 for \fourfingerpassleapns. Each optimized correction has up to 30 iterations; search-space proposal and primitive optimization follow the supplementary methods.

\paragraph{Human-takeover setup.}
\label{para:supp_human_takeover}
A Manus glove drives real-time retargeting through Sharpa's algorithm (\href{https://github.com/sharpa-robotics/sharpa-manus-sdk}{https://\allowbreak github.com/\allowbreak sharpa-robotics/\allowbreak sharpa-manus-sdk}) or \texttt{dex-retargeting}~\cite{qin2023anyteleop} for LEAP. After hardware practice, each operator is allowed up to 240 attempts per failure case and may stop early when they judge further attempts unlikely to yield meaningful improvement.

\noindent\textit{Full takeover.}
A button transfers whole-hand control until completion or failure.
Retargeted poses often disrupt contacts before correction, even in slow \trianglepass and \fourfingerpass grasps. In \trianglepassupns, grasping deliberately can lose two-finger support (Fig.~\ref{fig:human_takeover_failures}(C-i)); teleoperation from the start fares better, reaching approximately $90^\circ$, but occasional transfers to middle and ring fingers fail at thumb recontact before $180^\circ$. In \passns, unsuitable poses prevent grasping (C-v); closing supporting fingers sometimes avoids drops but not subsequent force and contact-transition errors (C-vi). No effective full-takeover recovery occurs for \thumbaround or \sonicdownns.

\noindent\textit{Partial takeover.}
Finger masks leave other fingers policy-controlled. Button-triggered intervention ends automatically after 10 steps for \thumbaround or 20 otherwise; the latter equals approximately $0.66\,\mathrm{s}$ for \passns, $1\,\mathrm{s}$ for \trianglepass and \fourfingerpassns, and $0.2\,\mathrm{s}$ for \sonicdownns. Masks can exclude supporting fingers. At the first \trianglepass failure, index obstruction or insufficient thumb and middle torque prevents completing approximately $90^\circ$. Index-only intervention preserves contacts but cannot reliably reposition within one second (Fig.~\ref{fig:human_takeover_failures}(C-ii)) or repair insufficient torque (C-iii); taking over supporting fingers reintroduces grasp disruption.

\paragraph{Fixed-pulse correction variants.}
The kinematic fixed-pulse variant applies a triangular pulse to each joint selected through human-guided kinematic correction:
\begin{equation}
    \delta a_j(t)=\sigma_j\,\frac{\pi}{18}\,
    \max\!\left(1-\frac{|t-t_{\mathrm{fail}}|}{10},0\right),
    \label{eq:supp_ablation_triangular_pulse}
\end{equation}
where $t$ is the frame index, $t_{\mathrm{fail}}$ is the annotated failure frame and $\sigma_j\in\{-1,+1\}$ is the proposed direction. Selected joints receive a $10^\circ$ peak with 10-frame rise and fall; others remain unchanged. The coarse-physics variant additionally uses candidates from coarse-physics-guided contact correction, with the same pulses. Neither optimizes parameters. Each candidate is tested twice against initialization variability.

Kinematic pulses resolve some slow-rotation obstructions or missing contacts; contact proposals modestly improve force-sensitive correction. Fixed timing and magnitude nevertheless remain inadequate and can disrupt policy continuation.

\paragraph{Kinematic proposals with real-world optimization.}
The \textit{Human-Guided Kinematic Correction + Real-World Optimization} variant retains only the joints and motion directions proposed by human-guided kinematic editing. It uses the full method's Gaussian primitives, parameter bounds, and real-world CMA-ES procedure, excluding coarse-contact-model-generated orientation and position candidates. Maximum single-episode progress matches the kinematic fixed-pulse variant on all ten tasks (Fig.~\ref{fig:human_takeover_failures}(B)). This suggests that searching the kinematic proposal space alone can leave critical contact-force corrections uncovered. Together, the ablations support combining contact-guided proposals with primitive optimization. Relative to fixed pulses, optimized primitives change temporal parameterization and parameter fitting together.

\subsubsection*{Stage 2: Model-Based Refinement}
\label{sec:supp_model_based_ablations}

\paragraph{Hybrid-component ablations.}
Contact-only and action-only compensation each undergo two iterations, with 200 real-world trajectories and 1,000 PPO epochs per iteration. 
Action-only compensation additionally uses 100,000 trajectories from the original simulator per iteration, collected with the initial policy in the first iteration and the updated policy in the second.
Evaluation follows Table~\ref{tab:supp_evaluation_protocols}, with combined compensation achieving the best performance (Fig.~\ref{fig:model_based_abl}(A)).

\paragraph{Real-world trajectory-tracking evaluation.}
Figure~\ref{fig:model_based_abl}(B)
evaluates the four \pass variants: \passvzerons, \passvonens, \passvtwons, and \passvthreens. Each source model is trained on 200 real-world trajectories from one variant and evaluated on all four variants at compensation scale $\beta_c=10.0$, giving four same-variant and twelve cross-variant source--target pairs. For each test condition, we collect 20 additional trajectories exclusively for held-out evaluation. We replay their recorded actions in MuJoCo Warp, measuring object-orientation error, position error, and tracking survival length.

For test trajectory $i$ at time $t$, let $\Delta q_i(t)=\hat q_i(t)\otimes q_i(t)^*$ be the relative unit quaternion between the predicted and reference object orientations. The angular error in radians is
\begin{equation}
    e_i(t)=2\arcsin\!\left(\min\!\left\{1,\left\|[\Delta q_i(t)]_{xyz}\right\|_2\right\}\right).
\end{equation}
We average over the $N_{\mathrm{test}}=20$ held-out trajectories and the $T$ evaluated time steps:
\begin{equation}
    E=\frac{1}{T}\sum_{t=0}^{T-1}
    \left[\frac{1}{N_{\mathrm{test}}}\sum_{i=1}^{N_{\mathrm{test}}}e_i(t)\right].
\end{equation}

Position error replaces $e_i(t)$ with $\|\hat{\mathbf p}_i(t)-\mathbf p_i(t)\|_2$ in the same average. Both errors are normalized as $E_{\mathrm{model}}/E_{\mathrm{baseline}}$ under the same test condition. 
Tracking survival length is reported relative to each trajectory's length, because rollout lengths differ across tasks. For test trajectory $i$ with $T_i$ recorded steps, let $L_i\le T_i$ be the number of steps replayed before
the simulated replay fails (the simulated object falls below
$z=0.45\,\mathrm{m}$ or the simulation state becomes nonfinite), with
$L_i=T_i$ when the replay reaches the end of the trajectory. We report
\begin{equation}
    S=\frac{100}{N_{\mathrm{test}}}\sum_{i=1}^{N_{\mathrm{test}}}\frac{L_i}{T_i},
\end{equation}
a percentage in $[0,100]$; higher values indicate longer tracking.
These evaluations measure trajectory prediction, not control-policy transfer across tasks.

Orientation and position errors are computed only before replay failure, or over the full trajectory if no failure occurs. We report these errors alongside tracking survival length to characterize both tracking accuracy and duration.

\paragraph{Tracking after policy updates.}
On \passvzero and \trianglepassupns, we refine the policy in the contact-compensated simulator while keeping the learned compensator fixed. We then replay held-out real-world actions from the original and updated policies with compensation enabled or disabled (Supplementary Section~S3).
The model retains lower orientation and position errors and longer tracking survival after the policy update, supporting generalization beyond the data-collection policy.

\subsubsection*{Stage 3: Real-World Reinforcement Learning}
\label{sec:supp_model_free_ablations}

Figure~\ref{fig:structural_and_residual_exploration} separates residual representation (A,B), temporal coordination (C), and execution through a learned action interface (D,E). Panel A covers all ten tasks; panels B--E use \passvzerons, \passvonens, \sonicdownns, and \trianglepassupns. Unless full-action execution is explicitly tested, residuals augment fixed base-policy commands recomputed at every control step.

Supplementary Section~S3 provides additional real-world sensitivity analyses on the same four tasks. These experiments characterize sensitivity to chunk length, latent dimension, and residual scales; they are not used for hyperparameter selection on hardware. Task-specific chunk lengths are selected in simulation, while $d_z=16$ and $\alpha_r=\beta_r=0.25$ are empirical defaults.

\paragraph{Joint-space residual comparison.} Both methods use the same fixed second-stage controller, SAC algorithm, task reward, large-scale simulation pretraining setting, real-world rollout budget, and checkpoint-selection rule. The joint-space residual policy is evaluated every control step; its output is bounded to $[-1,1]$, scaled by 0.22, and added to the current base command in the controller's action coordinates. Exploration noise is calibrated separately from this bound in simulation to retain at least 90\% of the noise-free base controller's mean return before large-scale pretraining. It then receives the additional two-environment, 1,000-epoch simulation search described in Sec.~\nameref{sec:supp_baselines}, including target entropy, noise-onset epoch, critic LayerNorm, actor-update frequency, and the base-distance penalty. Figure~\ref{fig:structural_and_residual_exploration}(A) reports real-world learning with these simulation-selected settings. The latent-residual method retains $\alpha_r=0.25$ and $\beta_r=0.25$, with chunked modulation and per-step residual execution.

This comparator shares the per-step additive residual formulation of ResFiT~\cite{Ankile2025ResidualOR}, with the base policy frozen. We use proprioceptive SAC and a base controller recomputed at every step, differing from ResFiT's TD3-style updates and the action-chunked BC bases used in its experiments. The additional base-distance penalty is our conservative modification.

\paragraph{Linear and temporally coordinated alternatives.}
In panel B, PCA is fitted to the same standardized action chunks used to train the nonlinear representation, with matching chunk length and latent dimension. Its principal directions linearly generate residual sequences, with inverse-standardization scales absorbed into the basis. The temporal-only alternative instead repeats one joint-residual vector throughout each chunk. Both introduce temporal dependence: chunk-constant residuals provide persistence alone, whereas PCA captures data-derived correlations across joints and time. Initial mean absolute per-joint residual magnitudes are matched over the first 50 control steps. Neither construction guarantees smooth residuals or continuity at chunk boundaries. The nonlinear representation outperforms both alternatives on the four tested tasks.
The panel B variants share the same base controller, reward, rollout budget, and initial residual magnitude, and retain per-step base-policy feedback.  

\paragraph{Temporally coordinated residual actions.}
Panel C retains the latent-residual formulation and per-step base-policy feedback, comparing temporally structured residual sequences with per-step residual generation. The latter removes explicit coordination across future control steps while retaining a learned action representation. Temporally structured residuals achieve higher final reward and episodes on all four tasks.

\paragraph{Base-policy execution and learned action interfaces.}
Panel D compares five configurations before real-world RL. Four BC alternatives predict action chunks: direct joint-command output or output through a learned action interface, each executed with a replanning interval of either one or the full chunk. With an interval of one, only the first action is executed before replanning.
The fifth configuration is our unchunked Stage 2 policy, which directly computes joint commands from the latest observation. All configurations use 20, 50, 50, and five evaluation trials on \passvzerons, \passvonens, \sonicdownns, and \trianglepassupns, respectively; bars report mean episodes and population SD. Per-step replanning improves both chunked variants, and direct joint-command output outperforms the learned interface at either interval.

\paragraph{Residual versus full-action exploration.}
Panel E compares chunk-wise residual exploration, chunk-wise full-action exploration, and per-step full-action exploration using the same learned latent space during real-world RL. The chunk-wise full-action variant executes an entire decoded chunk before replanning. The per-step variant retains the same representation and prediction horizon, but recomputes the latent and decoded chunk from the latest observation at every control step and executes only the first action. Residual exploration retains the decoded correction sequence while adding each correction to a base command recomputed every step. Per-step full-action execution restores frequent feedback but still routes complete commands through the representation and no longer explicitly preserves within-chunk temporal coordination. The comparison therefore does not isolate reconstruction error as the sole cause of its remaining gap.

\paragraph{Conservative tuning of full-action exploration.}
Both full-action variants in Fig.~\ref{fig:structural_and_residual_exploration}(E) undergo conservative tuning entirely in simulation. For each variant, we first calibrate the pre-decoder coefficient $\beta_r$ in $\mathbf z_t^{\mathrm{mod}}=\mathbf z_t+\beta_r\Delta\mathbf z_t$, selecting a scale that retains at least 90\% of the mean return obtained by decoding $\mathbf z_t$ alone under the same execution mode. This criterion concerns mean return, not success rate. We then perform large-scale simulation pretraining. Each additional search run starts from the pretrained actor and critic checkpoints and varies the initial standard deviation of latent-policy actions, the probability of selecting the SAC action, the learning rate, and the number of gradient updates. To limit exploration and policy changes, we also vary whether the critic uses LayerNorm, actor-update frequency, target entropy, and the onset of $\Delta\mathbf z_t$ at epoch $\{0,50,100,150,200\}$ of the additional simulation training. The simulation-selected settings are used for subsequent real-world tuning.

\paragraph{Additional chunk-length comparisons and discount matching.}
Supplementary Section~S3 provides a sensitivity test of 
real-world performance to the action chunk length by varying $T_c\in\{2,8,16\}$ (independent of the \{4, 8, 16\} candidates used for simulation-based selection)
with $d_z=16$ fixed, thereby changing the temporal extent and action-chunk compression ratio while preserving the latent exploration dimension.
Both simulation-based selection and real-world optimization use $\gamma_{\mathrm{step}}=0.99^{1/8}$ per control step:
\[
\gamma_{T_c}=\gamma_{\mathrm{step}}^{T_c}=0.99^{T_c/8}.
\]
Shortened chunks that retain bootstrapping use their actual executed duration $n_t$ in the exponent; within-chunk rewards remain undiscounted sums. Longer chunks ($T_c=16$) are less effective on \passvone and \sonicdown than on \passvzero and \trianglepassupns. Sequence variability, representation difficulty, and the coupled configuration changes may contribute.

\paragraph{Latent scale.}
Before decoding, $\mathbf z_t^{\mathrm{mod}}=\mathbf z_t+\beta_r\Delta\mathbf z_t$, with unscaled policy output $\Delta\mathbf z_t$ (Eq.~\eqref{eq:supp_latent_modulation}); $\beta_r$ differs from post-decoding action scale $\alpha_r$. 
In the sensitivity test on the scale (Supplementary Section~S3), 
$0.125$ works well but peaks below $0.25$, whereas $0.5$ and $0.75$ sometimes reach competitive performance with stronger fluctuations. This suggests that small offsets restrict improvement while large offsets complicate optimization.

\paragraph{Closed-loop responsiveness in simulation.}
\label{para:supp_closedloop}
Simulation BC bases use $H\in\{1,2,4,8\}$ with residual chunk $T_c=8$ on \passvzero in a separate simulation control.
For $H>1$, executing only the first action before replanning largely restores in-domain performance lost under full-chunk execution. Remaining long-chunk losses may reflect sequence-fitting difficulty.

We then train residuals with actuator P-gains scaled by $0.8$: four configurations, five seeds each (20 runs), eight RTX 4090 GPUs with $8{,}192$ environments each. Reward--environment-step curves remain impaired for less responsive bases despite massive parallel experience, indicating that sampling does not compensate for delayed feedback here.

\paragraph{Additional single-step full-action control.}
In a separate comparison with $T_c=1$ and $d_z=22$ (the hand-action dimension), full-action exploration barely improves \passvzero and \sonicdownns,
plateaus after early \trianglepassup gains, and deteriorates on \passvonens, whereas residual exploration improves substantially. These results are consistent with benefits from retaining base-policy commands even without temporal or dimensional compression, but do not isolate reconstruction error.

\paragraph{Starting-policy performance.}
Related RL methods often start from substantial success: RL-100's DP3 obtains 67.8\%, then 91.8\% after offline RL~\cite{rl100} (their Table~1); ZPRL starts at approximately 70\%, 60\%, 82.5\%, and 20\%~\cite{Yu2026BeyondAR} (their Fig.~8); DICE-RL starts at 46.67\%, 56.67\%, and 56.67\% on Gear Insertion, Light Bulb Insertion, and Belt Assembly~\cite{Sun2026FromPT} (their Fig.~8). ZPRL's Insert Bills improves from 20\% to 77.5\%, still below our reliability target. These examples establish no universal threshold. Only two of our tasks reach 30\%--70\% after correction; intermediate model-based refinement improves full-episode success and reduces the need to discover successful behavior through costly physical exploration.

\subsection*{Visualization and Supplementary Movies}

The interactive viewer in the corrective optimization stage is developed with 
Viser~\cite{yi2026viser}. Reliability and sustained-rotation movies are continuous, uncut recordings with accelerated playback and annotations; reliability clips also use slow motion to explain terminal-grasp/reset cases.

\paragraph{Related-work speed estimates.}
From public videos, we select stable finger-gaiting sequences spanning approximately $\pi$ radians. For timeline duration $T_{\mathrm{video}}$ and playback factor $s$ ($s=2$: $2\times$ playback), physical duration is $sT_{\mathrm{video}}$ and
$
\hat{\omega}
=\frac{\pi}{sT_{\mathrm{video}}}.
$
For approximately planar longitudinal-axis rotation, this approximates our pen-axis speed metric. Estimates characterize selected sequences:

\noindent\textit{PenSpin}~\cite{Wang2025PenSpin}: teaser/failure videos (\href{https://penspin.github.io/}{https://penspin.github.io/}) yield $0.214\pi$--$0.234\pi\,\mathrm{rad/s}$ assuming real-time playback. Visible human motion in failures supports this assumption qualitatively; teaser speed is unverified.
\textit{DexNDM}~\cite{Liu2025DexNDMCT}: cuboid demonstrations (\href{https://meowuu7.github.io/DexNDM/}{https://meowuu7.github.io/DexNDM/}) yield $0.132\pi$--$0.150\pi\,\mathrm{rad/s}$ after the explicitly labeled $2\times$ correction.
\textit{TeleDexter}~\cite{Li2026TeleDexter}: mid-spin recovery (\href{https://bigai-dex.github.io/blog/teledexter/}{https://bigai-dex.github.io/blog/teledexter/}) yields approximately $0.5\pi\,\mathrm{rad/s}$ assuming real time, excluding initial stationary adjustment and intermediate pauses; this is active-rotation speed, not whole-sequence average.
\textit{TopoRetarget}~\cite{Wu2026TopoRetarget}: first and third \texttt{hero.mp4} clips on its website, 
selected for visible contact transitions, yield $0.111\pi$--$0.254\pi\,\mathrm{rad/s}$ assuming real time.

Playback speed, visually estimated angles, and temporal boundaries introduce uncertainty; these are approximate motion comparisons, not standardized system evaluations.

Simulation-only studies~\cite{Charlesworth2020SolvingCD,Ma2023EurekaHR} are excluded.

%% file: supp/supp_additional_res_2.tex
\section{Additional Experimental Results}
\label{sec:supp_additional_results}

Protocols and metrics for the following failure cases, dynamics analyses, transfer tests, and rollout variations are defined in Sec.~\ref{sec:supp_exp_settings}.

\subsection*{Direct Transfer from Isaac Lab}
\label{sec:supp_alternative_simulator_transfer}
We trained policies in Isaac Lab and evaluated direct transfer on \passvzero and \fourfingerpassleapns, with 50 real-world trials per task. Neither policy completed a task episode (0/50 successes on each task). Execution was also qualitatively worse than with policies trained in the original MuJoCo Warp setup. \textbf{Changing the training simulator alone did not resolve the transfer difficulties in these two-task tests.}

\subsection*{Task-Specific Corrective Optimization}
\label{sec:supp_recovery_cases}

Below, \texttt{mcp}, \texttt{mcp\_aa}, and \texttt{pip} denote MCP flexion--extension, MCP abduction--adduction, and PIP joints, with signs following robot coordinates. Procedures follow Sec.~\nameref{sec:failure_recovery_supp}.

\paragraph{\passvzerons: three events.}
During the first ring--little to middle--ring transfer, premature middle-finger lowering obstructs rotation. While the ring and little fingers retain the object, the annotator repositions the middle finger, accepts the progressive-IK positive \texttt{mcp} correction, and targets rotation about the object's negative local $y$ axis. Contact analysis proposes strengthening either or both second-phalangeal contacts using minimal feasible \texttt{mcp} and \texttt{mcp\_aa} subsets. Because the hand edit removes an obstruction, we can optimize the kinematic candidate alone or jointly with an orientation correction candidate; CMA-ES succeeds with the kinematic candidate alone. Near the episode's end, index--middle contact loss causes a drop. Immediately after separation, both near-contact candidates remain; without a hand edit, the annotator targets rotation about the object's positive local $y$ axis. Of three proposals (strengthen index, middle, or both), only the highest-ranked dual-contact proposal is optimized, producing retention and driving torque. The first obstruction recurs in episode two and is resolved by reusing its primitive.

\paragraph{\passvonens: two events.}
Halfway through the episode, the middle--ring transfer stalls. The middle finger's small \texttt{mcp} angle prevents further lowering, so the annotator shifts the object along positive $y$ and $z$ toward a central grasp without editing the hand, removes the obstructing contact, and specifies the rotation direction. Position correction anchors seven recorded frames earlier, where relative contacts best match the target configuration; tangential corrections oppose downward slip. Orientation correction acts at the failure frame. Both propose strengthening ring, little, or both contacts through \texttt{mcp} and \texttt{mcp\_aa}. Joint optimization yields an earlier dual-finger anti-slip correction and subsequent rotational correction. A similar middle--ring to index--middle failure resists primitive reuse; separate optimization selects index-, middle-, and ring-finger joints.

\paragraph{\passvtwons: two events.}
Both resemble \passvonens; the same procedure selects largely the same key joints.

\paragraph{\passvthreens: one event.}
Halfway through the episode, the partially extended middle finger obstructs the ring--little to middle--ring transfer. Editing proposes positive \texttt{mcp} and \texttt{pip} corrections; contact analysis proposes strengthening either or both ring and little second-phalangeal contacts. Kinematic-only optimization fails. The effective primitive transiently suppresses obstructing middle-finger motion while strengthening both contacts to drive rotation.

\paragraph{\thumbaroundns: three events.}
For ineffective launch, the annotator specifies an orientation at launch without hand edits. Contact proposals strengthen thumb, middle, both, or thumb while softening little; CMA-ES selects thumb and middle. When the launched object travels away, annotation again targets launch; orientation-guided optimization, together with the launch correction, coordinates thumb, index, and middle fingers to return it. Finally, the returning object rebounds during catching. At index contact, the annotator edits the finger to lift and cup it. With no object-pose target, contact proposals are omitted; kinematic-space optimization yields a catching primitive.

\paragraph{\sonicdownns: two events.}
At the first slot transition, the object becomes trapped between index and middle fingers. Editing middle and ring fingers and proposing stronger index, middle, or both contacts yields optimized index-, middle-, and ring-finger corrections. A similar second-transition failure resists reuse and requires separate annotation and optimization.

\paragraph{\trianglepassupns{} and \trianglepassdownns: six events each.}
Each half-turn contains two drops from two-finger grasps near $90^\circ$ and $180^\circ$, plus a returning thumb that pushes the object away near the half-turn boundary. Drop corrections transfer to the second half; thumb corrections require separate optimization.

\paragraph{\trianglepassdownreversedns: six events.}
In each half-turn, the thumb obstructs rotation near the transition with insufficient driving torque, then recontacts at an unsuitable location; the index next returns to the wrong position on the opposite side. The first correction requires reoptimization in the second half; the others follow the same procedure.

\paragraph{\fourfingerpassleapns: two events.}
Near $90^\circ$, trapping between thumb and ring stalls rotation. Its correction also resolves the analogous second-half event.

\paragraph{Correction-policy training data.}
Table~\ref{tab:supp_s1_correction_data} summarizes the real-world rollouts collected after primitive optimization to train the stage 1 correction policy (Sec.~\ref{sec:corrective_policy_details}).

\begin{table}[htbp]
\centering
\small
\caption{\footnotesize
\textbf{Stage 1 Correction-Policy Training Data.}
Retained successful real-world rollouts and environment steps per rollout for each task, collected within the 3-hour cap per task. Counts exclude failed rollouts from the same collection, primitive-search trials, and subsequent-stage rollouts.
}
\label{tab:supp_s1_correction_data}
\resizebox{0.8\textwidth}{!}{%
\begin{tabular}{lrr}
\toprule
Task & Rollouts & Env. steps per rollout \\
\midrule
\thumbaroundns & 421 & 20 \\
\sonicdownns & 280 & 400 \\
\passvzerons & 1{,}000 & 100 \\
\passvonens & 270 & 78 \\
\passvtwons & 200 & 78 \\
\passvthreens & 350 & 60 \\
\trianglepassupns & 150 & 400 \\
\trianglepassdownns & 100 & 400 \\
\trianglepassdownreversedns & 60 & 400 \\
\fourfingerpassleapns & 30 & 400 \\
\midrule
Total & 2{,}861 & -- \\
\bottomrule
\end{tabular}
}
\end{table}

\subsection*{Two-Operator Evaluation of Failure Recovery}
\label{sec:supp_two_operator_recovery}

\paragraph{Protocol.}
Two operators independently address two selected failure cases on each of \passvtwons, \thumbaroundns, and \trianglepassupns. Both receive standardized interface tutorials, scoring instructions, and author demonstrations, using practice cases separate from formal evaluation. Teleoperation familiarization and task-specific takeover practice are capped at 30 minutes. During formal experiments, operators independently select failure frames, target poses, and kinematic edits without viewing each other's annotations or resulting corrections. For later failures, shared, previously optimized primitives handle preceding failures, so intervention or optimization targets only the selected failure.
The two operators were members of the authors' laboratories who were not involved in developing the method and took part voluntarily; the evaluation concerns the recovery methods, and only task outcomes, trial counts, and timing were recorded.

\paragraph{Budgets and timing.}
All methods have a 240-trial cap per operator and failure case. Operators may stop early when they judge that further attempts are unlikely to produce effective recovery, resulting in different takeover trial counts. Human time depends on trial duration, task, and operator pacing between attempts; our method additionally involves guidance and deliberation when scoring outcomes. It therefore need not scale proportionally with the number of trials.

\paragraph{Results.}
Neither full-hand nor partial takeover achieves recovery in any evaluated case for either operator (Table~\ref{tab:supp_two_operator_recovery}). Corrective optimization achieves nonzero recovery in all six cases for both operators, with mean recovery rates of 64.2\% and 55.8\%, respectively. These results show recovery benefits across both operators, with remaining variation across operators and failure cases.

\begin{table*}[htbp]
\centering
\small
\setlength{\tabcolsep}{5pt}
\renewcommand{\arraystretch}{1.12}
\caption{\footnotesize
\textbf{Two-Operator Evaluation of Failure Recovery.}
C1 and C2 denote two selected failure cases per task, shared by both operators.
Recovery reports successful/total trials over 20 fresh evaluations, using fixed optimized corrections for our method and live interventions for takeover. Success requires avoiding the targeted failure and continuing normally through a predefined recovery window.
Search trials count case-specific attempts before final evaluation. For our method, these are real-world CMA-ES evaluations (eight per generation); the search stops at the first proposal that recovers the failure, which can occur mid-generation, so counts need not be multiples of eight. Human time includes case-specific guidance, attempts, scoring, and resets, excluding final evaluation and hardware cooldowns.
}
\label{tab:supp_two_operator_recovery}
\resizebox{\textwidth}{!}{%
\begin{tabular}{lll|ccc|ccc}
\toprule
\multirow{2}{*}{Task}
& \multirow{2}{*}{Case}
& \multirow{2}{*}{Method}
& \multicolumn{3}{c|}{Operator 1}
& \multicolumn{3}{c}{Operator 2} \\
\cmidrule(lr){4-6}\cmidrule(lr){7-9}
& & & Recovery & Search trials & Human time (min)
    & Recovery & Search trials & Human time (min) \\
\midrule

\multirow{6}{*}{\passvtwons}
& \multirow{3}{*}{C1}
& Full takeover      & 0/20 & 121 & 25 & 0/20 & 78 & 15 \\
& & Partial takeover & 0/20 & 47 & 14 & 0/20 & 29 & 10 \\
& & Ours             & 12/20 & 31 & 18 & 14/20 & 36 & 22 \\
\cmidrule(lr){2-9}
& \multirow{3}{*}{C2}
& Full takeover      & 0/20 & 62 & 17 & 0/20 & 40 & 11 \\
& & Partial takeover & 0/20 & 23 & 6 & 0/20 & 31 & 9 \\
& & Ours             & 5/20 & 19 & 15 & 6/20 & 27 & 19 \\
\midrule

\multirow{6}{*}{\thumbaroundns}
& \multirow{3}{*}{C1}
& Full takeover      & 0/20 & 11 & 2 & 0/20 & 9 & 2 \\
& & Partial takeover & 0/20 & 13 & 3 & 0/20 & 15 & 5 \\
& & Ours             & 15/20 & 40 & 15 & 10/20 & 31 & 14 \\
\cmidrule(lr){2-9}
& \multirow{3}{*}{C2}
& Full takeover      & 0/20 & 8 & 2 & 0/20 & 5 & 2 \\
& & Partial takeover & 0/20 & 9 & 2 & 0/20 & 11 & 4 \\
& & Ours             & 7/20 & 47 & 18 & 9/20 & 35 & 15 \\
\midrule

\multirow{6}{*}{\trianglepassupns}
& \multirow{3}{*}{C1}
& Full takeover      & 0/20 & 29 & 13 & 0/20 & 32 & 15 \\
& & Partial takeover & 0/20 & 16 & 8 & 0/20 & 12 & 5 \\
& & Ours             & 18/20 & 15 & 17 & 15/20 & 27 & 23 \\
\cmidrule(lr){2-9}
& \multirow{3}{*}{C2}
& Full takeover      & 0/20 & 20 & 10 & 0/20 & 24 & 12 \\
& & Partial takeover & 0/20 & 89 & 35 & 0/20 & 52 & 20 \\
& & Ours             & 20/20 & 11 & 7 & 13/20 & 18 & 14 \\
\bottomrule
\end{tabular}%
}
\end{table*}

\subsection*{Model-Based Refinement: Additional Analysis}
\label{sec:supp_additional_model_analysis}

\subsubsection*{Fully Neural and NeRD-Style Hand--Object Dynamics}
\namedlabel{para:pure_neural_wm}{Whole-system Neural WM}

We tested architectures and teacher-forced and autoregressive training for fully neural hand--object dynamics. The best configuration followed~\cite{Li2025RoboticWM}: eight-frame recurrent histories, 16-step autoregressive supervision, and several million simulated trajectories. State comprises joint positions and a six-dimensional unsigned object axis, with commanded targets in the history; it does not represent the full physical state.

Held-out action replay reproduced complete rotation cycles relatively accurately. Closed-loop evaluation, feeding each predicted state to the policy at every step, instead produced brief plausible motion followed by a nearly stationary object axis. Policy optimization at a very small learning rate yielded slight initial gains before stagnating or degrading. These observations concern the tested models and partial-state representation. The joint-wise delta-action surrogate retains their recurrent history, 16-step objective, and unsigned-axis representation.

We also tested a NeRD-style hybrid~\cite{Xu2025NeuralRD}, trained on a comparable amount of simulated data (several million trajectories). The observations were the same: held-out action replay was relatively accurate, whereas closed-loop evaluation and policy optimization degraded as described above.

\subsubsection*{Collision-Geometry Approximation}

\begin{figure}[htbp]
  \centering
  \includegraphics[width=\textwidth]{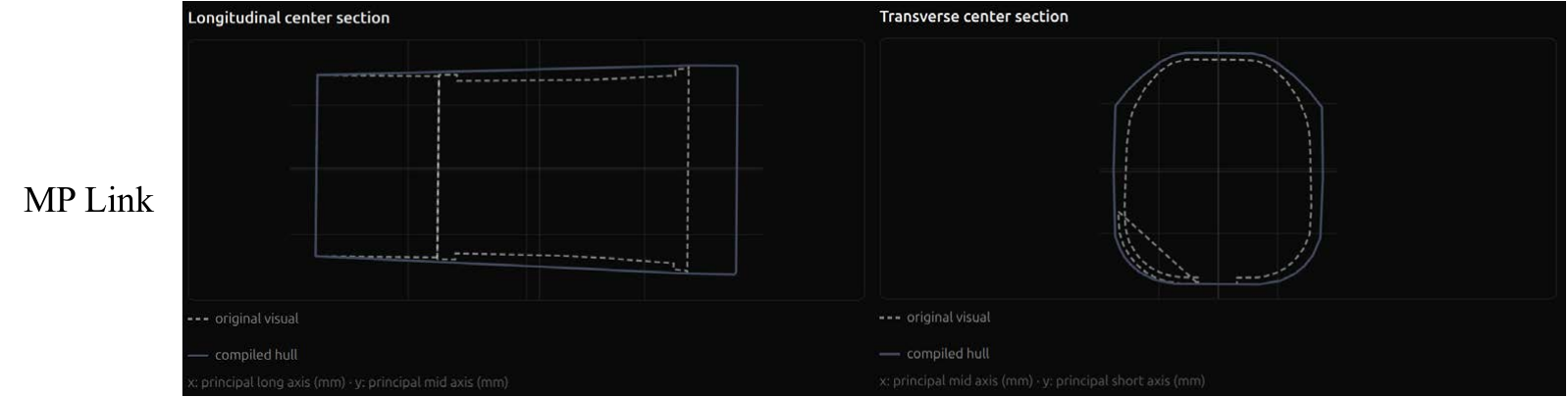}
  \caption{\footnotesize
  \textbf{Collision-Geometry Approximation of the Sharpa Hand: MP Links.} Comparison of original mesh geometry and convex-decomposed collision geometry (MP links).
  }
  \label{fig:collision_geometry_convexify_viz_mp}
\end{figure}

\begin{figure}[htbp]
  \centering
  \includegraphics[width=\textwidth]{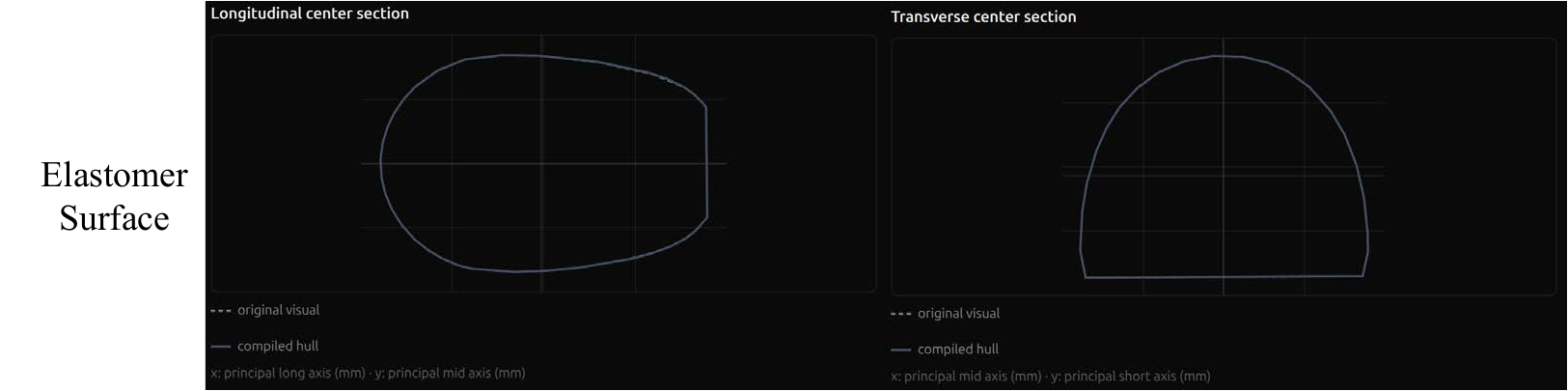}
  \caption{\footnotesize
  \textbf{Collision-Geometry Approximation of the Sharpa Hand: Fingertip Elastomer.} Comparison of original mesh geometry and convex-decomposed collision geometry (fingertip elastomer surfaces).
  }
  \label{fig:collision_geometry_convexify_viz_elastomer}
\end{figure}

Convex decomposition alters collision surfaces and detected contacts (Figs.~\ref{fig:collision_geometry_convexify_viz_mp} and~\ref{fig:collision_geometry_convexify_viz_elastomer}); our compensator changes contact constraints, not geometry.

Fingertip elastomers have 97.4\% solid IoU and $0.67\,\mathrm{mm}$ maximum bidirectional surface distance. MP links dominate decomposition error: 90.1\% IoU, $7.66\,\mathrm{mm}$ maximum distance, and 95\% of sampled distances below $3.07\,\mathrm{mm}$. LEAP's planar inner and lateral fingertips use boxes; capsule and cuboid objects use parametric primitives, avoiding decomposition.

Contact positions in hand-body coordinates and geometry identifiers (Table~\ref{tab:hybrid_actor_obs}) permit region-dependent constraint corrections; collision geometry and detected contact positions remain unchanged.

\subsubsection*{Analysis of the Contact Compensator}
\label{sec:supp_compensation_consistency}

\begin{figure}[htbp]
  \centering
  \includegraphics[width=0.85\textwidth]{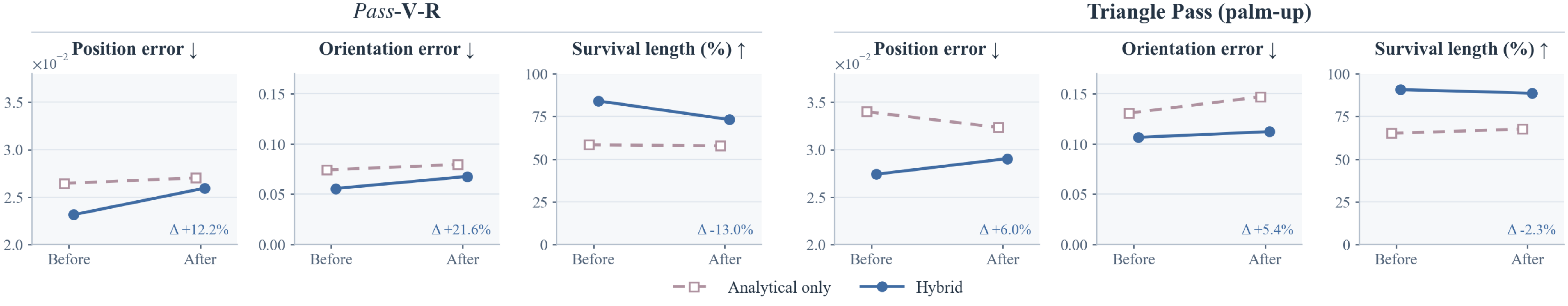}
  \caption{\footnotesize
  \textbf{Tracking Generalization of the Hybrid Simulator.} 
  Tracking of held-out real-world trajectories from the original and refined \passvzero and \trianglepassup policies, comparing the calibrated analytical simulator with a hybrid simulator whose contact compensator remains fixed. Position error is shown in units of $10^{-2}\,\mathrm{m}$ and orientation error in radians; $\Delta$ gives the hybrid simulator's relative change after the policy update.
  }
  \label{fig:supp_stage2_additional_abl}
\end{figure}
\begin{figure}[htbp]
  \centering
  \includegraphics[width=0.85\textwidth]{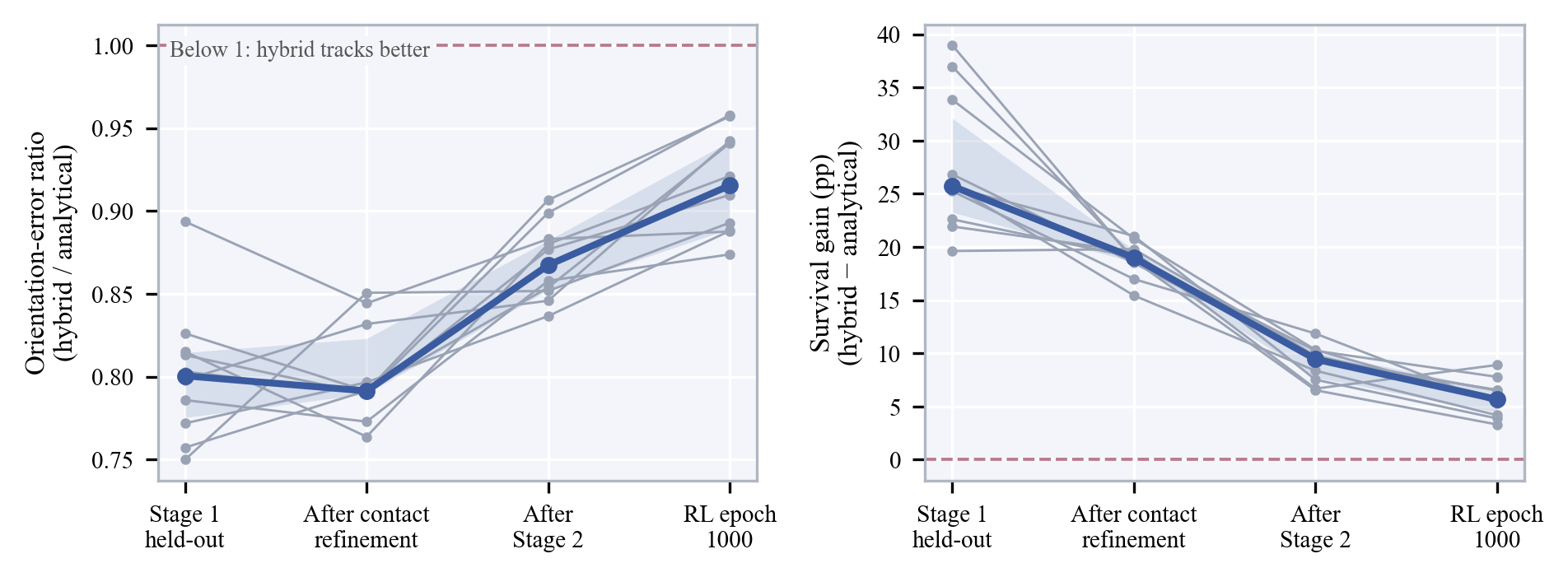}
  \caption{\footnotesize
  \textbf{Tracking across Policy Updates with a Fixed Contact Compensator.} Held-out real-world trajectories from successive policies are replayed with the compensator trained once on Stage-1 data. Left: ratio of hybrid to analytical orientation error ($<1$ favors the hybrid simulator). Right: survival gain of the hybrid simulator in percentage points. Thin lines denote tasks; thick lines and bands denote the median and interquartile range across the ten tasks.
  }
  \label{fig:supp_compensation_policy_updates}
\end{figure}

A compensator trained on one \pass variant reduces orientation error to 0.75–0.97 and position error to 0.45–0.99 of the calibrated analytical simulator’s values on held-out trajectories of every variant, with equal or longer tracking survival in all 16 source–target pairs (Fig.~\ref{fig:model_based_abl}(B)). 
With the compensator fixed after Stage-1 training, the hybrid simulator also tracks trajectories from the refined Pass-V-R and Triangle Pass (palm-up) policies better than the analytical simulator, although its errors rise by 5–22\% and its survival falls by 2–13\% after the update (Fig.~\ref{fig:supp_stage2_additional_abl}).
On all ten tasks, the hybrid simulator likewise tracks trajectories from all later policies better than the analytical simulator, but the advantage narrows as the policy changes (median orientation-error ratio from 0.80 to 0.92 and survival gain from 26 to 6 percentage points; Fig.~\ref{fig:supp_compensation_policy_updates}), consistent with the diminishing returns of model-based refinement (Fig.~\ref{fig:framework_stages}(B)).

\subsection*{Model-free RL Tuning: Additional Ablations}

\begin{figure}[!htbp]
  \centering
  \includegraphics[width=\textwidth]{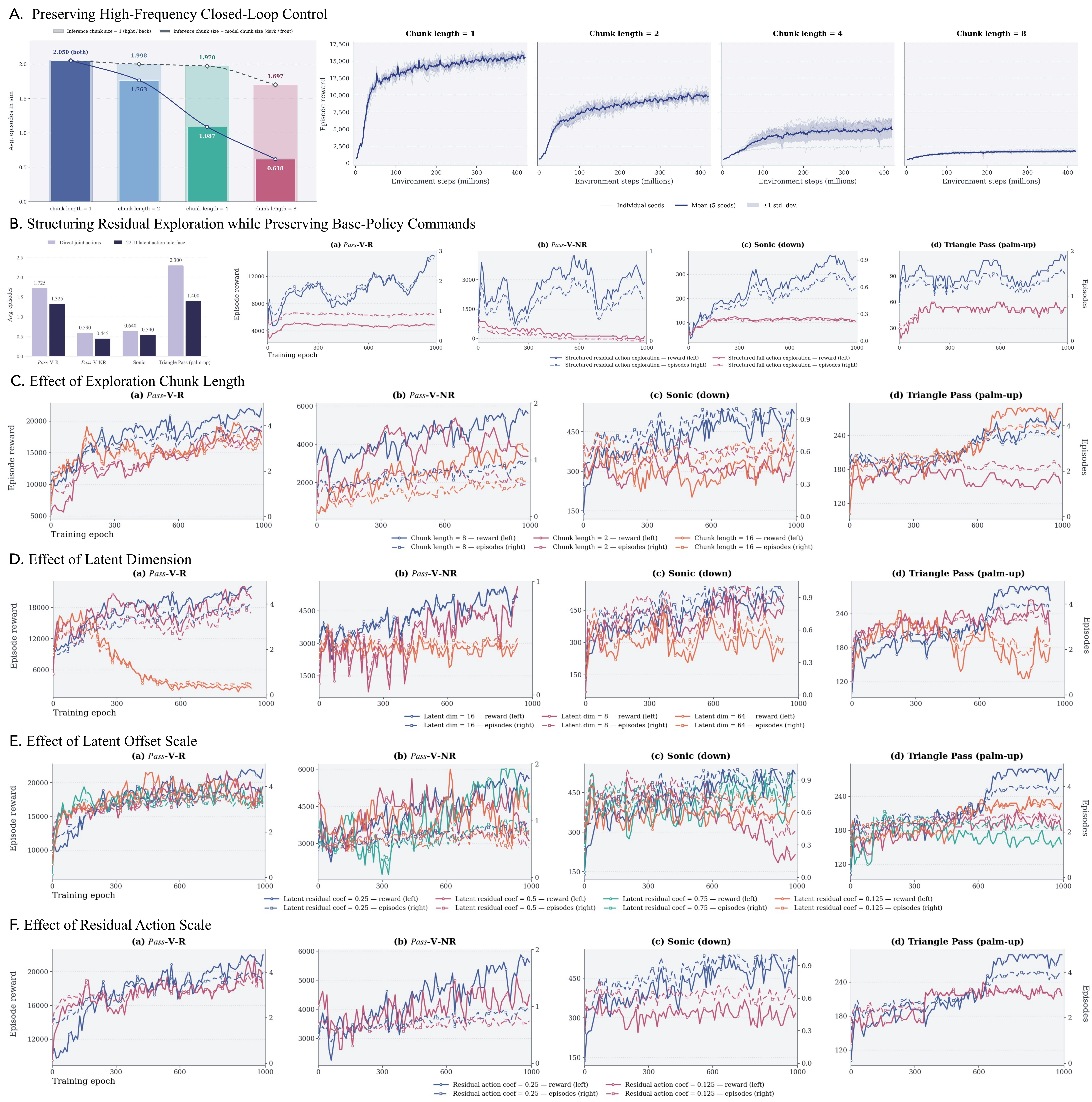}
\caption{\footnotesize
    \textbf{Additional Ablations of Structured Residual RL.}
    \textbf{(A)} Simulated \passvzerons: BC policies with different prediction horizons evaluated with per-step replanning or full-chunk execution (left), followed by latent residual RL over fixed BC bases under reduced actuator P-gains (right; mean $\pm$ SD across five seeds).
    \textbf{(B)} Single-step, 22-dimensional comparisons: direct joint-action versus latent-interface BC (left), and residual versus full decoded-action execution during real-world RL (right).
    \textbf{(C--F)} Additional real-world parameter sensitivity analyses: exploration chunk length with $d_z=16$ fixed (C), latent dimension with task-specific chunk lengths fixed (D), latent offset scale $\beta_r$ (E), and residual action scale $\alpha_r$ (F).
    These comparisons assess sensitivity, rather than select hyperparameters on hardware. 
    Learning curves in (B--F) show training-rollout reward (solid) and episodes (dashed).
}
  \label{fig:ablations_stage3}
\end{figure}

We further examine base-policy feedback, execution through a learned action interface, and the temporal extent and magnitude of residual exploration (Fig.~\ref{fig:ablations_stage3}). Panels C--F provide additional real-world sensitivity analyses, not hyperparameter selection on hardware. They assess variation around simulation-selected chunk lengths and the empirical defaults $d_z=16$ and $\alpha_r=\beta_r=0.25$. Experimental protocols are detailed in Sec.~\ref{sec:supp_model_free_ablations}.

\paragraph{Base-policy feedback.}
Per-step replanning mitigates the performance loss of longer-chunk BC policies in simulation (A, left). Under reduced actuator P-gains, subsequent latent residual RL over fixed, less responsive BC bases retains substantial performance gaps despite extensive parallel training (A, right). Thus, residual learning does not readily compensate for reduced base-policy feedback in this setting.

\paragraph{Preserving base-policy commands.}
With single-step chunks and $d_z=22$ matching the hand-action dimension, a learned action interface lowers BC performance on all four tasks (B, left), and full decoded-action execution learns less effectively than residual execution (B, right). These results support retaining fine-grained base-policy commands even without temporal or dimensional compression; they do not isolate reconstruction error as the sole cause.

\paragraph{Exploration chunk-length sensitivity.}
With $d_z=16$ fixed and time-matched discounts, eight-step residual chunks perform well across the three dynamic tasks, while sixteen-step chunks perform best on \trianglepassup (C), consistent with simulation-based selection. Shorter chunks generally yield smaller gains, whereas extending the horizon to sixteen steps does not uniformly improve learning.

\paragraph{Latent-dimension sensitivity.}
At task-specific default chunk lengths, the empirical default $d_z=16$ provides the strongest overall performance; $d_z = 64$ underperforms it on the four tasks. The eight-dimensional variant can improve quickly but does not consistently match the final performance of 16 dimensions (D). 
In this batch of latent-dimension runs, training was stopped early when, on entering the final 100 epochs, the operator judged further improvement unlikely, to save hardware time; panel D therefore plots all curves, including the default, only up to the last epoch reached by the shortest run (epoch 930), so that the settings are compared over the same epoch range.

\paragraph{Latent offset sensitivity.}
The latent offset scale $\beta_r$ controls modulation before decoding. The empirical default $0.25$ provides the most consistent performance across tasks (E). Smaller offsets ($0.125$) yield lower final performance, while larger offsets ($0.5$ or $0.75$) can be competitive but produce greater fluctuations or degradation.

\paragraph{Residual action-scale sensitivity.}
The post-decoding scale $\alpha_r$ controls how strongly residuals modify base commands.
The empirical default $\alpha_r=0.25$ reaches higher final reward than $0.125$ on \passvzerons, \passvonens, \sonicdownns, and \trianglepassup (F).

\subsection*{Cross-Rollout Variation in Object Motion}
\label{sec:supp_additional_rollout_variation}

\subsubsection*{Initialization and Evaluation Scope}

\begin{figure}[htbp]
  \centering
  \includegraphics[width=0.8\textwidth]{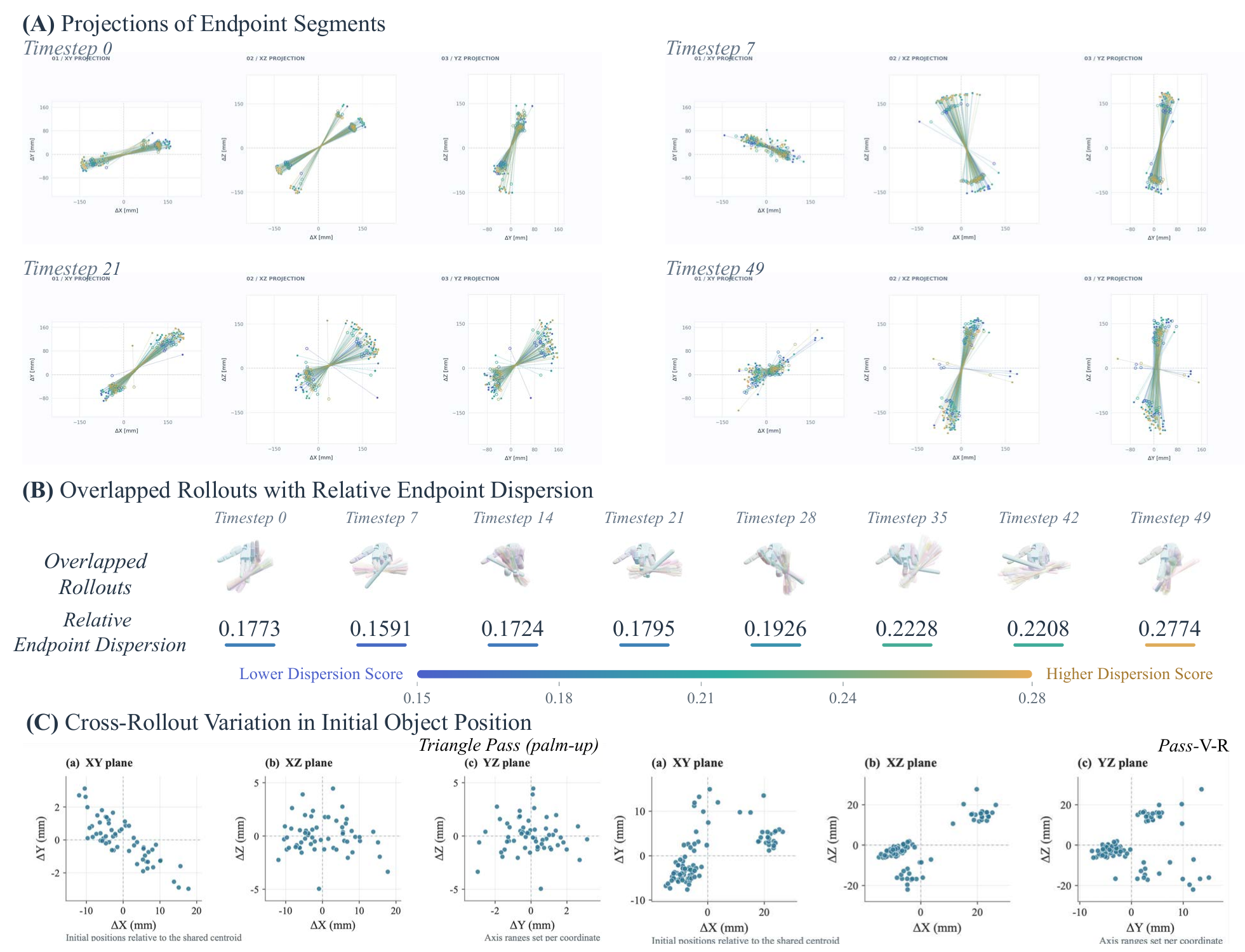}
  \caption{\footnotesize
  \textbf{Cross-Rollout Variation.} \textbf{(A)} Object endpoint segments projected onto the $xy$, $yz$, and $xz$ planes and overlaid at corresponding timesteps across 93 repeated \passvzero rollouts. \textbf{(B)} Overlaid object configurations and cross-rollout endpoint dispersion. \textbf{(C)} Initial object center positions across 59 successful \trianglepassup and 93 \passvzero rollouts, projected onto the same planes. Relative endpoint dispersion is $D/L$ for the $L = 300$~mm reference segment
(longer than the 200~mm pen). 
  }
  \label{fig:rollout_distribution}
\end{figure}

\begin{figure}[htbp]
  \centering
  \includegraphics[width=0.8\textwidth]{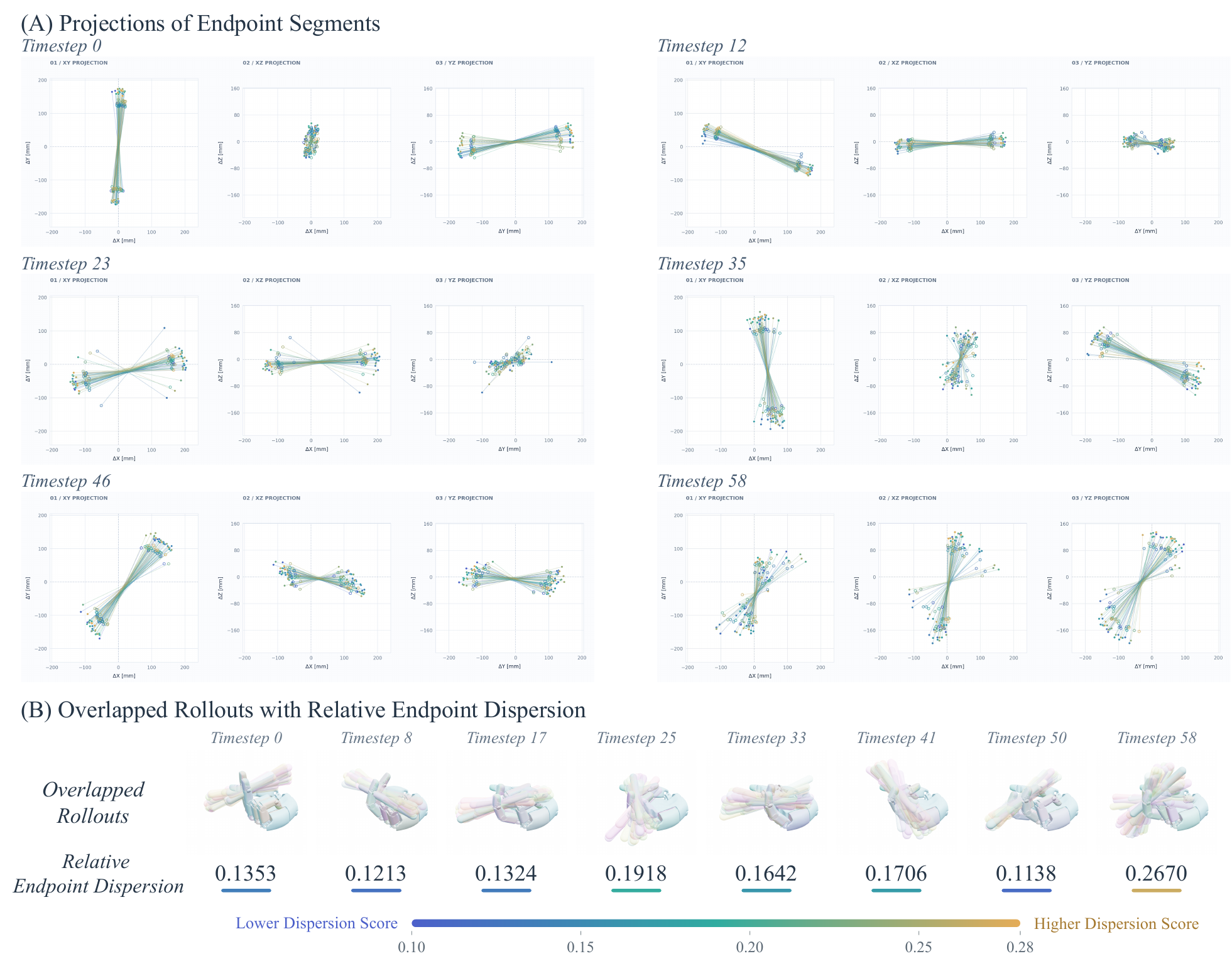}
  \caption{\footnotesize
  \textbf{Cross-Rollout Variation on \passvthreens.} \textbf{(A)} Projected endpoint segments overlaid at corresponding timesteps across 59 repeated rollouts. \textbf{(B)} Overlaid object configurations and cross-rollout endpoint dispersion. Relative endpoint dispersion is $D/L$ for the $L = 300$~mm reference segment
(longer than the physical pen).
  }
  \label{fig:passv3_objecttrajectoryrollout_distribution}
\end{figure}

Figure~\ref{fig:rollout_distribution}(A,B) overlays repeated \passvzero rollouts, showing increasing projected-endpoint dispersion even among successful trials. 
We calculate ``cross-rollout endpoint dispersion'' as a metric for describing
cross-rollout object variations at each timestep. For rollout $i$ at frame $t$,
let $\mathbf{p}_i \in \mathbb{R}^3$ be the tracked object center and
$\boldsymbol{\ell}_i$ the unit vector along the object's longitudinal axis,
expressed in the common transformed frame in millimetres. We represent the
object by a reference segment of length $L = 300~\mathrm{mm}$, centered on
$\mathbf{p}_i$ and aligned with $\boldsymbol{\ell}_i$, with endpoints
$\mathbf{a}_i = \mathbf{p}_i + \tfrac{L}{2}\boldsymbol{\ell}_i$ and
$\mathbf{b}_i = \mathbf{p}_i - \tfrac{L}{2}\boldsymbol{\ell}_i$. This is the
segment used for visualization in Figs.~\ref{fig:rollout_distribution}
and~\ref{fig:passv3_objecttrajectoryrollout_distribution}, and it extends beyond the 200~mm pen.
For $N_t \ge 2$ valid trials at frame $t$, the endpoint-order-invariant pair
distance and RMS dispersion are
\begin{align}
  d_{ij}^2(t) &= \tfrac{1}{2}\min\Bigl\{
      \lVert \mathbf{a}_i-\mathbf{a}_j \rVert^2
      + \lVert \mathbf{b}_i-\mathbf{b}_j \rVert^2,\;
      \lVert \mathbf{a}_i-\mathbf{b}_j \rVert^2
      + \lVert \mathbf{b}_i-\mathbf{a}_j \rVert^2
    \Bigr\}, \label{eq:pair_distance}\\
  D(t) &= \Biggl[\frac{2}{N_t(N_t-1)}
      \sum_{i<j} d_{ij}^2(t)\Biggr]^{1/2}. \label{eq:rms_dispersion}
\end{align}
Equivalently,
$d_{ij}^2 = \lVert \mathbf{p}_i-\mathbf{p}_j \rVert^2
 + \tfrac{L^2}{4}\min\bigl\{
   \lVert \boldsymbol{\ell}_i-\boldsymbol{\ell}_j \rVert^2,
   \lVert \boldsymbol{\ell}_i+\boldsymbol{\ell}_j \rVert^2 \bigr\}$,
so $D$ combines variation in object position and in axis direction, with $L$
setting their relative weight. $D$ vanishes for coincident segments, is
invariant to common rigid transforms, and excludes axial rotation. 
$D$ also includes the orientation-dependent reference-point offset described below.
The relative
endpoint dispersion reported in Figs.~\ref{fig:rollout_distribution}(B)
and~\ref{fig:passv3_objecttrajectoryrollout_distribution}(B) is $\widetilde{D} = D/L$ with
$L = 300~\mathrm{mm}$; it is a normalized summary rather than a fraction of the
pen length.

Figure~\ref{fig:passv3_objecttrajectoryrollout_distribution} provides the corresponding trajectory visualization for \passvthreens.

\subsubsection*{Variation in Initial Object Position}

Successful \trianglepassup and \passvzero initial positions appear in Fig.~\ref{fig:rollout_distribution}(C). For \trianglepassupns, spans approach $3\,\mathrm{cm}$ along $x$ and $1\,\mathrm{cm}$ along $z$, with $y$ (object width) constrained by grasping. The vertical span is approximately half the object height; the longitudinal span, though below its $15\,\mathrm{cm}$ length, is comparable to the approximately $3\,\mathrm{cm}$ finger-contact separation, materially changing coordination requirements.


For \passvzerons, the tracked reference point lies about $1.8\,\mathrm{cm}$ from the pen's midpoint, so rollouts with the pen in the opposite end-to-end orientation form a second group displaced by about 3.6 cm along the pen axis (C, right). After removing this orientation-dependent offset, the vertical span is approximately $3\,\mathrm{cm}$. Excessive height above the palm hinders the second transfer after the short end rotates upward; insufficient height hinders the first. Extreme offsets remain challenging. These successful-trial distributions demonstrate observed tolerance, not uniform success throughout their enclosed ranges.

\begin{figure}[htbp]
  \centering
  \includegraphics[width=\textwidth]{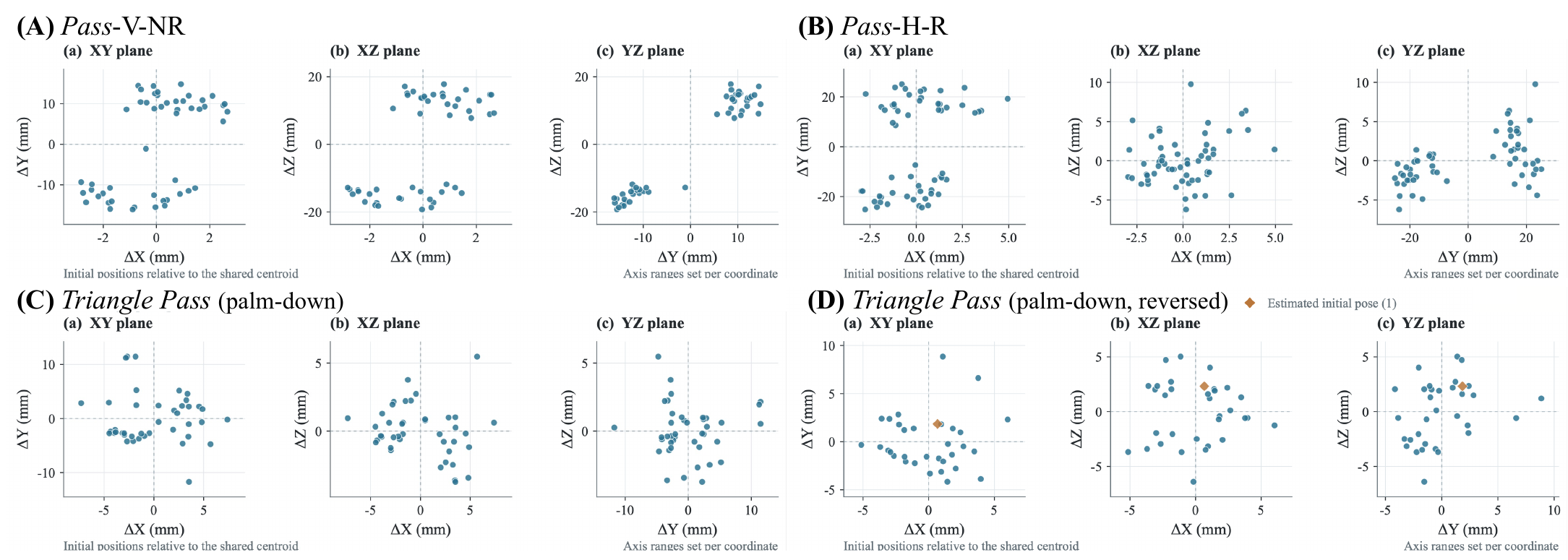}
  \caption{\footnotesize
  \textbf{Variation in Initial Object Position.} Initial object center positions across successful rollouts, projected onto the $xy$, $xz$, and $yz$ planes. \textbf{(A)} \passvonens, 50 rollouts. \textbf{(B)} \passvthreens, 59 rollouts. \textbf{(C)} \trianglepassdownns, 41 rollouts. \textbf{(D)} \trianglepassdownreversedns, 33 rollouts, including one estimated initial position (diamond). The \passvone and \passvthree panels contain the same two orientation groups.
  }
  \label{fig:rollout_distribution_initial_obj_position_others}
\end{figure}

Figure~\ref{fig:rollout_distribution_initial_obj_position_others} shows the initial-position distributions for \passvonens, \passvthreens, \trianglepassdownns, and \trianglepassdownreversedns.

\subsection*{Baseline Analyses and Representative Failures}

\paragraph{Simulation-based conservative tuning.}
BPG-SAC, RFS, and the joint-space residual baseline in Fig.~\ref{fig:structural_and_residual_exploration}(A) undergo additional tuning entirely in simulation. Before large-scale pretraining, exploration noise is calibrated to retain at least 90\% of the frozen starting controller's noise-free mean return. Subsequent searches start from pretrained policies and critics, using conservative exploration and update settings and a penalty on deviation from base commands. SAC tuning uses two environments for 1,000 epochs; RFS collects a fixed dataset in two simulation environments and performs 1,000 offline TD3+BC training epochs, matching its hardware-stage algorithm. Real-world training uses the simulation-selected hyperparameters; detailed protocols and search ranges appear in Supplementary Section~S2.
The two full-action variants in Fig.~\ref{fig:structural_and_residual_exploration}(E) also receive simulation-only latent-offset calibration and conservative tuning from pretrained actors and critics. Their calibration retains at least 90\% of the mean return from unmodulated reference latents under the corresponding execution mode; the full protocol is given in Sec.~\nameref{sec:supp_model_free_ablations}.

\begin{figure}[htbp]
  \centering
  \includegraphics[width=\textwidth]{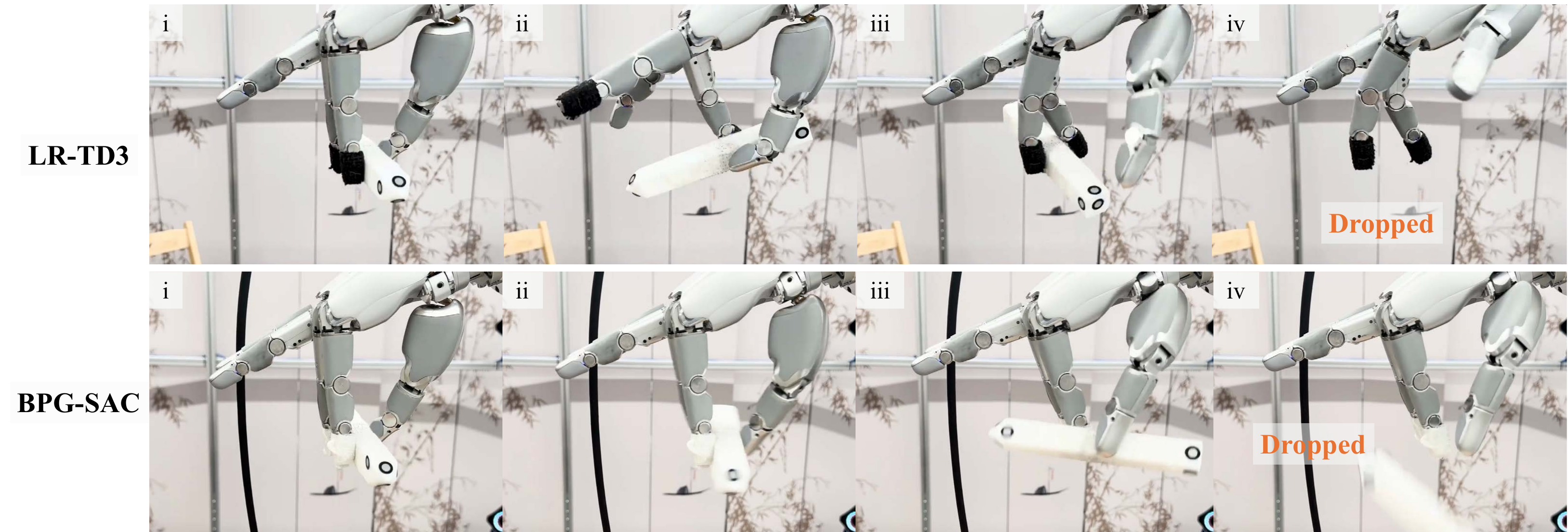}
  \caption{\footnotesize
  \textbf{Representative Baseline Failures.} Rollout frames (i--iv) of LR-TD3 and BPG-SAC policies on \trianglepassdownns, each ending in an object drop.
  }
  \label{fig:baseline_failures}
\end{figure}

Even slow rotation tasks require responsive closed-loop control and fine-grained actions during both execution and real-world training, as fingertip-supported rotation is highly sensitive to contact variations. We summarize each baseline below; training protocols and parameter sweeps are given in Supplementary Section~S2.

\paragraph{Joint-wise dynamics refinement (NDM).}
NDM retains DexNDM's proprioceptive formulation, learning and aligning joint transitions from Chaos Box data. After two refinement iterations from the first-stage policy, it achieves lower average rotation than our second-stage policy on all ten tasks. The ``w/ delta-action model'' ablation in Fig.~\ref{fig:model_based_abl}(A) complements this baseline with NDM-style joint-wise compensation trained on our mocap-annotated task trajectories with joint-position and object-axis supervision.

\paragraph{Chunk-based methods (LR-TD3 and SERNF).}
LR-TD3 achieves 0\% success across the ten tasks after real-world fine-tuning. Its reference flow exhibits execution drift on hardware before any RL update, despite sweeps of prediction and execution horizons, and fine-tuning does not recover successful execution (Fig.~\ref{fig:baseline_failures}). Because complete actions pass through the learned representation, reduced feedback responsiveness and distortion of fine-grained commands may contribute. SERNF achieves lower mean single-episode success and shorter sustained rotations than our method, with degradation already evident after action-chunk distillation on some tasks. It executes ten-action chunks on all tasks except \thumbaroundns, and reduced feedback responsiveness and difficulty optimizing its normalizing-flow architecture and high-dimensional chunk outputs may contribute. Both outcomes are consistent with the sensitivity to action chunking and learned action interfaces in Fig.~\ref{fig:structural_and_residual_exploration}(D).

\paragraph{Per-step joint-space methods (BPG-SAC and RFS).}
Both remain below the second-stage controller on all ten tasks despite conservative tuning. BPG-SAC combines guidance from the refined base and fixed stage-one correction policies, replay priors, and conservatively tuned updates; nevertheless, real-world exploration can disrupt critical contacts, and the resulting drops may limit useful learning signals (Fig.~\ref{fig:baseline_failures}). RFS uses online PPO for large-scale simulation pretraining, followed by simulation-only parameter selection for offline TD3+BC adaptation on real-world data. Because human takeover does not achieve effective recovery in our setup (Fig.~\ref{fig:human_takeover_failures}; Table~\ref{tab:supp_two_operator_recovery}), its offline dataset is collected through policy exploration. RFS retains performance after behavior cloning and improves in simulation, but deteriorates during offline adaptation. Its $2n_{\mathrm{dof}}$-dimensional output lacks temporal structure and its action residual lacks a learned spatial prior, potentially yielding poorly coordinated exploration that disrupts contacts and limits useful offline-data coverage.

\subsection*{Comparison with More Existing Work}
\label{sec:supp_existing_work_comparison}

\paragraph{Wuji-MJLab.}
We qualitatively compare our task conditions with the publicly demonstrated PenSpin setup in Wuji-MJLab (\href{https://github.com/wuji-technology/wuji-mjlab/releases}{https://github.com/wuji-technology/wuji-mjlab/releases}).

\noindent\textit{Motion and contact demands.}
Our ten tasks draw on human pen-spinning styles and robot-adapted variants (\href{https://www.youtube.com/watch?v=fayyJk60ZQg}{www.youtube.com/watch?v=fayyJk60ZQg}, \href{https://www.youtube.com/watch?v=9vF9OmoKlF4}{www.youtube.com/watch?v=9vF9OmoKlF4}), testing forceful actions, dynamic object motion, underactuated rotation, and constrained finger gaiting. Even slower tasks repeatedly rely on two-finger support, offering less contact redundancy than the thumb-assisted rotation shown in the public demonstration.

\noindent\textit{Object geometry and friction.}
The documented Wuji pen has an outer diameter of approximately $1.69\,\mathrm{cm}$ including ribs and permits friction-enhancing tape wrapping. Our main-experiment cylinders have diameters of $1.4\,\mathrm{cm}$ (\pass and \thumbaroundns) and $1.08\,\mathrm{cm}$ (\sonicdownns). Millimetre-scale differences affect finger placement and available contact area. Ribs, tape, and the demonstrated glove covering can enhance grip and reduce slipping. Our pens receive only simple anti-slip treatment, and Sharpa is used without a glove; only the three \trianglepass tasks add fingertip tape to compensate for slippery nail surfaces.

\noindent\textit{Deployment sensing.}
Wuji-MJLab uses camera-based pen-pose observations during deployment. Our policies use hand proprioception alone, without external object-state observations or tactile sensing, requiring contact and motion regulation from more limited feedback.

\paragraph{TopoRetarget.}
Similar differences apply to TopoRetarget~\cite{Wu2026TopoRetarget}: its demonstrated setup uses a glove-covered hand, a larger pen, and camera-based pen-pose observations during deployment. Our experiments use smaller pens, operate Sharpa without a glove, and rely solely on hand proprioception during deployment.

\paragraph{TeleDexter pen-spinning demonstration.}
We qualitatively compare with the public SharpaWave pen-spinning demonstration of TeleDexter~\cite{Li2026TeleDexter} (\href{https://bigai-dex.github.io/blog/teledexter/}{https://bigai-dex.github.io/blog/teledexter/}). 

\noindent\textit{Motion and contact conditions.}
In the right demo, the pen and several finger surfaces extending below the fingertips appear extensively tape-wrapped, which can enhance grip. The pen begins low across the palm and exhibits visible out-of-plane tilting during rotation. This low placement may allow the tilted pen to contact the palm and receive additional support; such support, if present, can ease contact maintenance and alter the demands of underactuated object motion. Our tasks emphasize dynamic rotations and constrained finger gaiting with limited support. 
The left, uninterrupted take includes finger-slot transfers resembling \passvzerons.
At the displayed playback rate, TeleDexter's transfers appear slower and less uniform than our \passvzero executions.

\noindent\textit{Deployment sensing.}
TeleDexter uses a tracking-based controller with external hand and object-pose information. Our final manipulation policies use hand proprioception alone, without external object-state feedback.

\noindent\textit{Repeated-trial evaluation.}
The website presents an uninterrupted take and a recovery example, but reports neither a pen-specific evaluation over repeated independent trials nor aggregate continuous-run statistics. In comparison, our evaluation emphasizes execution reliability through repeated trials across all ten tasks.

\paragraph{ContactNets.}
ContactNets~\cite{Pfrommer2020ContactNetsLO} identifies contact geometry and friction mappings from observed state transitions by learning signed-distance functions and contact Jacobians, including a friction-weighted tangential mapping. Both approaches retain physics-based contact solvers; ContactNets keeps the prescribed contact-law formulation. Its evaluation focuses on trajectory prediction, without predictive-control or model-based RL demonstrations, leaving its expressivity--generalization balance for policy improvement untested.
We retain analytical contact geometry and Jacobians and instead compensate contact-constraint responses through shared per-contact modulation of reference accelerations and inverse regularization weights for policy refinement.

\paragraph{NeuralSim.}
NeuralSim~\cite{Heiden2020NeuralSimAD} augments analytical simulation with learned friction and passive-force models, trained by differentiating trajectory losses through the simulator. In its planar-pushing experiment, friction corrections use object pose and velocity together with contact geometry; transfer of shared contact-local corrections across complex articulated configurations is not evaluated. Its quadruped experiment concerns MPC-solver imitation, leaving learned-model policy refinement for high-DoF dynamic dexterous manipulation untested. Our shared per-contact constraint compensation learns through RL trajectory tracking without simulator derivatives and supports the manipulation policy refinement.

\paragraph{NeRD and FlashNeRD.}
NeRD~\cite{Xu2025NeuralRD} predicts robot-level state transitions from state, torque, and contact histories, replacing analytical dynamics and contact solvers. FlashNeRD~\cite{Ataei2026FlashNeRDPC} adds faster inference and variable-contact encoding; its manipulation mode predicts robot free dynamics before a coupled analytical contact solve. 
We instead learn shared per-contact corrections to constraint responses, using this finer granularity to encourage cross-contact generalization while retaining analytical dynamics and the coupled solver.
In our preliminary NeRD-inspired implementation, the learned hybrid did not support effective policy refinement in our setup.

\paragraph{Latent-space policy refinement.}
ZPRL~\cite{Yu2026BeyondAR} perturbs an observation bottleneck; VLA post-training~\cite{Zhu2026TowardsHD} and LAMP~\cite{Yang2026LAMPLM} refine latent hand commands decoded into full actions. RFS~\cite{Su2026RFS} combines flow-noise steering with action-space residuals. These methods also exploit learned priors to guide exploration. Our distinction is to impose spatial and temporal structure on the residual corrections themselves: low-dimensional latent modulation generates coordinated, task-relevant correction sequences as differences between chunks decoded from modulated and reference latents, reducing the search dimension relative to direct joint-action sequences. These corrections are added to base-policy commands recomputed from the latest observation at every control step. This separates the temporal structure of exploration from execution feedback frequency, preserving high-frequency closed-loop control without reconstructing complete base-policy commands.

%% file: supp/supp_caption.tex

\paragraph{Caption for Movie S1.}
\textbf{Overview.}
Real-time (1$\times$) execution of each task, reliability evaluation over 50 repeated trials, consecutive multi-episode execution, zero-shot generalization and adaptation to new task variants, and robustness to human disturbances.

\paragraph{Caption for Movie S2.}
\textbf{\textit{Pass}-V-R.}
Continuous multi-episode rotation, single-episode evaluation (50/50 successful trials), and sustained rotation across 20 trials (mean 4.025 episodes).

\paragraph{Caption for Movie S3.}
\textbf{\textit{Pass}-V-NR.}
Continuous multi-episode rotation and single-episode evaluation (50/50).

\paragraph{Caption for Movie S4.}
\textbf{\textit{Pass}-H-NR.}
An alternative transition found by real-world training, in which the pen spins about 180$^\circ$ around the middle fingernail; continuous multi-episode rotation; and single-episode evaluation (45/50).

\paragraph{Caption for Movie S5.}
\textbf{\textit{Pass}-H-R.}
Continuous multi-episode rotation and single-episode evaluation (48/50).

\paragraph{Caption for Movie S6.}
\textbf{\textit{Thumb-Around}.}
Single-episode demonstrations and single-episode evaluations on the nominal object (49/50) and, after adaptation, on a compliant TPU object (48/50).

\paragraph{Caption for Movie S7.}
\textbf{\textit{Sonic} (down).}
Single-episode demonstrations and single-episode evaluations of the nominal style (50/50) and a variant with more Charge-style rotations between transfers (50/50).

\paragraph{Caption for Movie S8.}
\textbf{\textit{Triangle Pass} (palm-up).}
A continuous 91-episode run, single-episode evaluation (50/50), and sustained rotation across five trials (mean 39.9 episodes).

\paragraph{Caption for Movie S9.}
\textbf{\textit{Triangle Pass} (palm-down).}
Single-episode evaluation (50/50) and sustained rotation across five trials (mean 40.1 episodes; up to 77).

\paragraph{Caption for Movie S10.}
\textbf{\textit{Triangle Pass} (palm-down, reversed).}
Single-episode evaluation (50/50) and sustained rotation across five trials (mean 10.7 episodes; up to 29).

\paragraph{Caption for Movie S11.}
\textbf{\textit{Four-Finger-Pass} (palm-down, LEAP).}
Continuous rotation and single-episode evaluations on the nominal object (48/50) and, after adaptation, on an elliptical cylinder (50/50).

\paragraph{Caption for Movie S12.}
\textbf{Beyond pen spinning: \textit{Four-Finger-Pass} with a board.}
An uninterrupted 134-episode run lasting over 30 minutes (real time), rotating a 15 $\times$ 10 $\times$ 2 cm board on the Sharpa hand.

\paragraph{Caption for Movie S13.}
\textbf{Zero-shot generalization to new objects.}
Policies applied without additional training to objects including hexagonal and octagonal prisms, compliant (bendable) objects, L-, C-, and T-shaped objects, a long board, and a 100 g book-like object.

\paragraph{Caption for Movie S14.}
\textbf{Adaptation to new task variants.}
Policies adapted with 250--500 additional real-world rollouts per variant, including a compliant TPU object (\textit{Thumb-Around}), an elliptical cylinder (\textit{Four-Finger-Pass}, LEAP), and thinner objects, cylinders, octagonal prisms, and a 15 $\times$ 2 $\times$ 5 cm board (\textit{Triangle Pass}).

\paragraph{Caption for Movie S15.}
\textbf{Robustness to human disturbances.}
Closed-loop execution under repeated human pushes and drags on \textit{Triangle Pass} (palm-up), \textit{Pass}-V-R, and \textit{Triangle Pass} (palm-down).

\paragraph{Caption for Movie S16.}
\textbf{Human takeover.}
Full-hand and finger-masked partial takeover, teleoperated with a Manus glove, fail to recover critical failures on \textit{Pass}-V-NR, \textit{Pass}-H-NR, and \textit{Triangle Pass} (palm-up).

\paragraph{Caption for Data S1.}
\textbf{Evaluation and analysis data.}
Values plotted in Figs.~2, 3, 5, and 6, figs.~S7 and S8, and the bar charts in Fig.~7 and fig.~S9.

\paragraph{Caption for Data S2.}
\textbf{Learning-curve data.}
Real-world and simulated training curves in Fig.~7 and fig.~S9.

\paragraph{Caption for Data S3.}
\textbf{Cross-rollout variation data.}
Values plotted in figs.~S10 to S12.